\documentclass[10pt,twocolumn,letterpaper]{article}

\usepackage{iccv}
\usepackage{times}
\usepackage{epsfig}
\usepackage{graphicx}
\usepackage{amsmath}
\usepackage{amssymb}
\usepackage{subfigure}
\usepackage{multirow}
\usepackage{color}
\usepackage{tabu}
\usepackage{array}

\usepackage[breaklinks=true,bookmarks=false]{hyperref}

\iccvfinalcopy 

\def\iccvPaperID{****} 

\begin{document}

\title{Pixel-aware Deep Function-mixture Network for Spectral Super-Resolution}

\author{Lei Zhang$^1$, Zhiqiang Lang$^2$, Peng Wang$^3$, Wei Wei$^2$, Shengcai Liao$^1$, Ling Shao$^1$, Yanning Zhang$^2$\\
$^1$Institute of Artificial Intelligence, United Arab Emirates\\
$^2$School of Computer Science and Engineering, Northwestern Polytechnical University, Xi’an, 710072, China\\
$^3$School of Computer Science, The University of Adelaide, Australia\\
}

\maketitle

\begin{abstract}
Spectral super-resolution (SSR) aims at generating a hyperspectral image (HSI) from a given RGB image. Recently, a promising direction for SSR is to learn a complicated mapping function from the RGB image to the HSI counterpart using a deep convolutional neural network. This essentially involves mapping the RGB context within a size-specific receptive field centered at each pixel to its spectrum in the HSI. The focus thereon is to appropriately determine the receptive field size and establish the mapping function from RGB context to the corresponding spectrum. Due to their differences in category or spatial position, pixels in HSIs often require different-sized receptive fields and distinct mapping functions. However, few efforts have been invested to explicitly exploit this prior.

To address this problem, we propose a pixel-aware deep function-mixture network for SSR, which is composed of a new class of modules, termed function-mixture (FM) blocks. Each FM block is equipped with some {\textbf{basis functions}}, i.e., parallel subnets of different-sized receptive fields. Besides, it incorporates an extra subnet as a {\textbf{mixing function}} to generate pixel-wise weights, and then linearly mixes the outputs of all basis functions with those generated weights. This enables us to pixel-wisely determine the receptive field size and the mapping function. Moreover, we stack several such FM blocks to further increase the flexibility of the network in learning the pixel-wise mapping. To encourage feature reuse, intermediate features generated by the FM blocks are fused in late stage, which proves to be effective for boosting the SSR performance. Experimental results on three benchmark HSI datasets demonstrate the superiority of the proposed method.
\end{abstract}

\begin{figure}[htbp]
\centering
\includegraphics[height=2in, width=3.2in]{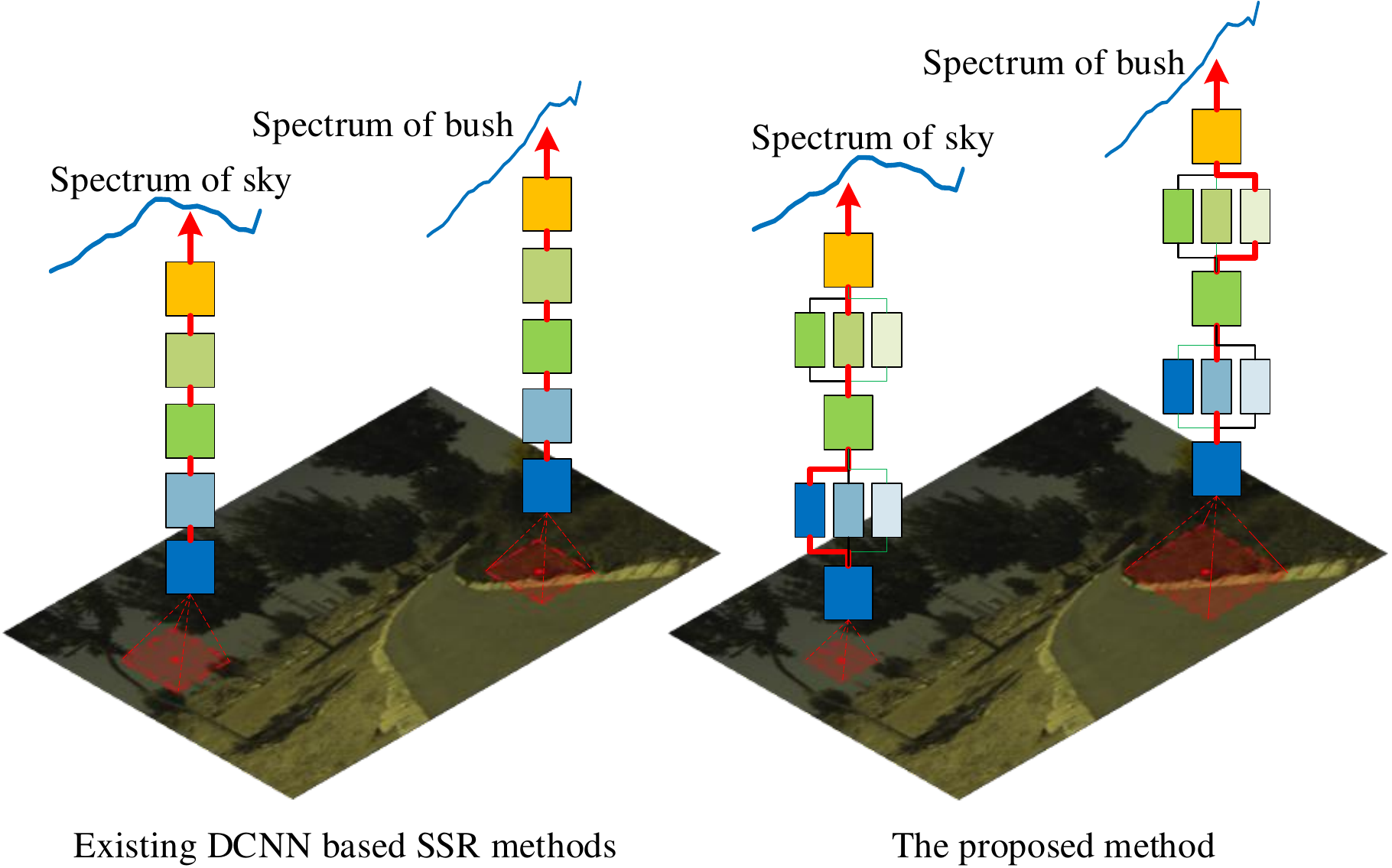}
\caption{The difference between existing DCNN based SSR methods and the proposed method. Existing methods often take a fixed-sized receptive field and learn a universal mapping function for all pixels, while the proposed method can adaptively determine the receptive field size and the mapping function for each pixel.}
\label{fig:idea}
\vspace{-0.3cm}
\end{figure}

\section{Introduction}
Hyperspectral imaging is the technique that captures the reflectance of scenes with extremely high spectral resolution (\eg, 10$nm$)~\cite{chakrabarti2011statistics}. The captured hyperspectral image (HSI) often contains hundreds or thousands of spectral bands, and each pixel has a spectrum~\cite{chakrabarti2011statistics,zhang2018cluster}. Profiting from the abundant spectral information, HSIs have been widely applied to various tasks, \eg, classification~\cite{akhtar2018nonparametric}, detection~\cite{manolakis2002detection} and tracking~\cite{van2010tracking} \etc However, the expense of obtaining such spectral information is to increase the pixel size on the sensor, which inevitably limits the spatial resolution of HSIs~\cite{lanaras2015hyperspectral}. Thus, it is crucial to investigate how to generate high-spatial-resolution (HSR) HSIs.

Different from convnetioanl HSIs super-resolution~\cite{mei2017hyperspectral,zhang2018exploiting} that directly improves the spatial resolution of a given HSI, spectral super-resolution (SSR)~\cite{arad2016sparse,xiong2017hscnn} adopts an alternative way and attempts to produce an HSR HSI by increasing the spectral resolution of a given RGB image with satisfactory spatial resolution. Early SSR methods~\cite{arad2016sparse,aeschbacher2017defense,jia2017rgb} often formulate SSR as a linear inverse problem, and exploit the inherent low-level statistic of HSR HSIs as priors. However, due to the limited expressive capacity of their handcrafted prior models, these methods fail to well generalize to challenging cases. Recently, witnessing the great success of deep convolutional neural networks (DCNNs) in a wide range of tasks~\cite{simonyan2014very,he2016deep,he2017mask}, increasing efforts have been invested to learn a DCNN based mapping function to directly transform the RGB image into an HSI~\cite{alvarez2017adversarial,arad2017filter,shi2018hscnn+,fu2018joint}. These methods essentially involve mapping the RGB context within a size-specific receptive field centered at each pixel to its spectrum in the HSI, as shown in Figure~\ref{fig:idea}. The focus thereon is to appropriately determine the receptive field size and establish the mapping function from RGB context to the corresponding spectrum. Due to the difference in category or spatial position, pixels in HSIs often necessitate collecting different RGB information and adopting various recovery schemes for SSR. 
Therefore, to obtain an accurate DCNN based SSR approach, it is crucial to adaptively determine the receptive field size and the RGB-to-spectrum mapping function for each pixel. However, most existing DCNN based SSR methods treat all pixels in HSIs equally and learn a universal mapping function with a fixed-sized receptive field, as shown in Figure~\ref{fig:idea}.

In this study, we present a pixel-aware deep function-mixture network for SSR, which is flexible to pixel-wisely determine the receptive field size and the mapping function. Specifically, we first develop a new module, termed the function-mixture (FM) block. Each FM block consists of some parallel DCNN based subnets, among which one is termed the {\textit{mixing function}} and the remaining are termed {\textit{basis functions}}. The basis functions take different-sized receptive fields and learn distinct mapping schemes; while the mixture function generates pixel-wise weights to linearly mix the outputs of the basis functions. In this way, the pixel-wise weights can determine a specific information flow for each pixel and consequently benefit the network to choose appropriate RGB context as well as the mapping function for spectrum recovery. Then, we stack several such FM blocks to further improve the flexibility of the network in learning the pixel-wise mapping. Furthermore, to encourage feature reuse, the intermediate features generated by the FM blocks are fused in late stage, which proves to be effective for boosting the SSR performance. Experimental evaluation on three benchmark HSI datasets shows the superiority of the proposed approach for SSR.

In summary, we mainly contribute in three aspects. {\textbf{\textit{i)}} We present an effective pixel-aware deep function-mixture network for SSR, which is flexible to learn the pixel-wise RGB-to-spectrum mapping. To our best knowledge, this is the first attempt to explore this in SSR. {\textbf{\textit{ii)}} We design a new FM module, which is flexible to plug in any modern DCNN architectures; {\textbf{\textit{iii)}} We demonstrate new state-of-the-art performance on three benchmark SSR datasets.

\section{Related Work}
We first review the existing approaches for SSR and then introduce some techniques related to this work.  

\vspace{-0.4cm}
\paragraph{Spectral Super-resolution} Early methods mainly focus on exploiting appropriate image priors to regularize the linear inverse SSR problem. For example, Arad and Aeschbacher \etal~\cite{arad2016sparse,aeschbacher2017defense} investigated the sparsity of the latent HSI on a pre-trained over-complete spectral dictionary. Jia \etal~\cite{jia2017rgb} considered the manifold structure of HSIs in a low-dimensional space. Recently, most methods turn to learning a deep mapping function from the RGB image to an HSI. For example, Alvarez-Gila et al.~\cite{alvarez2017adversarial} implemented the mapping function using an U-Net architecture~\cite{ronneberger2015u} and trained it based on both the mean-square-error (MSE) loss and the adversarial loss~\cite{goodfellow2014generative}. Shi \etal~\cite{shi2018hscnn+} developed a deep residual network consisting of residual blocks to learn the mapping function. Despite obtaining impressive performance for SSR, these methods are limited by learning a universal RGB-to-spectrum mapping function for all pixels in HSIs. This leaves space for learning more flexible and adaptive mapping function.

\vspace{-0.4cm}
\paragraph{Receptive Field in DCNNs} Receptive field is an important concept in the DCNN, which determines the sensing space of a convolutional neuron. There are many efforts dedicating to adjusting the size or shape of the receptive field~\cite{yu2015multi,wei2017learning,dai2017deformable} to meet the requirement of specific tasks at hand. Thereinto, dilated convolution~\cite{yu2015multi} or kernel separation~\cite{seif2018large} were often utilized to enlarge the receptive field. Recently, Wei \etal~\cite{wei2017learning} changed the receptive field by inflating or shrinking the feature maps using two affine transformation layers. Dai \etal~\cite{dai2017deformable} proposed to adaptively determine the context within the receptive field by estimating the offsets of pixels to the central pixel using an additional convolution layer. In contrast, we take a totally different direction and learn the pixel-wise receptive field size by mixing some basis function with different receptive field sizes.

\begin{figure*}
\centering
\includegraphics[height=1.1in, width=6.2in]{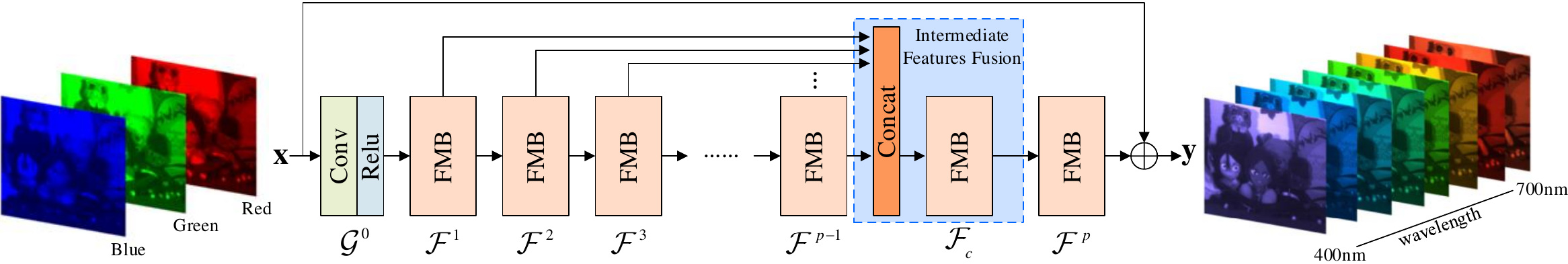}
\caption{Architecture of the proposed pixel-aware deep function-mixture network. FMB denotes the function-mixture block.}
\label{fig:FMNet}
\vspace{-0.3cm}
\end{figure*}

\vspace{-0.4cm}
\paragraph{Multi-column Network} Multi-column network~\cite{cirecsan2012multi} is a specicial type of network that feeds the input into several parallel DCNNs (\ie, columns), and then aggregates their outputs for final prediction. With the ability of using more context information, the multi-column network (MCNet) often shows better generalization capacity than that with only a single column in various tasks, \eg, classification~\cite{cirecsan2012multi}, image processing~\cite{agostinelli2013adaptive}, counting~\cite{zhang2016single} \etc. Although we also adopt a similar multi-column structure in our module design, the proposed network is obviously different from these existing multi-column networks~\cite{cirecsan2012multi,zhang2016single,agostinelli2013adaptive}. First, MCNet employs a separation-and-aggregation architecture which processes the input with parallel columns and then aggregates the outputs of all columns for output. In contrast, we adopt a recursive separation-and-aggregation architecture by stacking multiple FM modules, each of which can be viewed as an enhanced multi-column module, as shown in Figure~\ref{fig:idea},~\ref{fig:FM}. Second, when applied to SSR, MCNet still learns a universal mapping function and fails to flexibly handle each pixel in an explicit way. In contrast, the proposed FM block incorporates a mixing function to generate pixel-wise weights and mix the outputs of all basis functions. This enables to flexibly customize the pixel-wise mapping function. In addition, we fuse the intermediate feature generated by FM blocks in the network for feature reuse.

\section{Proposed Network}
In this section, we present the technical details of the proposed pixel-aware deep function-mixture network, as shown in Figure~\ref{fig:FMNet}. The proposed network adopts a global residual architecture as~\cite{kim2016accurate}. Its backbone is constructed by stacking multiple FM blocks and fusing the intermediate features generated by previous FM block with skip connections. In the following, we will first introduce the basic FM block. Then, we will introduce how to incorporate multiple FM blocks and the intermediate features fusion into the proposed network for performance enhancement.

\subsection{Function-mixture Block}\label{subsec:FMB}
The proposed network essentially establishes an end-to-end mapping function from an RGB image to the HSI counterpart, and thus each FM block plays the role of a mapping subfunction. In this study, we attempt to utilize the FM block to adaptively determine the receptive field size and the mapping function for each pixel, \ie, to obtain a pixel-dependent mapping subfunction. To this end, a direct solution is to introduce an additional hypernetwork~\cite{ha2016hypernetworks,jia2016dynamic} to adaptively generate the subfunction parameters for each pixel. However, this will greatly increase the computational complexity as well as the training difficulty~\cite{ha2016hypernetworks}. To avoid this problem, we borrow the idea in function approximation~\cite{cybenko1989approximation} and assume that all pixel-dependent subfunctions can be accurately approximated by mixing some {\textit{basis functions}} with pixel-wise weights. Due to being shared by all subfunctions, these basis functions can be learned by DCNNs. While the pixel-wise mixing weights can be viewed as the pixel-wise channel attention~\cite{Sato2014Deep}, which also can be directly generated by a DCNN.

Following this idea, we construct the FM block with a separation-and-aggregation structure, as shown in Figure~\ref{fig:FM}. First, a convolutional block, \ie a convolutional layer followed by a Rectified Linear Unit (ReLu)~\cite{nair2010rectified}, is utilized for initial feature representation. Then, the obtained features are fed into multiple parallel subnets. Thereinto, one subnet is utilized to generate the pixel-wise mixing weights. For simplicity, we term it the {\textit{mixing function}}. While the remaining subnets represent the basis functions. Finally, the outputs of all basis functions are linearly mixed based on the generated pixel-wise weights. Let $\mathbf{x}^{u-1}$ denote the input for the $u$-th FM block $\mathcal{F}^{u}$ and $n$ denote the number of basis functions in $\mathcal{F}^{u}$. The output $\mathbf{x}^{u}$ of $\mathcal{F}^{u}$ can be formulated as
\begin{equation}\label{eq:eq1}
\begin{aligned}
&\mathbf{x}^{u}=\mathcal{F}^{u}(\mathbf{x}^{u-1})= \sum\nolimits^n_{i=1} f^{u}_{i}(\bar{\mathbf{x}}^{u},\theta^{u}_i)\odot w^{u}(\bar{\mathbf{x}}^{u},\vartheta^u)[i]\\
&{\rm{s.t.}}, {\kern 2pt} \bar{\mathbf{x}}^{u}=\mathcal{G}^u(\mathbf{x}^{u-1}, \omega^{u}),\\
&{\kern 14pt}\sum\nolimits^n_{i=1} w^{u}(\bar{\mathbf{x}}^{u}, \vartheta^u)[i] = \mathbf{1}, w^{u}(\bar{\mathbf{x}}^{u},\vartheta^u)\succeq 0,
\end{aligned}
\end{equation} 
where $f^{u}_i(\cdot,\theta^u_i)$ denotes the $i$-th basis function parameterized by $\theta^u_i$ and $w^{u}(\cdot, \vartheta^u)$ represents the mixing function parameterized by $\vartheta^u$. When $f^{u}_{i}(\bar{\mathbf{x}}^{u},\theta^{u}_i)$ is of size $c\times h\times w$ (\ie, channel $\times$ height $\times$ width), $w^{u}(\bar{\mathbf{x}}^{u}, \vartheta^u)$ is of size $n\times h \times w$, and $w^{u}(\bar{\mathbf{x}}^{u}, \vartheta^u)[i]$ represents the mixing weights of size $h\times w$ generated for all pixels corresponding to the $i$-th basis function. $\odot$ denotes the point product. $\bar{\mathbf{x}}^u$ denotes the features output by the convolutional block $\mathcal{G}^u(\cdot,\omega^{u})$ in $\mathcal{F}^{u}$, and $\omega^{u}$ represents the convolutional filters. Inspired by~\cite{everitt2005finite}, we also require the mixing weights to be non-negative and the summation across all basis functions is equal to 1, as shown in Eq.~\eqref{eq:eq1}.

\begin{figure}
\centering
\includegraphics[height=2.2in, width=3.5in]{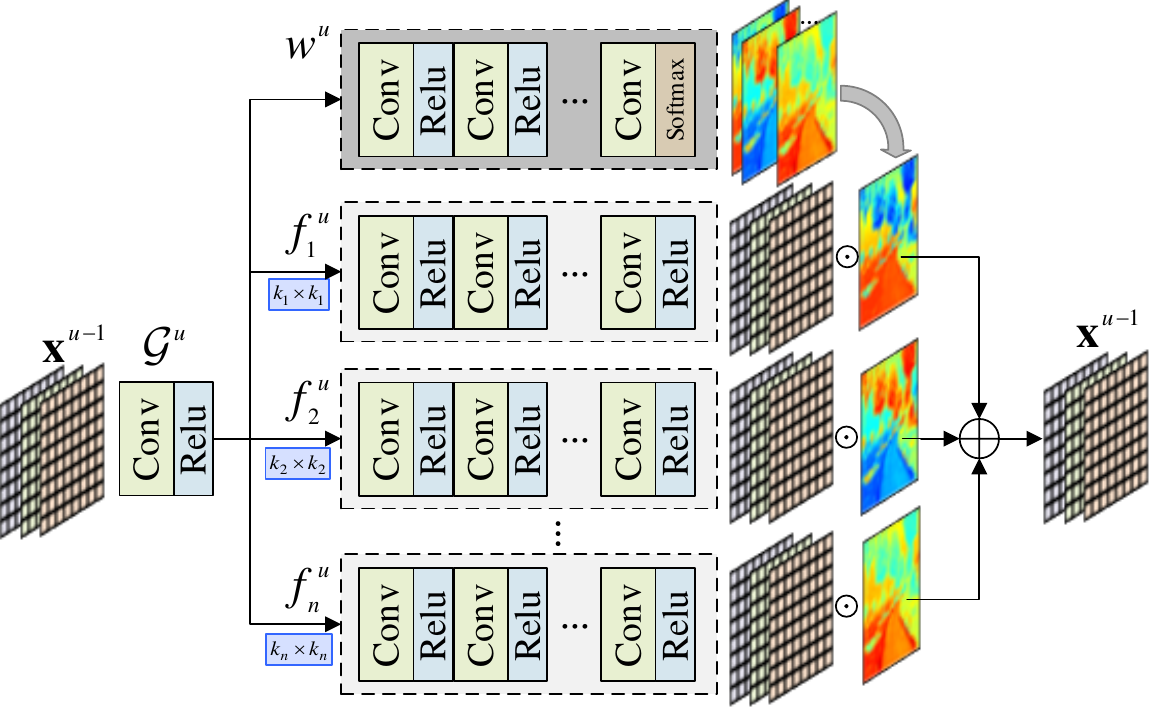}
\caption{Architecture of the proposed function-mixture block where $k_i$ ($i=1,\cdots,n$) denotes the convoluational filter size in the $i$-th basis function $f^u_i$.}
\label{fig:FM}
\vspace{-0.3cm}
\end{figure}

In this study, we implement the basis functions as well as the mixing function by stacking $m$ consecutive convolutional blocks, as shown in Figure~\ref{fig:FM}. Moreover, we equip these basis functions with different-sized convolutional filters to ensure they take different-sized receptive fields and learn distinct mapping schemes. For the mixing function, we introduce a Softmax unit at the end to comply with the constraints in Eq.~\eqref{eq:eq1}. Apparently, profiting from such a pixel-wise mixture architecture, the proposed FM block is able to determine the appropriate receptive field size and the mapping function for each pixel.

\subsection{Multiple FM Blocks}\label{subsec:MFMB}
As shown in Figure~\ref{fig:FMNet}, in the proposed network, we first introduce an individual convolutional block, and then stack multiple FM blocks for the intermediate feature representation and the ultimate output. For an input RGB image $\mathbf{x}$, the output of the network with $p$ FM blocks can be given as 
\begin{equation}\label{eq:eq2}
\begin{aligned}
&\mathbf{y} = \mathbf{x} + \mathcal{F}^{p}\left(\mathcal{F}^{p-1}\left(\cdots\mathcal{F}^2\left(\mathcal{F}^{1}\left(\mathbf{x}^0\right)\right)\right)\right),\\
&{\rm{s.t.}}, \quad\mathbf{x}^0 = \mathcal{G}^{0}\left(\mathbf{x},\omega^{0}\right),
\end{aligned}
\end{equation}
where $\mathbf{y}$ denotes the generated HSI and $\mathbf{x}^0$ represents the output of the first convolutional block $\mathcal{G}^{0}(\cdot,\omega^{0})$ parameterized by $\omega^{0}$. It is worth noting that in this study we increase the spectral resolution of $\mathbf{x}$ to the same as that of $\mathbf{y}$ by the bilinear interpolation. In addition, $\mathcal{F}^{1}, \cdots,\mathcal{F}^{p-1}$ show the same architecture, while the output of $\mathcal{F}^{p}$ will be adjusted according to the number of spectral bands in $\mathbf{y}$.

It has been shown that the layers in an DCNN from bottom to top take increasingly larger receptive fields and extract different levels of features from the input signal~\cite{zhou2016learning}. Therefore, by stacking multiple FM blocks, we can further increase the flexibility of the proposed network in learning the pixel-wise mapping, viz., adjust the receptive field size and the mapping function for each pixel at multiple levels. In addition, considering that each FM block defines the mapping subfunction for each pixel, the ultimate mapping function obtained by stacking $p$ FM blocks can be viewed as a composition function of $p$ subfunctions. Since each subfunction is approximated by the mixture of $n$ basis functions, the ultimate mapping function can be viewed as the mixture of $n^p$ basis functions, which show much larger expressive capacity than a single FM block in pixel-wisely fitting an appropriate mapping function. 

\subsection{Intermediate Features Fusion}\label{subsec:IFF}
As previously mentioned, the FM blocks in the porposed network extract different levels of features from the input. Inspired by~\cite{kim2016deeply,zhang2018residual}, to reuse these intermediate features for performance enhancement, we employ skip connections to aggregate the intermediate features generated by each FM block before the ultimate output block with a concatenation operation, as shown in Figure~\ref{fig:FMNet}. To better utilize all of these features for pixel-wise representation, we introduce an extra FM block $\mathcal{F}_{c}$ to fuse the concatenation result. With such an intermediate feature fusion operation, the output of the proposed network can be reformulated as
\begin{equation}\label{eq:eq3}
\begin{aligned}
\mathbf{y} = \mathbf{x} + \mathcal{F}^{p}\left(\mathcal{F}_c\left(\left[\mathcal{F}^{p-1}\left(\cdots\mathcal{F}^{1}\left(\mathbf{x}^{0}\right)\right),\cdots,\mathcal{F}^{1}\left(\mathbf{x}^{0}\right)\right]\right)\right)
\end{aligned}
\end{equation}

\begin{table*}\small
\caption{Numerical results of different methods on three benchmark SSR datasets. The best results are in bold.}
\label{table:numerical}
\renewcommand{\arraystretch}{1.1}
\begin{center}
\begin{tabular}{l|c|c|c|c|c|c|c|c|c|c|c|c}
\hline
\multirow{2}{*}{Methods} & \multicolumn{4}{c|}{NTIRE2018} & \multicolumn{4}{c|}{CAVE} & \multicolumn{4}{c}{Harvard}\\
\cline{2-13}
& {RMSE} & {PSNR} & {SAM} & {SSIM} & {RMSE} & {PSNR} & {SAM} & {SSIM} & {RMSE} & {PSNR} & {SAM} & {SSIM}\\
\hline
BI~\cite{hou1978cubic} & 15.41 & 25.73 & 15.30 & 0.8397 & 26.60 & 21.49 & 34.38 & 0.7382 & 30.86 & 19.44 & 39.04 & 0.5887\\
Arad~\cite{arad2016sparse} & 4.46 & 35.63 & 5.90 & 0.9082 & 10.09 & 28.96 & 19.54 & 0.8695 & 7.85 & 31.30 & 8.32 & 0.8490\\
Aitor~\cite{alvarez2017adversarial} & 1.97 & 43.30 & 1.80 & 0.9907 & 6.80 & 32.53 & 17.50 & 0.8768 & 3.29 & 39.21 & 4.93 & 0.9671\\
HSCNN+~\cite{xiong2017hscnn} & 1.55 & 45.38 & 1.63 & 0.9931 & 4.97 & 35.66 & 8.73 & 0.9529 & 2.87 & 41.05 & 4.28 & 0.9741\\
\hline
DCNN & 1.23 & 47.40 & 1.30 & 0.9939 & 5.77 & 34.09 & 11.35 & 0.9275 & 2.88 & 40.83 & 4.24 & 0.9724\\
MCNet & 1.11 & 48.43 & 1.13 & 0.9951 & 4.84 & 35.92 & 8.98 & 0.9555 & 2.83 & 40.70 & 4.26 & 0.9689\\
\hline
Ours & {\textbf{1.03}} & {\textbf{49.29}} & {\textbf{1.05}} & {\textbf{0.9955}} & {\textbf{4.54}} & {\textbf{36.33}} & {\textbf{7.07}} & {\textbf{0.9611}} & {\textbf{2.54}} & {\textbf{41.54}} & {\textbf{3.76}} & {\textbf{0.9796}}\\
\hline
\end{tabular}
\end{center}
\vspace{-0.5cm}
\end{table*}

\section{Experiment}
In this section, we will conduct extensive comparison experiments and carry out an ablation study to demonstrate the effectiveness of the proposed method in SSR.

\subsection{Experimental Setting}

\paragraph{Datasets}
In this study, we adopt three benchmark HSI datasets, including NTIRE2018~\cite{timofte2018ntire}, CAVE~\cite{yasuma2010generalized} and Harvard~\cite{chakrabarti2011statistics}. NTIRE2018 dataset is the benchmark for the SSR challenge in NTIRE2018. 
In NTIRE2018 dataset, there are 255 paired HSIs and RGB images which have the same spatial resolution, \eg, 1392 $\times$ 1300. Each HSI consists of $31$ successive spectral bands ranging from 400$nm$ to 700$nm$ with a 10$nm$ interval. CAVE dataset contains 32 HSIs of indoor objects. Similar to NTIRE2018, each HSI contains $31$ spectral bands ranging from 400$nm$ to 700$nm$ with a 10$nm$ interval but with the spatial resolution 512 $\times$ 512. Harvard dataset is another common benchmark for HSIs. It consists of 50 HSIs with spatial resolution 1392$\times$1040. Each image contains 31 spectral bands captured from 420$nm$ to 720$nm$ with a 10$nm$ interval. For the CAVE and Havard datasets, inspird by~\cite{dong2016hyperspectral,zhang2018exploiting}, we adopt the spectral response function of Nikon D700 camera~\cite{dong2016hyperspectral} to generate the corresponding RGB image for each HSI. In the following experiments, we randomly select 200 paired images from the NTIRE2018 dataset as the training set and the remaining 55 paired images for testing. For the CAVE dataset, we randomly choose 22 paired images for training and the remaining 10 paired images for testing. While in the Harvard dataset, 30 paired images are randomly chosen as the training set and the remaining 20 paired images are utilized for testing.
\vspace{-0.4cm}
\paragraph{Comparison Methods} In this study, we compare the proposed method with 6 existing methods including the bilinear interpolation (BI)~\cite{hou1978cubic}, Arad~\cite{arad2016sparse}, Aitor~\cite{alvarez2017adversarial}, HSCNN+~\cite{xiong2017hscnn}, deep convolution neural network (DCNN) and the multi-column network (MCNet). Among them, the BI utilizes the bilinear interpolation to increase the spectral resolution of the input RGB image. The Arad is a sparsity induced conventional SSR method. The Aitor and HSCNN+ are two recent DCNN based state-of-the-art SSR methods. The DCNN and MCNet are two baselines for the proposed method. The DCNN is a variant of the proposed method that is implemented by replacing each FM block in the proposed method with a convolutional block. For the MCNet, we implement it following the basic architecture in~\cite{cirecsan2012multi,zhang2016single} with the convolutional blocks. Moreover, the column number is set as $n$ and the convolutions in $n$ columns are equipped with $n$ kinds of different-sized filters, which is similar as the proposed method. We further control the depth of each column to make sure the model complexity of the MCNet is comparable to the proposed method. By doing this, the only difference between the MCNet and the proposed network is the network architecture. For fair comparison, all these DCNN based competitors and the spectral dictionary in the Arad~\cite{arad2016sparse} are retrained on the training set utilized in the experiments.
\vspace{-0.4cm}
\paragraph{Evaluation Metrics} To objectively evaluate the SSR performance of each method, we employ four commonly utilized metrics, including the root-mean-square error (RMSE), peak signal-to-noise ratio (PSNR), spectral angle sapper (SAM) and structural similarity index (SSIM). The RMSE and PSNR measure the numerical difference between the reconstructed image and the reference image. The SAM computes the average spectral angle between two spectra from the reconstructed image and the reference image at the same spatial position to indicate the reconstruction accuracy of spectrum. The SSIM is often utilized to measure the spatial structure similarity between two images. In general, a larger PSNR or SSIM and a smaller RMSE or SAM indicate better performance. 
\vspace{-0.4cm}
\paragraph{Implementation Details}
In the proposed method, we adopt 4 FM blocks (\ie, including $\mathcal{F}_c$ for feature fusion and $p$=3), and each block contains $n=3$ basis functions. The basis functions and the mixing functions consist of $m$=2 convolutional blocks. Each convolutional block contains 64 filters. In each FM block, three basis functions are equipped with three different-sized filters for convolution, \ie, 3$\times$3, 7$\times$7 and 11$\times$11. While the filter size in all other convolutional blocks is fixed as 3$\times$3.

In this study, we implement the proposed method on the Pytorch platform~\cite{ketkar2017introduction} and train the network using the following model
\begin{equation}\label{eq:eq4}
\begin{aligned}
\min\limits_{\theta} \frac{1}{N}\sum\nolimits^N_{i=1} \|\mathbf{y}_i - f(\mathbf{x}_i,\theta)\|_1,
\end{aligned}
\end{equation}
where $\{(\mathbf{y}_i, \mathbf{x}_i)\}$ denotes the $i$-th paired HSI and RGB image, respectively. $N$ denotes the number of training pairs. $f$ denotes the ultimate mapping function defined by the proposed network and $\theta$ represents all involved parameters. $\|\cdot\|_1$ represents the $\ell_1$ norm based loss. In the training stage, we employ the Adam optimizer~\cite{kingma2014adam} with the weight decay 1e-6. The learning rate is initially set as 1e-4 and halved in every 20 epochs. The batch size is 128. We terminate the optimization at the $100$-th epoch.

\begin{figure*}[htbp]
\centering
\includegraphics[height=1.0in, width=0.9in]{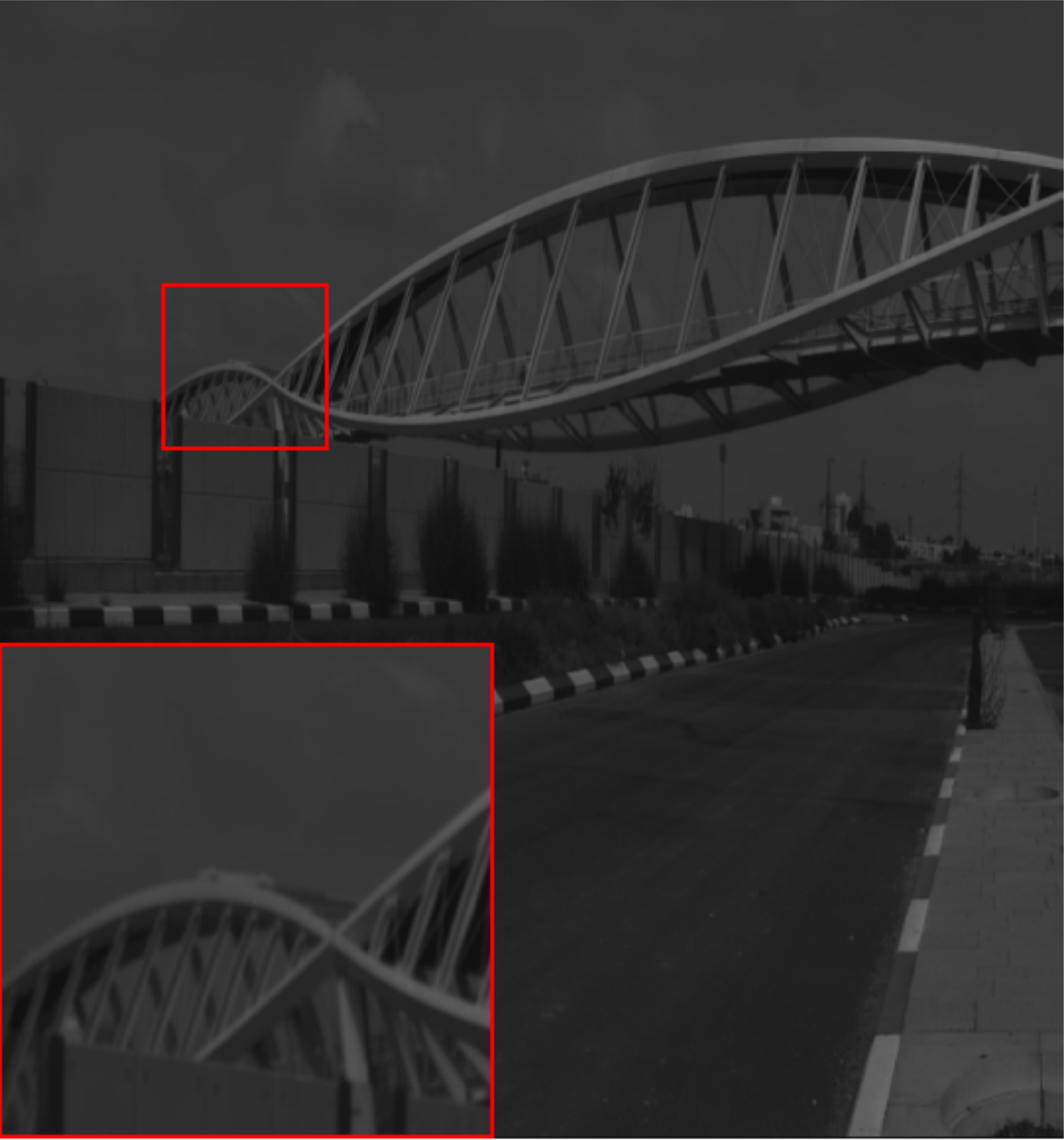}
\hspace{-0.15cm}
\includegraphics[height=1.0in, width=0.9in]{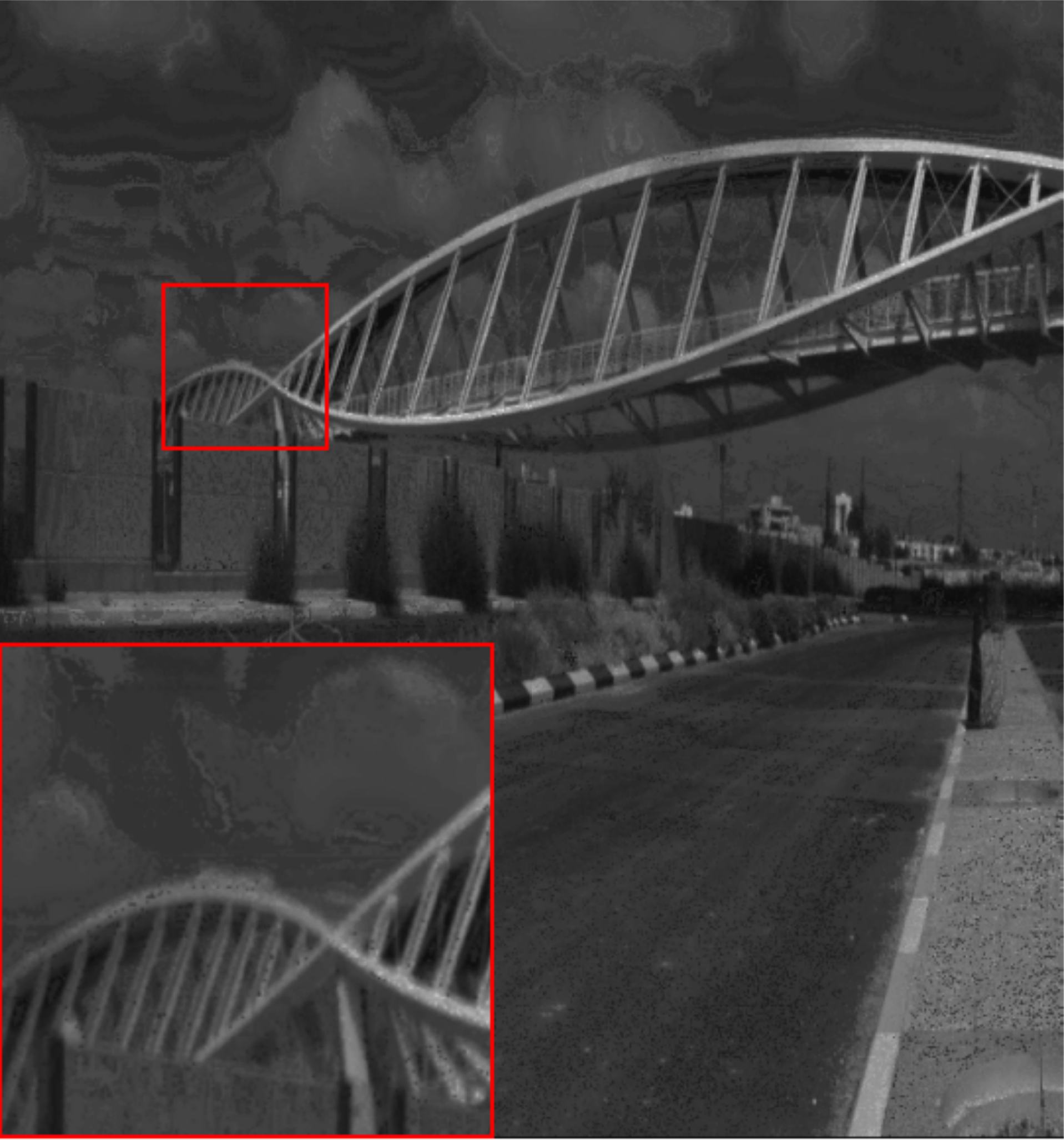}
\hspace{-0.15cm}
\includegraphics[height=1.0in, width=0.9in]{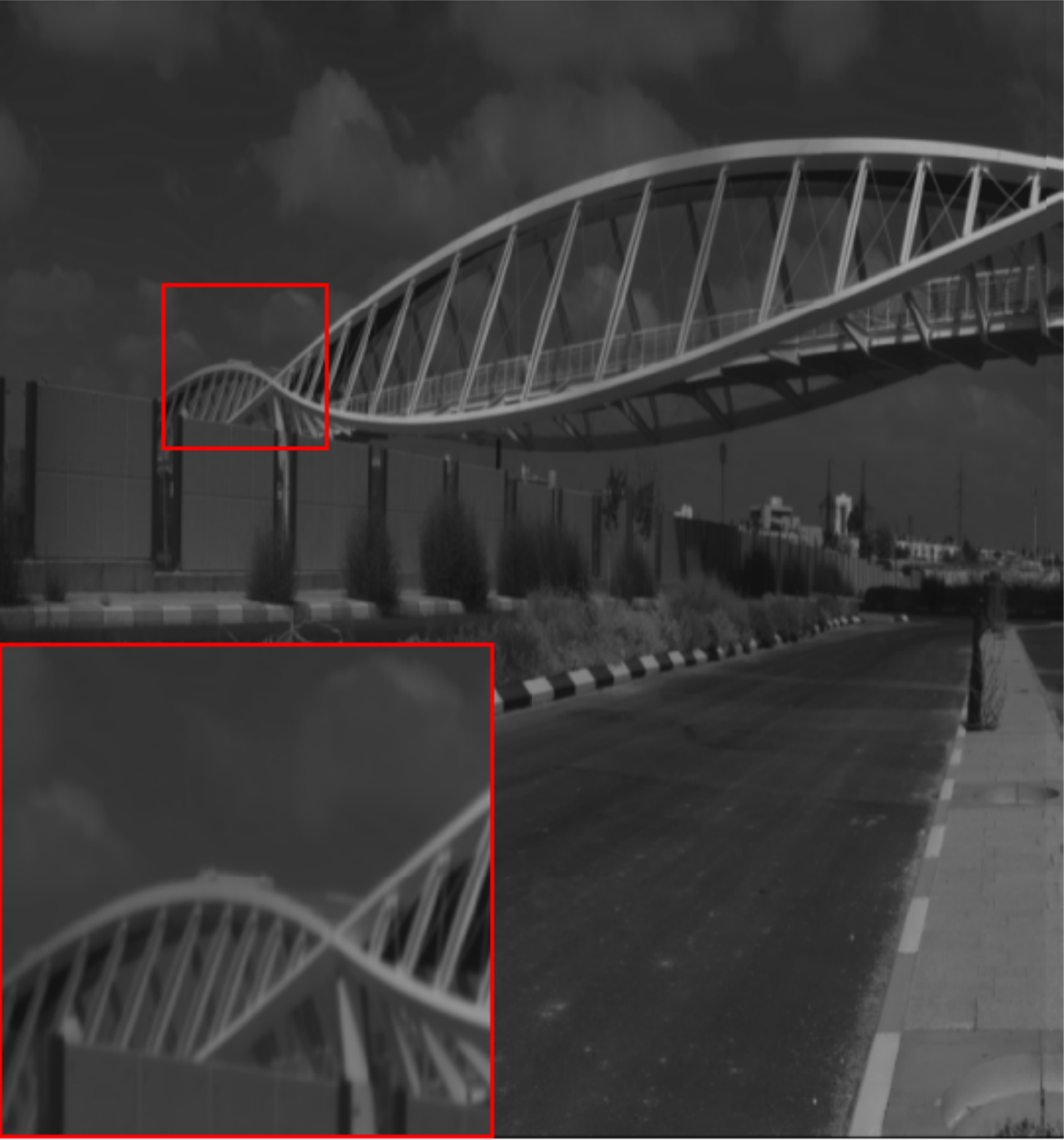}
\hspace{-0.15cm}
\includegraphics[height=1.0in, width=0.9in]{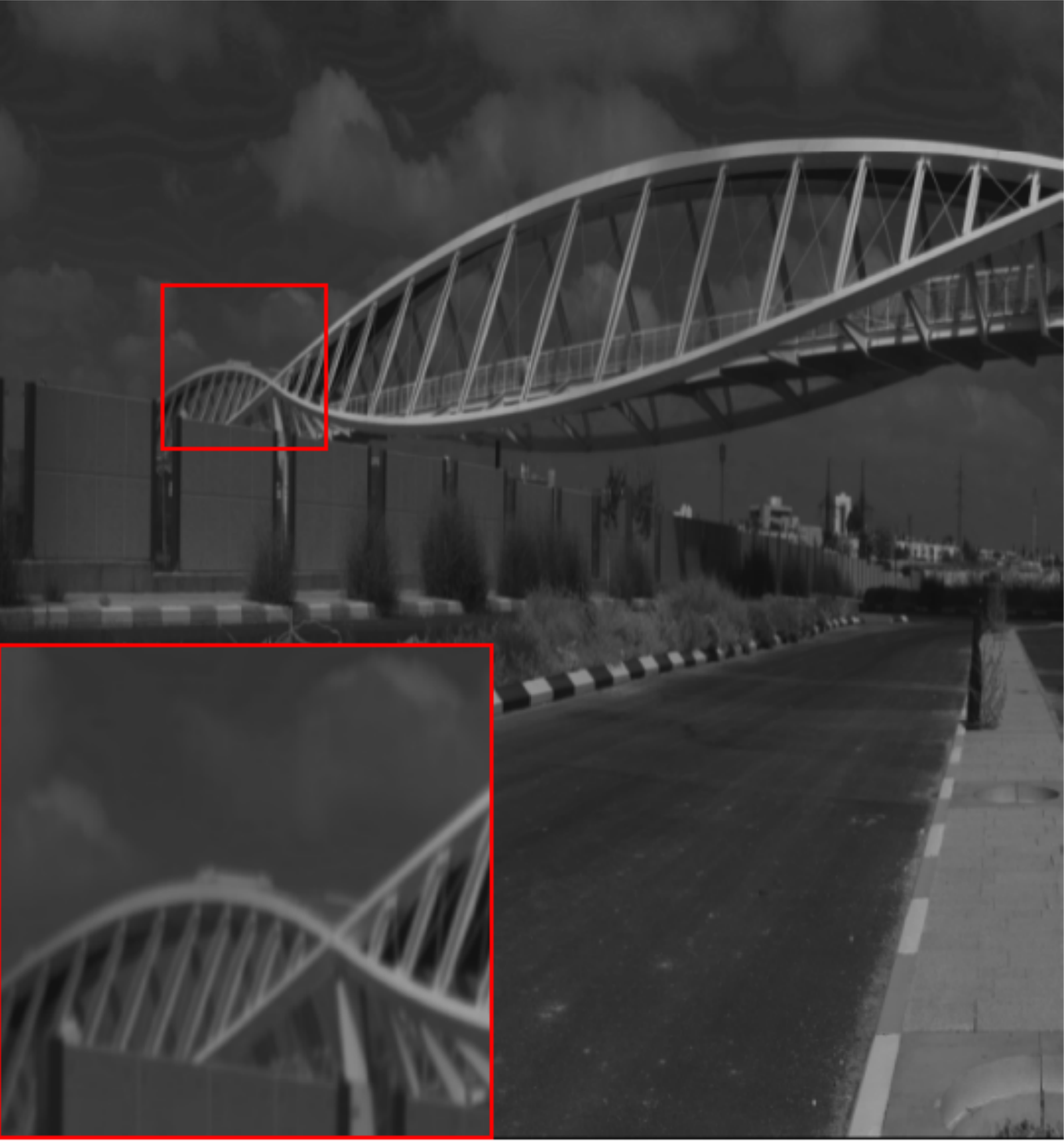}
\hspace{-0.15cm}
\includegraphics[height=1.0in, width=0.9in]{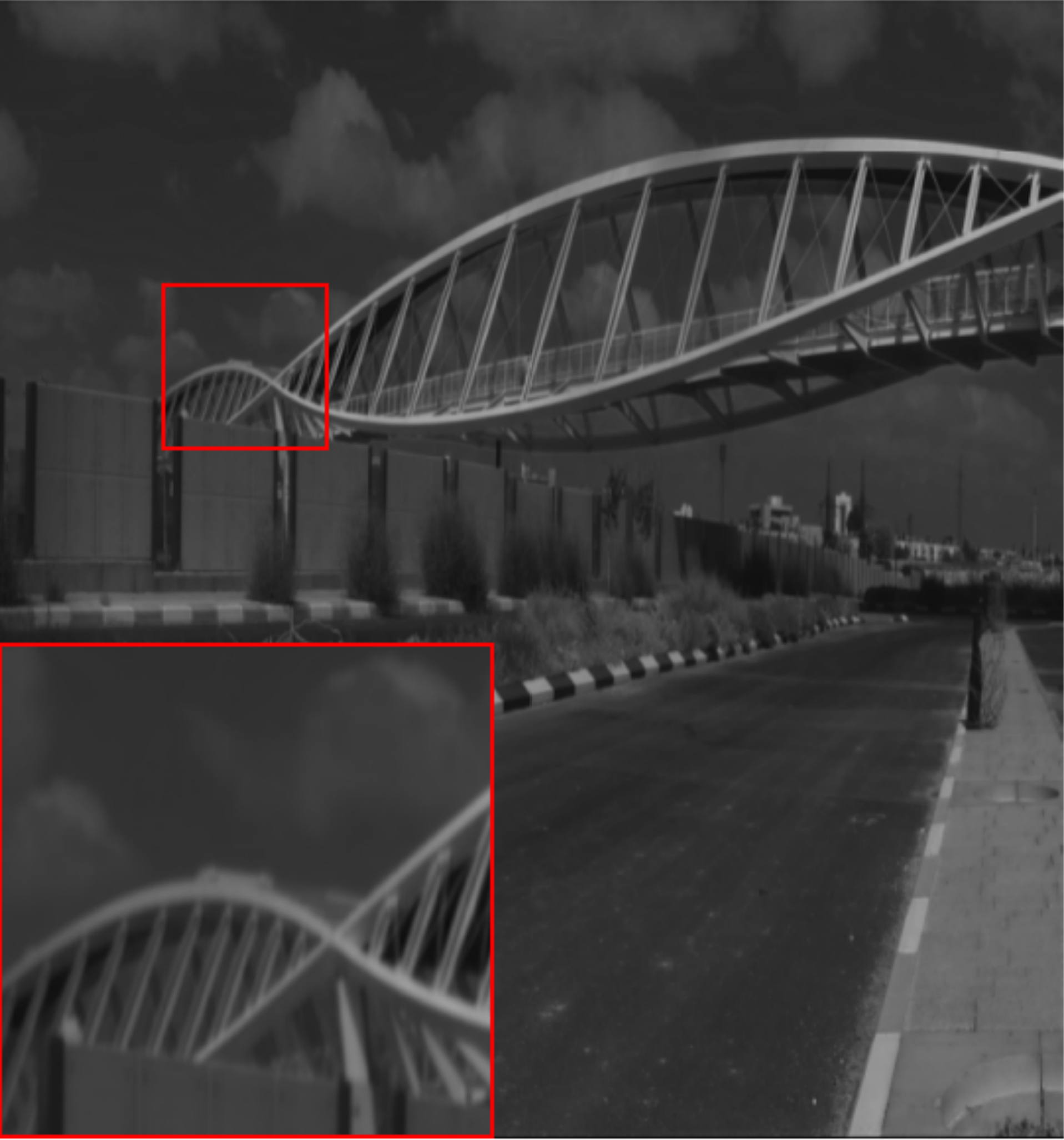}
\hspace{-0.15cm}
\includegraphics[height=1.0in, width=0.9in]{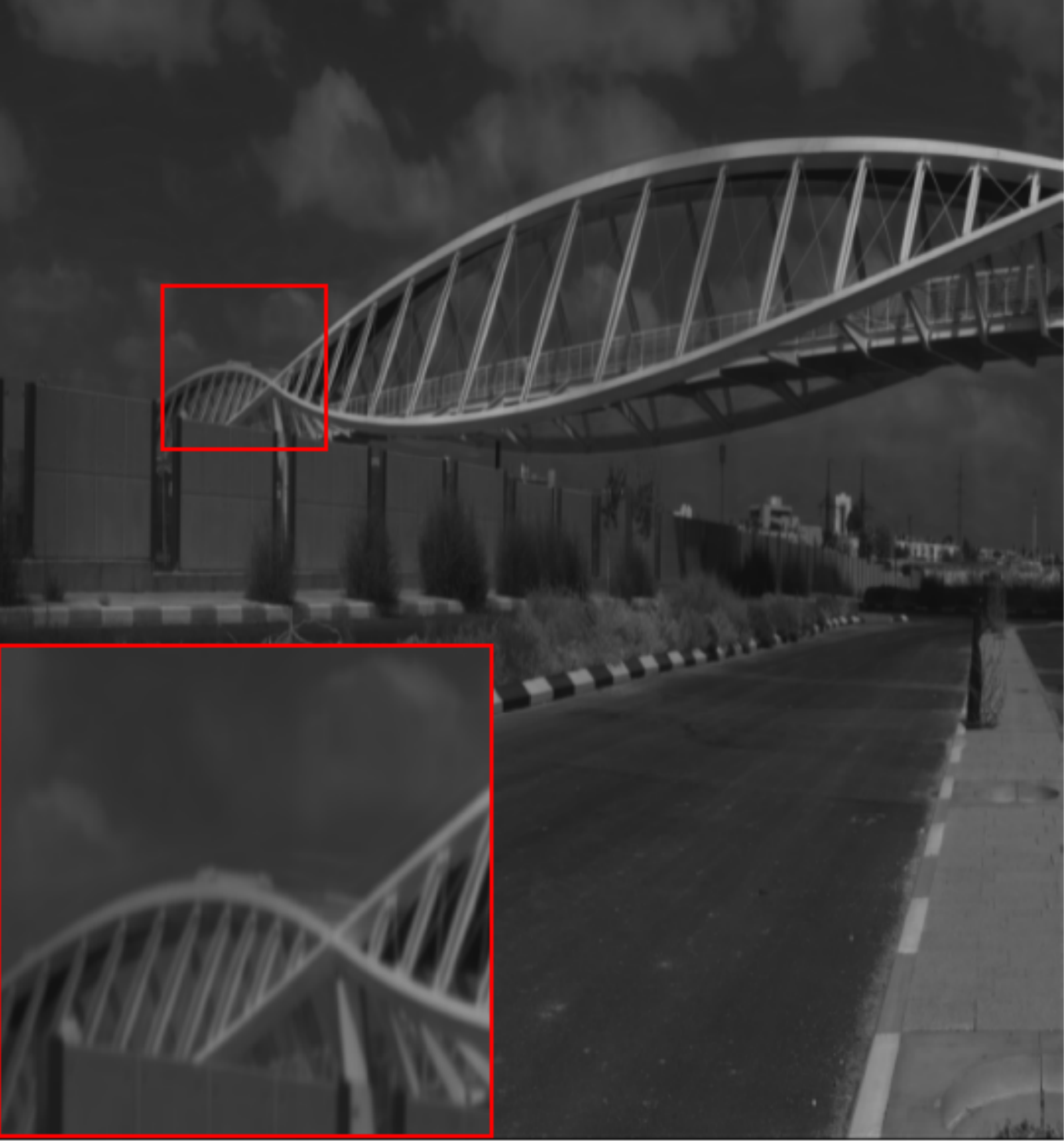}
\hspace{-0.15cm}
\includegraphics[height=1.0in, width=0.9in]{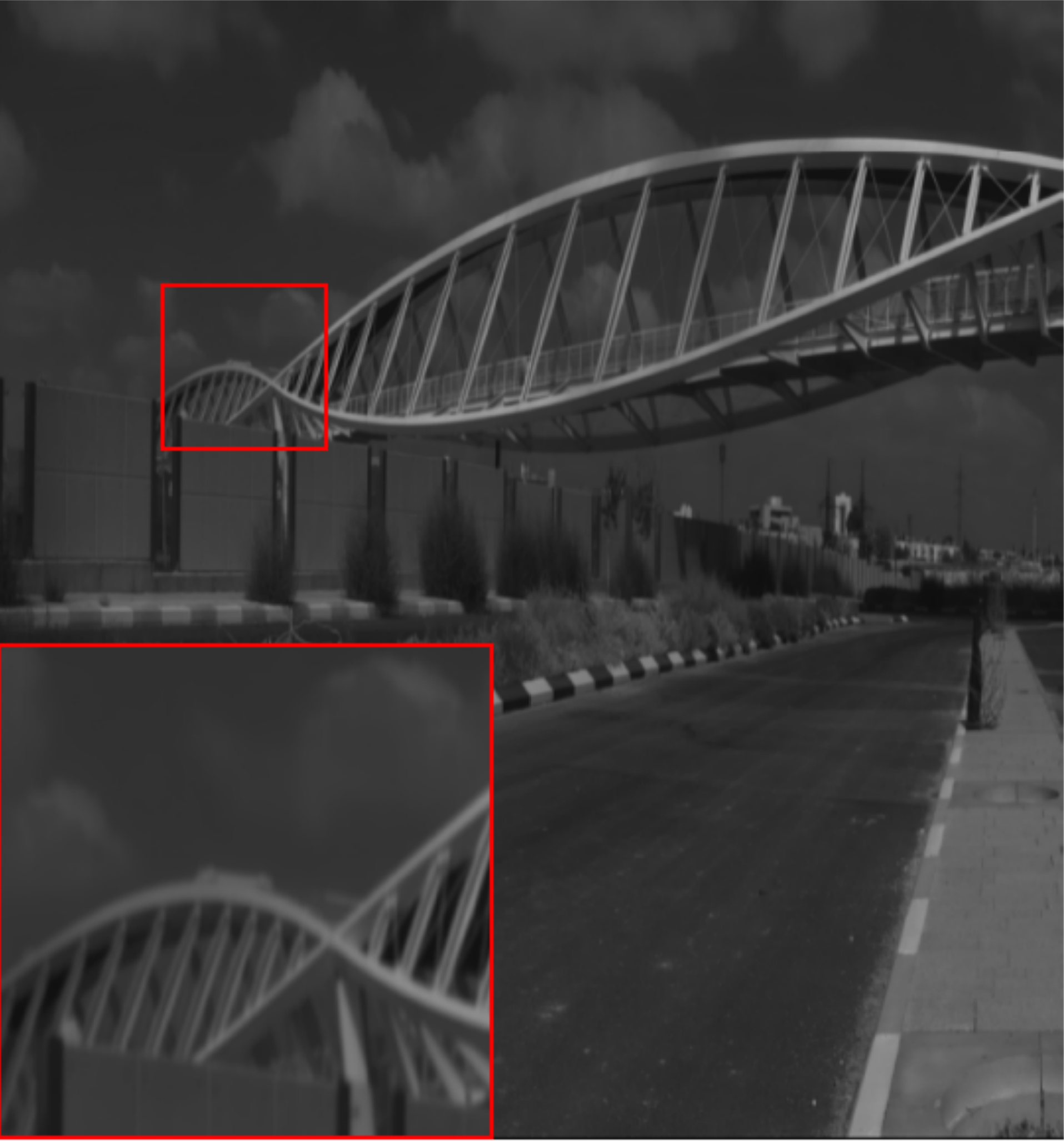}
\hspace{-0.15cm}
\includegraphics[height=1.0in, width=0.2in]{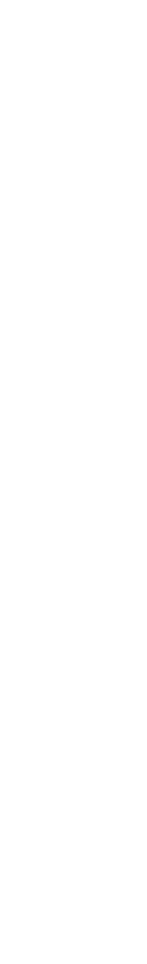}
\\
\vspace{-0.18cm}
\subfigure[BI~\cite{hou1978cubic}]{\includegraphics[height=1.0in, width=0.9in]{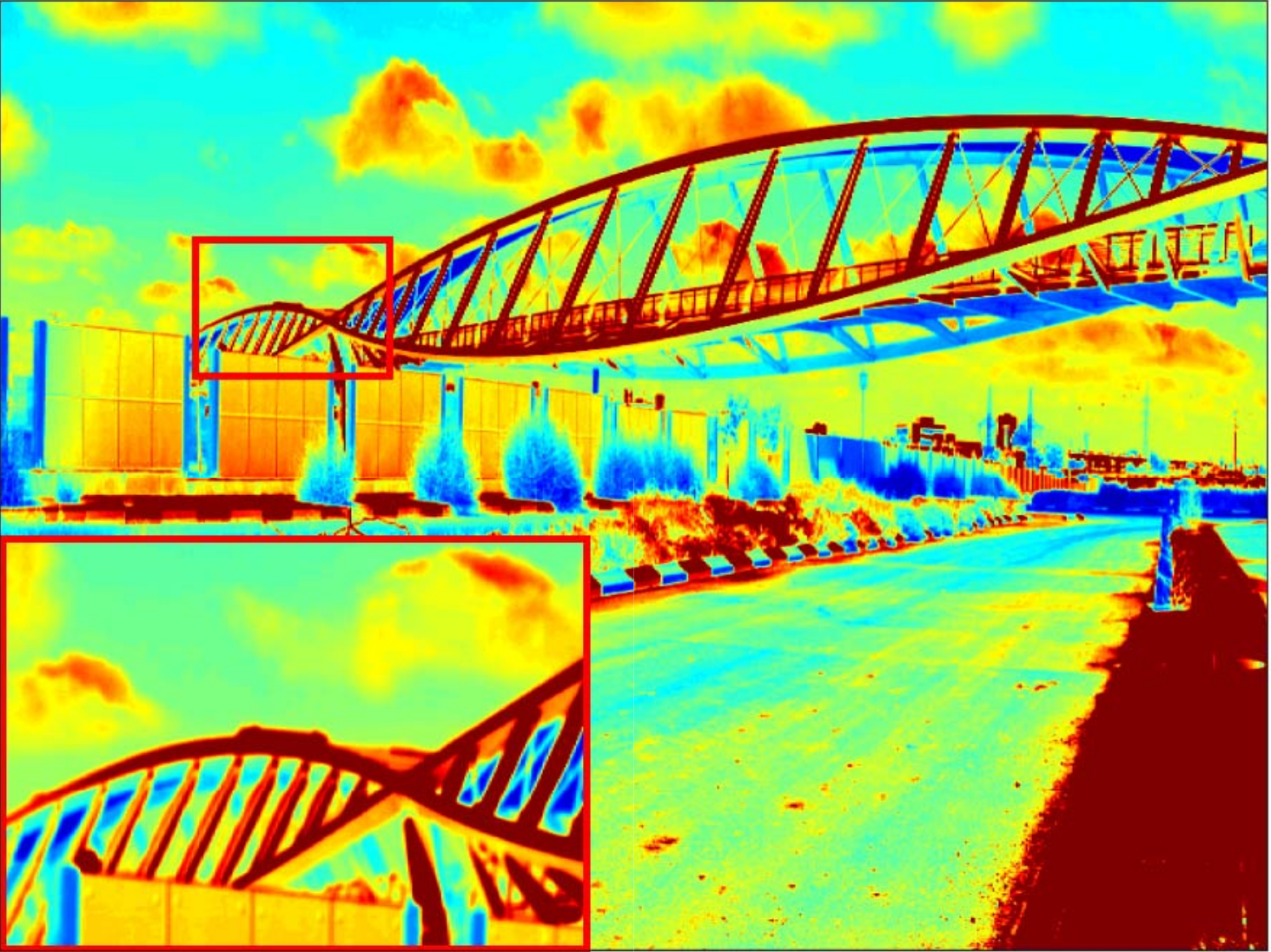}}
\hspace{-0.15cm}
\subfigure[Arad~\cite{arad2016sparse}]{\includegraphics[height=1.0in, width=0.9in]{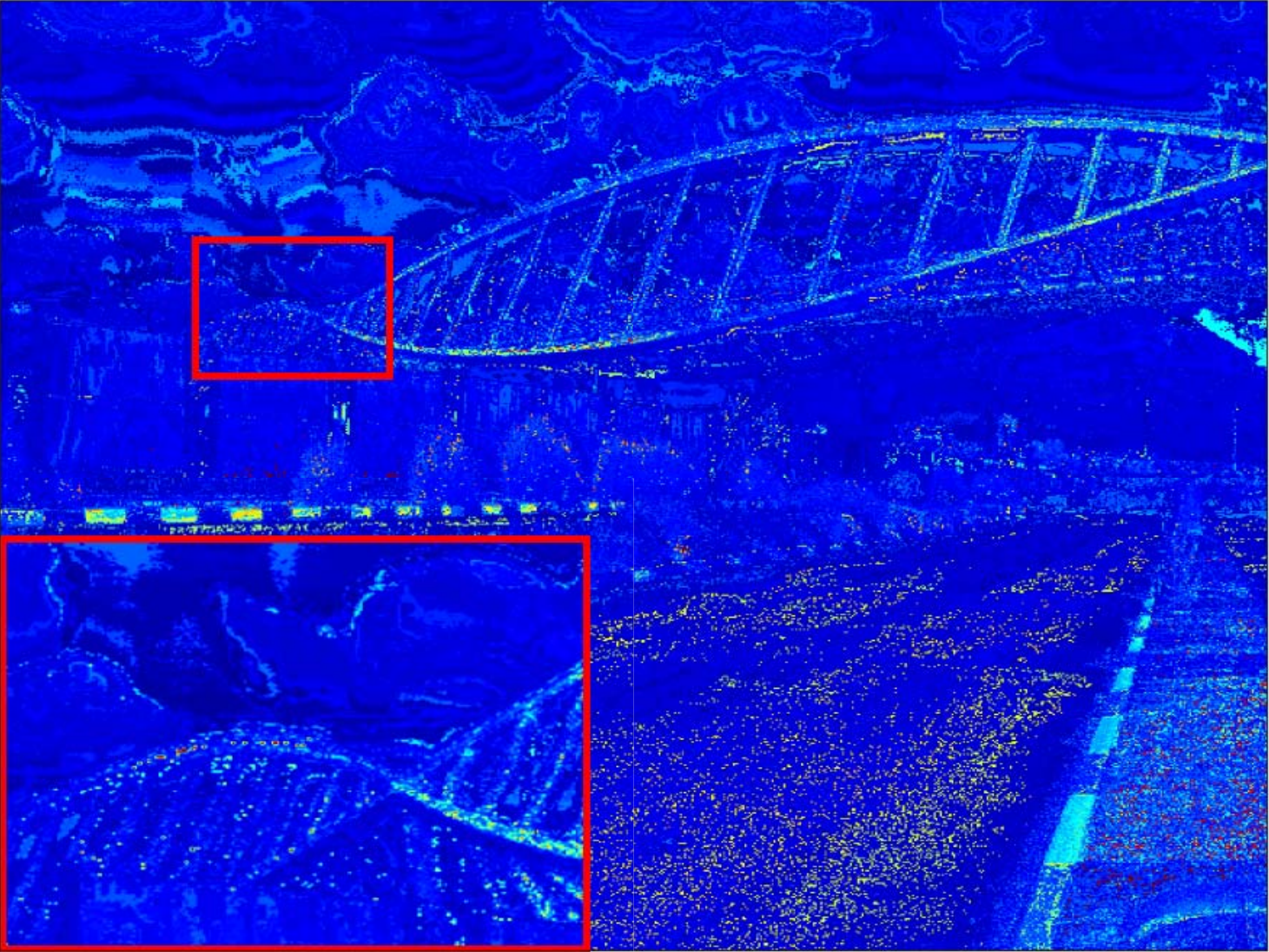}}
\hspace{-0.15cm}
\subfigure[Aitor~\cite{alvarez2017adversarial}]{\includegraphics[height=1.0in, width=0.9in]{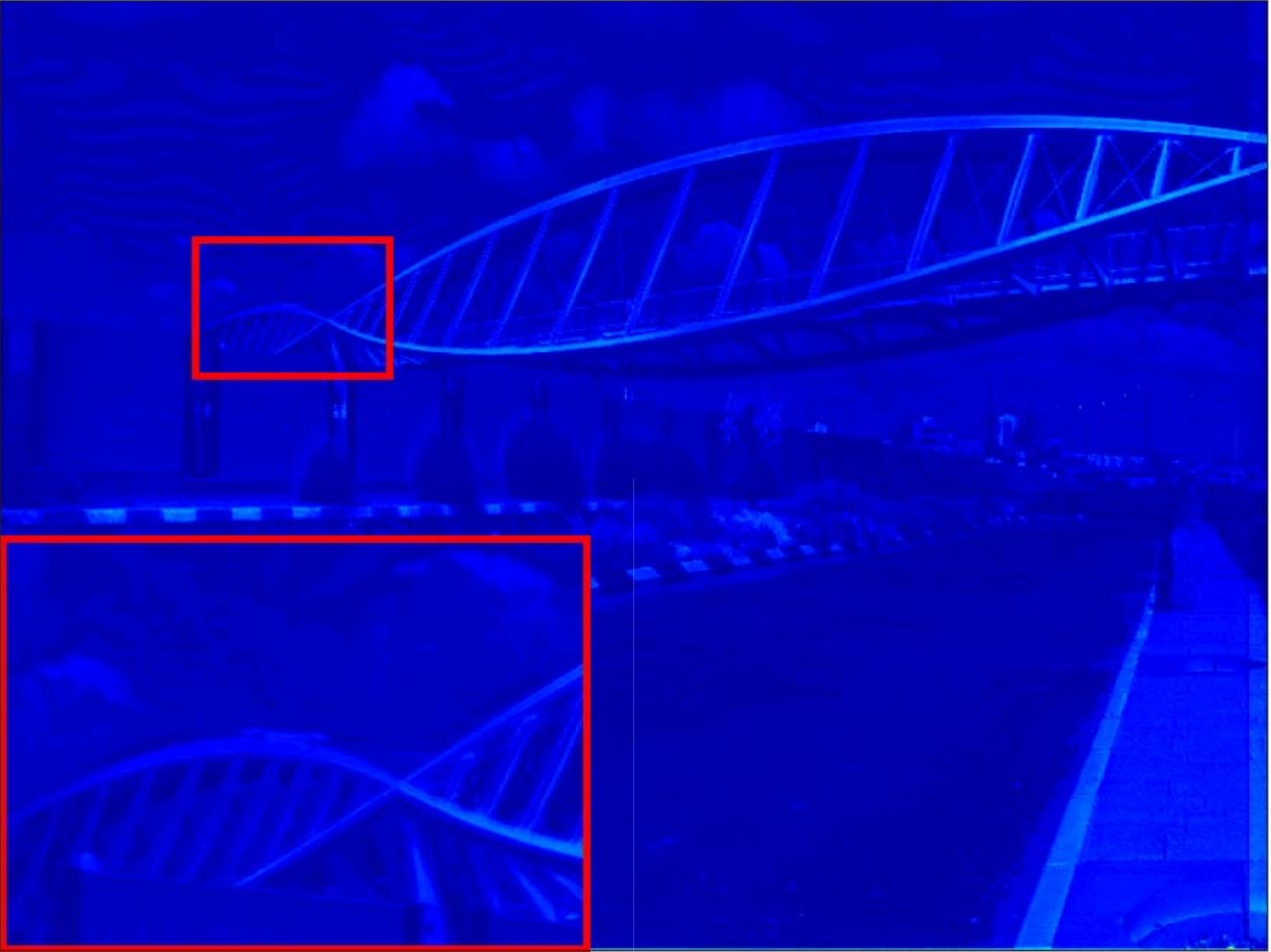}}
\hspace{-0.15cm}
\subfigure[HSCNN+~\cite{xiong2017hscnn}]{\includegraphics[height=1.0in, width=0.9in]{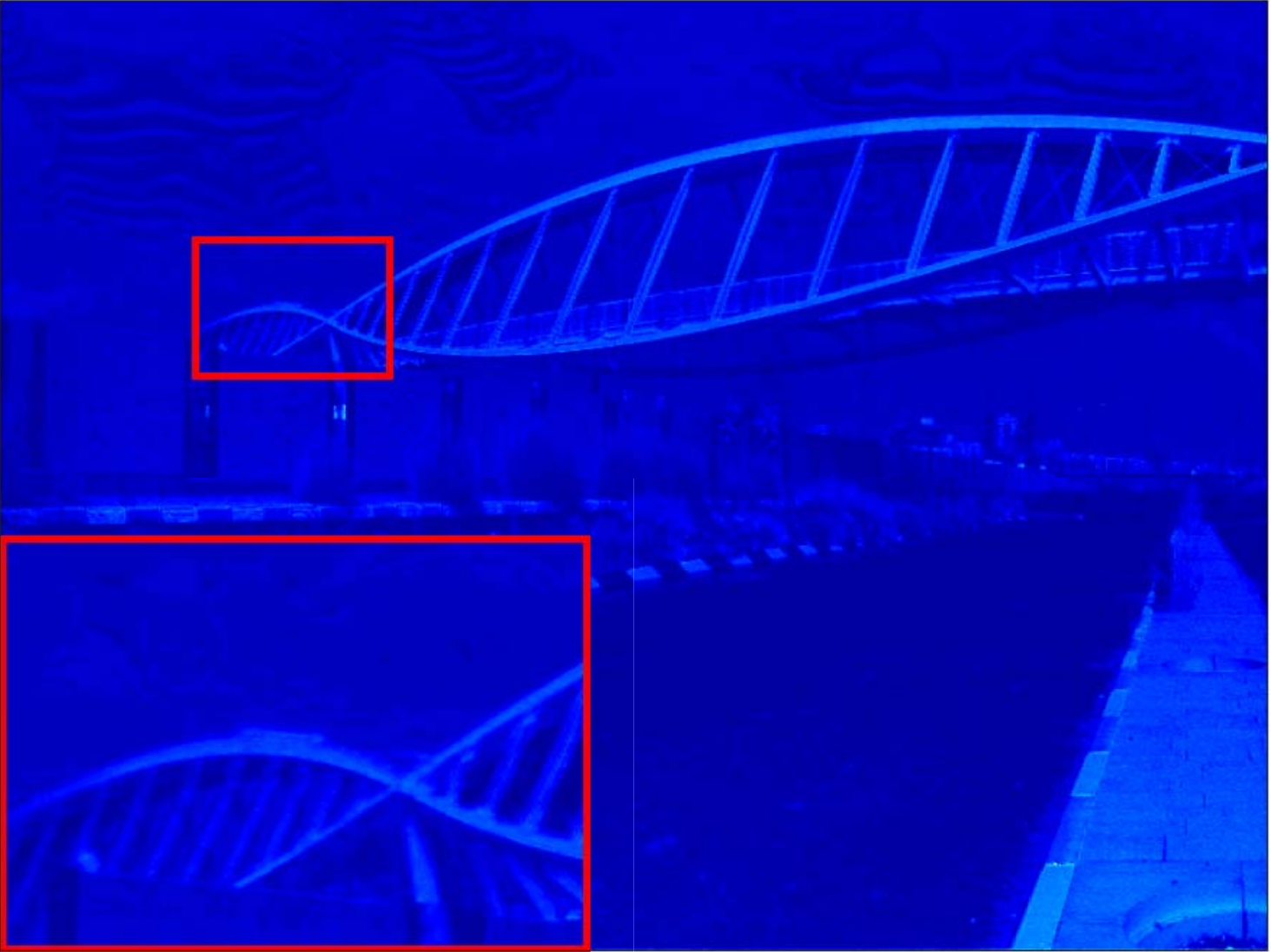}}
\hspace{-0.15cm}
\subfigure[DCNN]{\includegraphics[height=1.0in, width=0.9in]{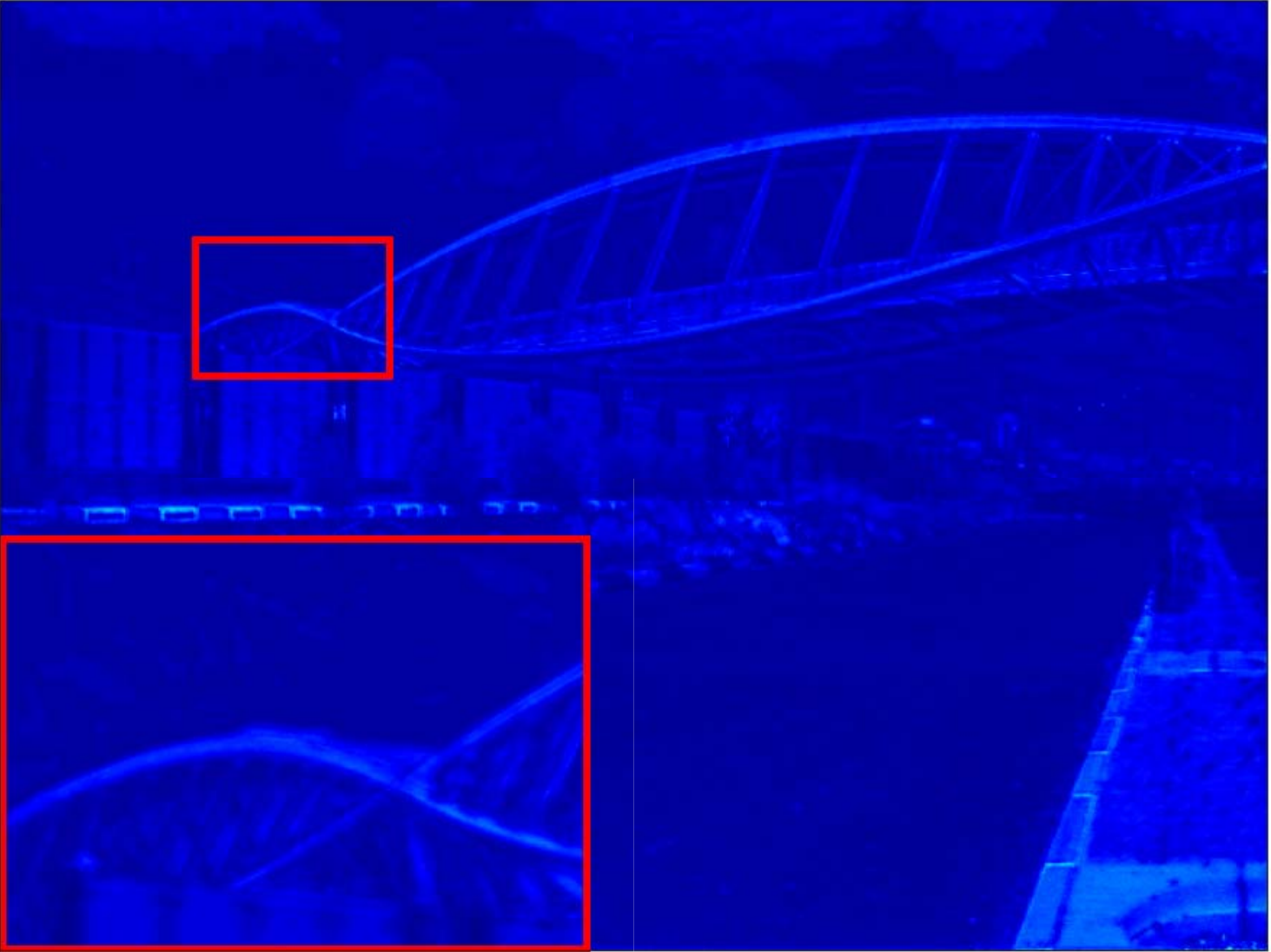}}
\hspace{-0.15cm}
\subfigure[MCNet]{\includegraphics[height=1.0in, width=0.9in]{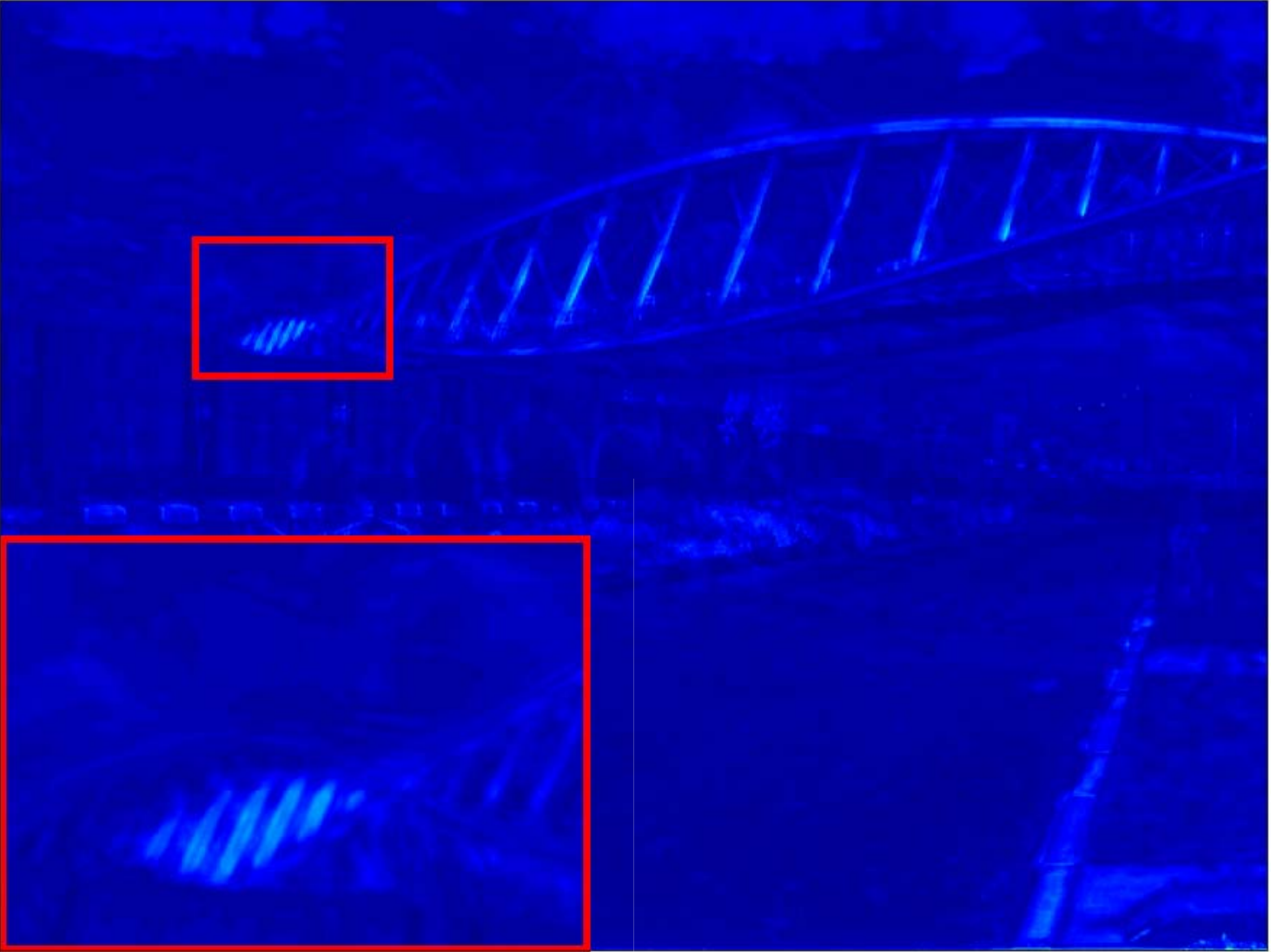}}
\hspace{-0.15cm}
\subfigure[Ours]{\includegraphics[height=1.0in, width=0.9in]{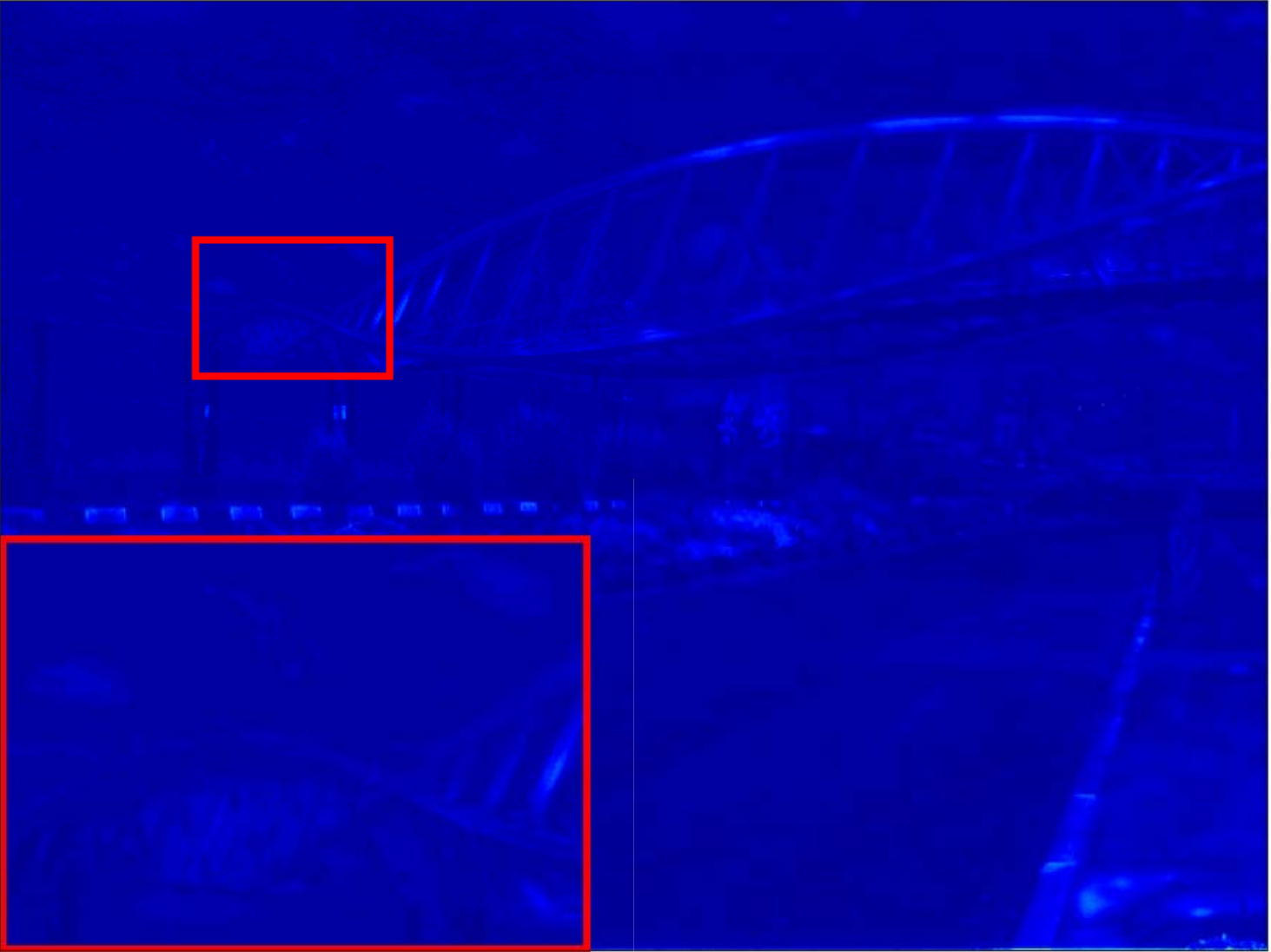}}
\hspace{-0.15cm}
\includegraphics[height=1.0in, width=0.2in]{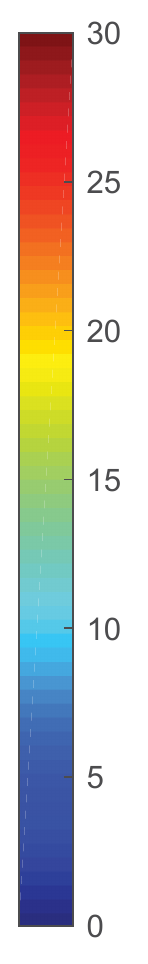}
\caption{Visual super-resolution results of the 31-th band and the reconstruction error maps of an example image from the NTIRE2018 dataset for different methods. The reconstruction error is obtained by computing the mean-square error between two spectrum vectors from the super-resolution result and the ground truth at each pixel. Best view on the screen.}
\label{fig:visual-ntr}
\vspace{-0.3cm}
\end{figure*}

\begin{figure*}[htbp]
\centering
\includegraphics[height=1.0in, width=0.9in]{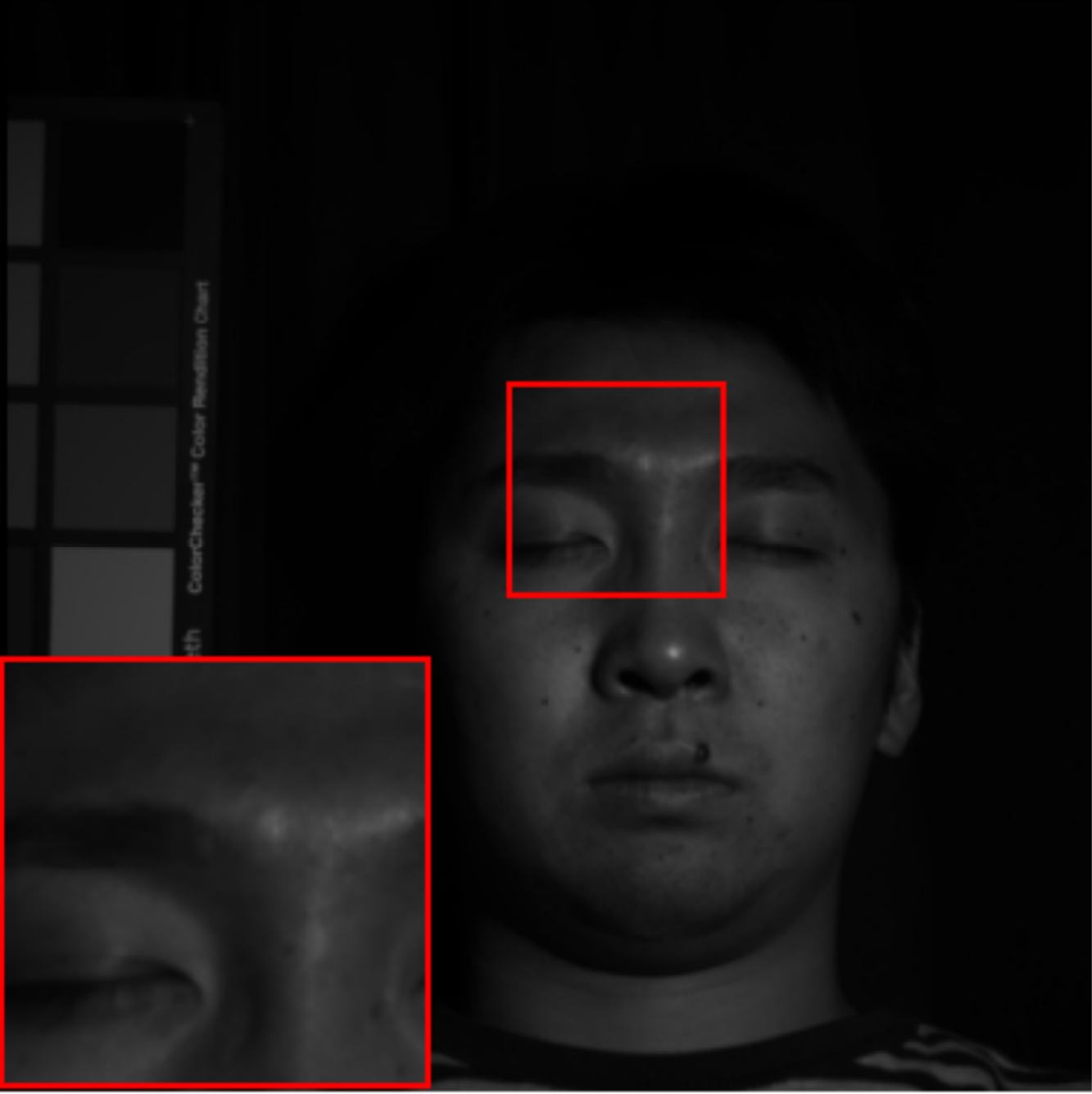}
\hspace{-0.15cm}
\includegraphics[height=1.0in, width=0.9in]{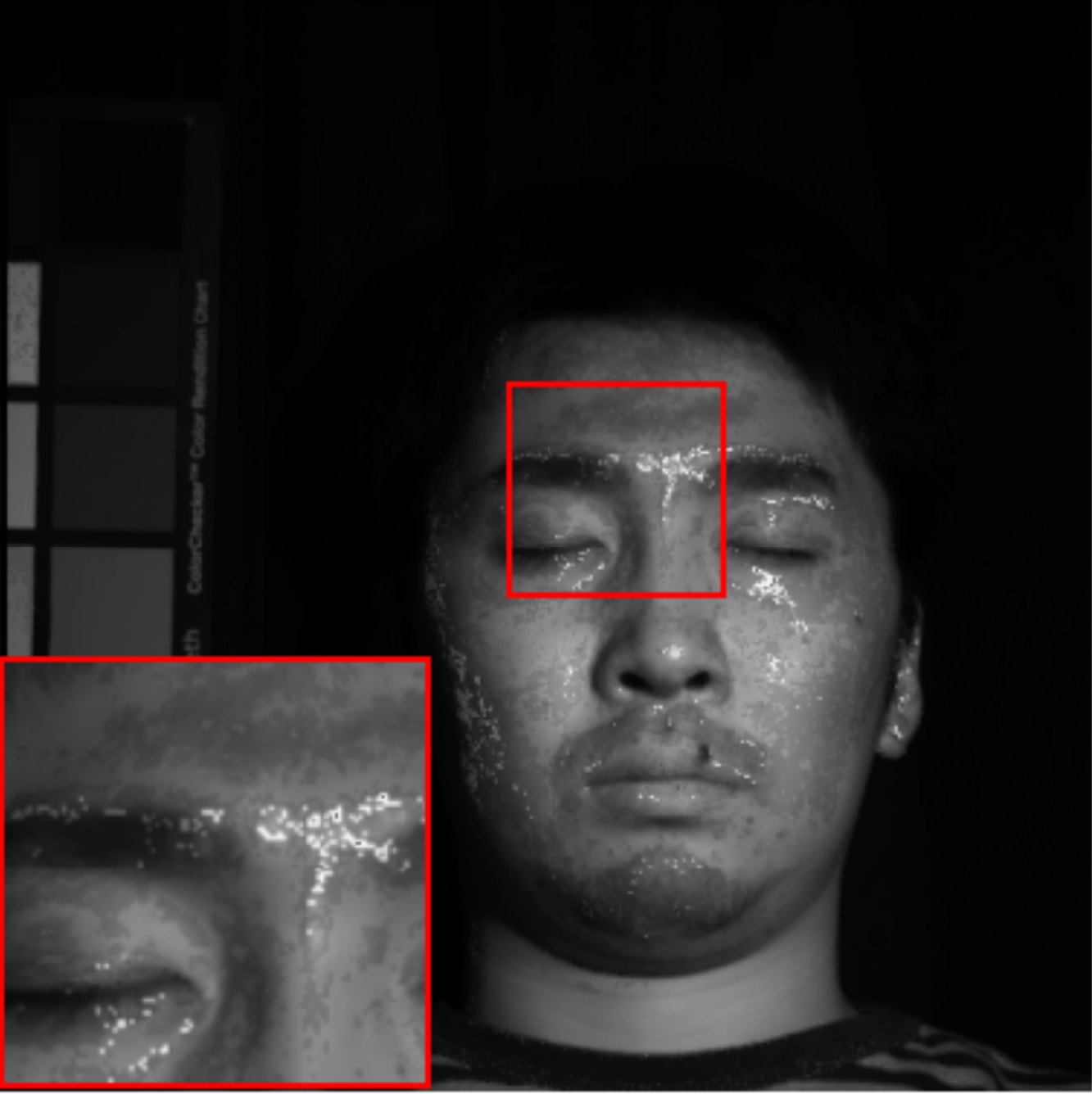}
\hspace{-0.15cm}
\includegraphics[height=1.0in, width=0.9in]{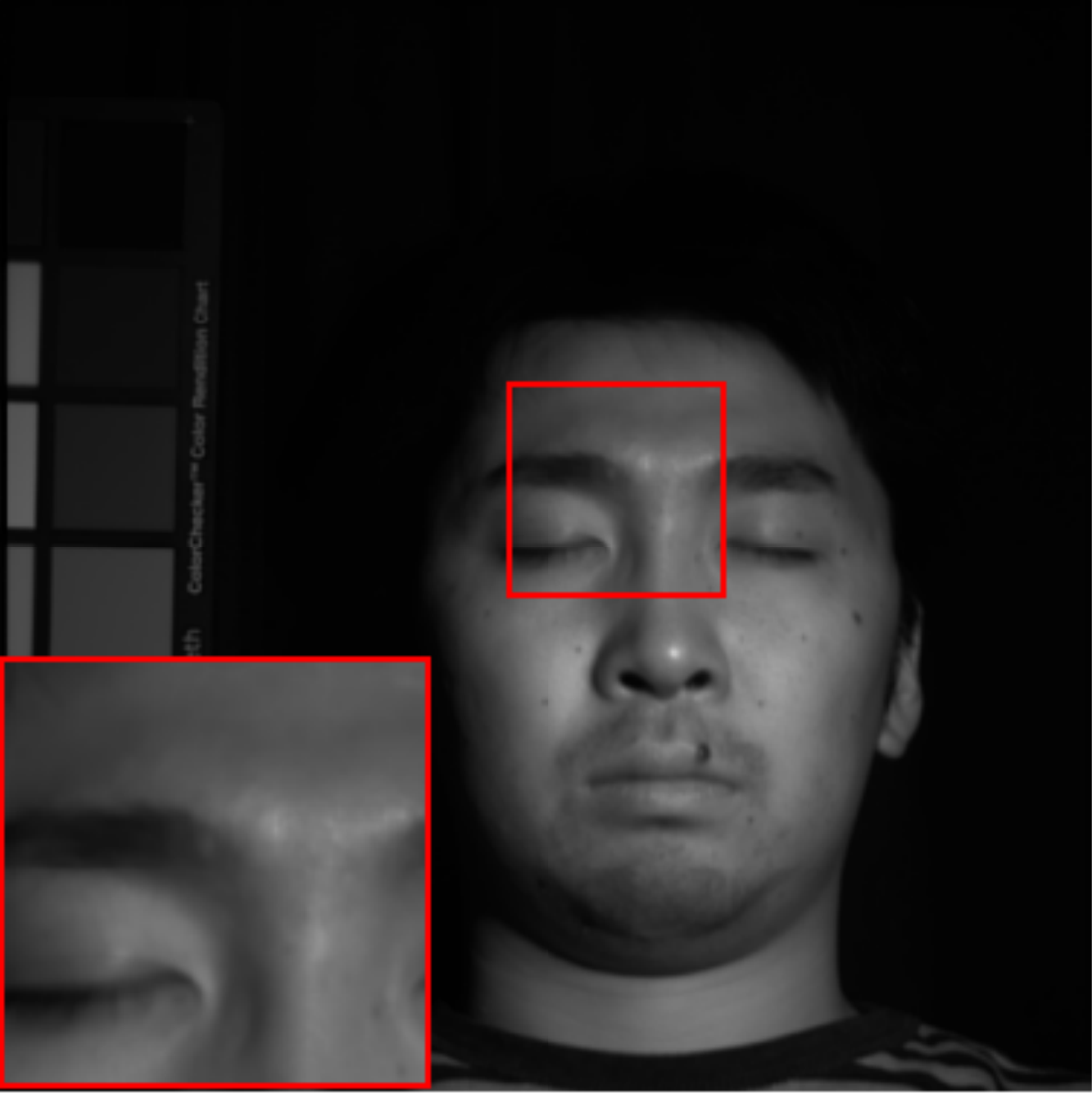}
\hspace{-0.15cm}
\includegraphics[height=1.0in, width=0.9in]{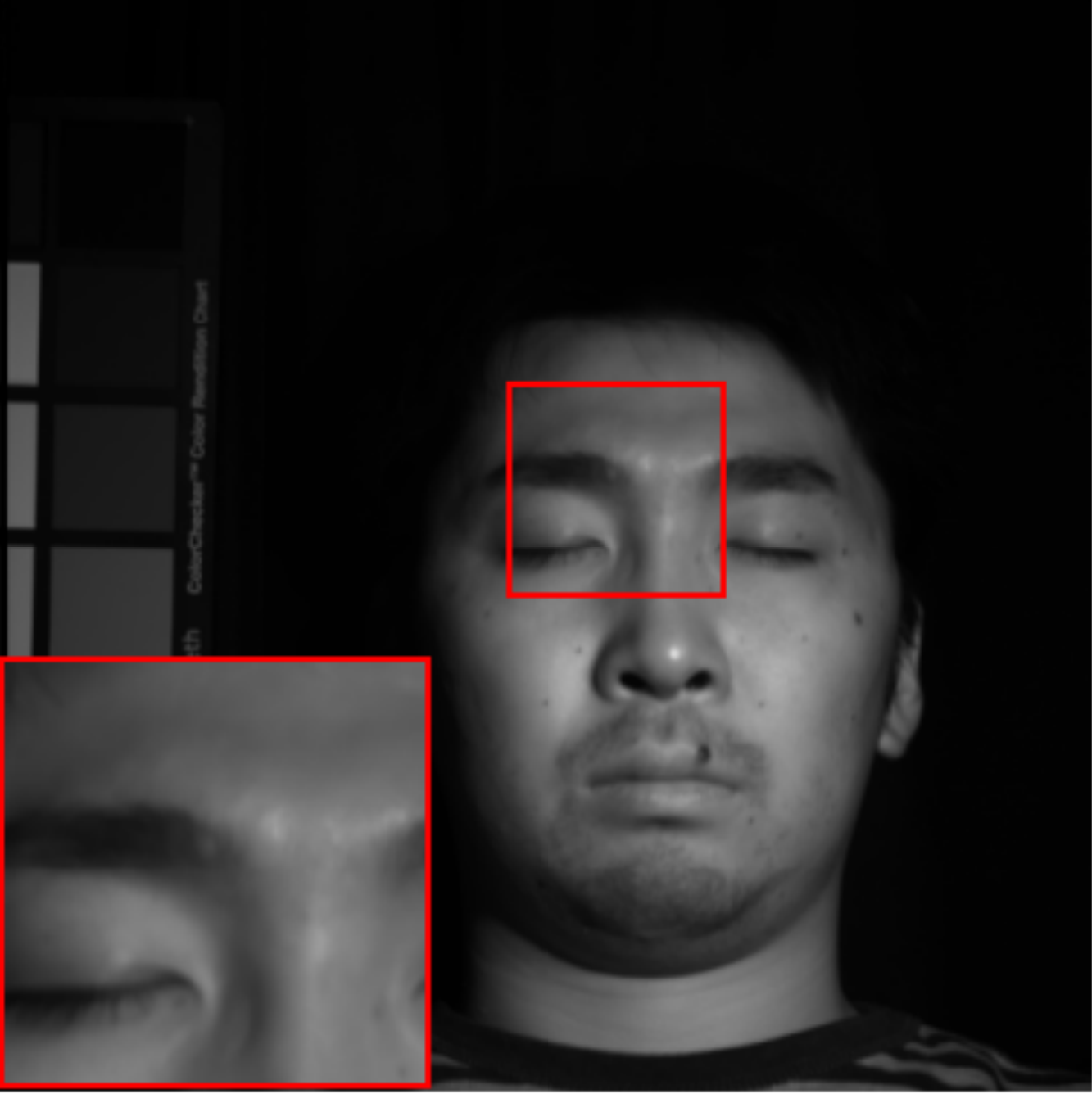}
\hspace{-0.15cm}
\includegraphics[height=1.0in, width=0.9in]{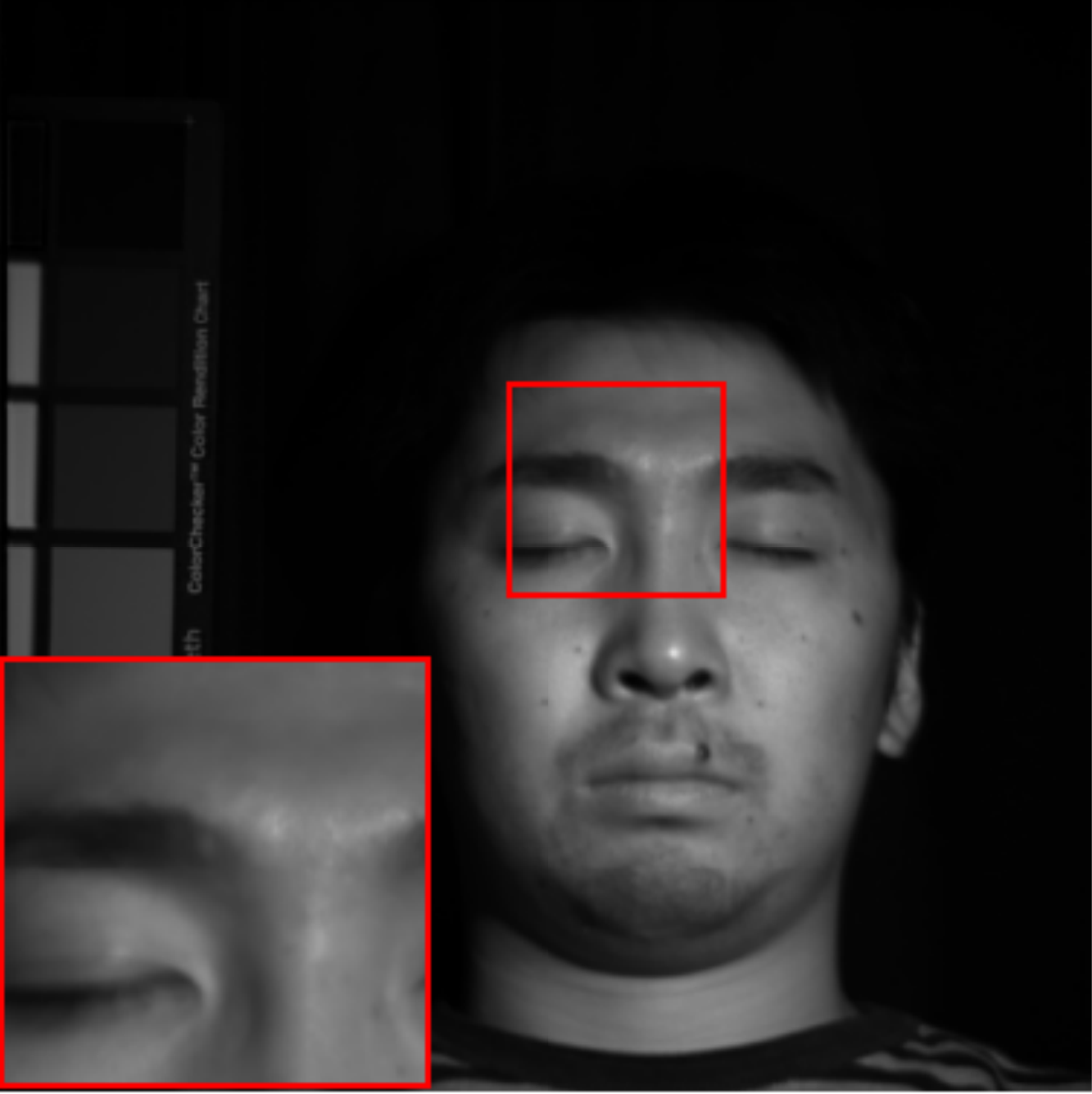}
\hspace{-0.15cm}
\includegraphics[height=1.0in, width=0.9in]{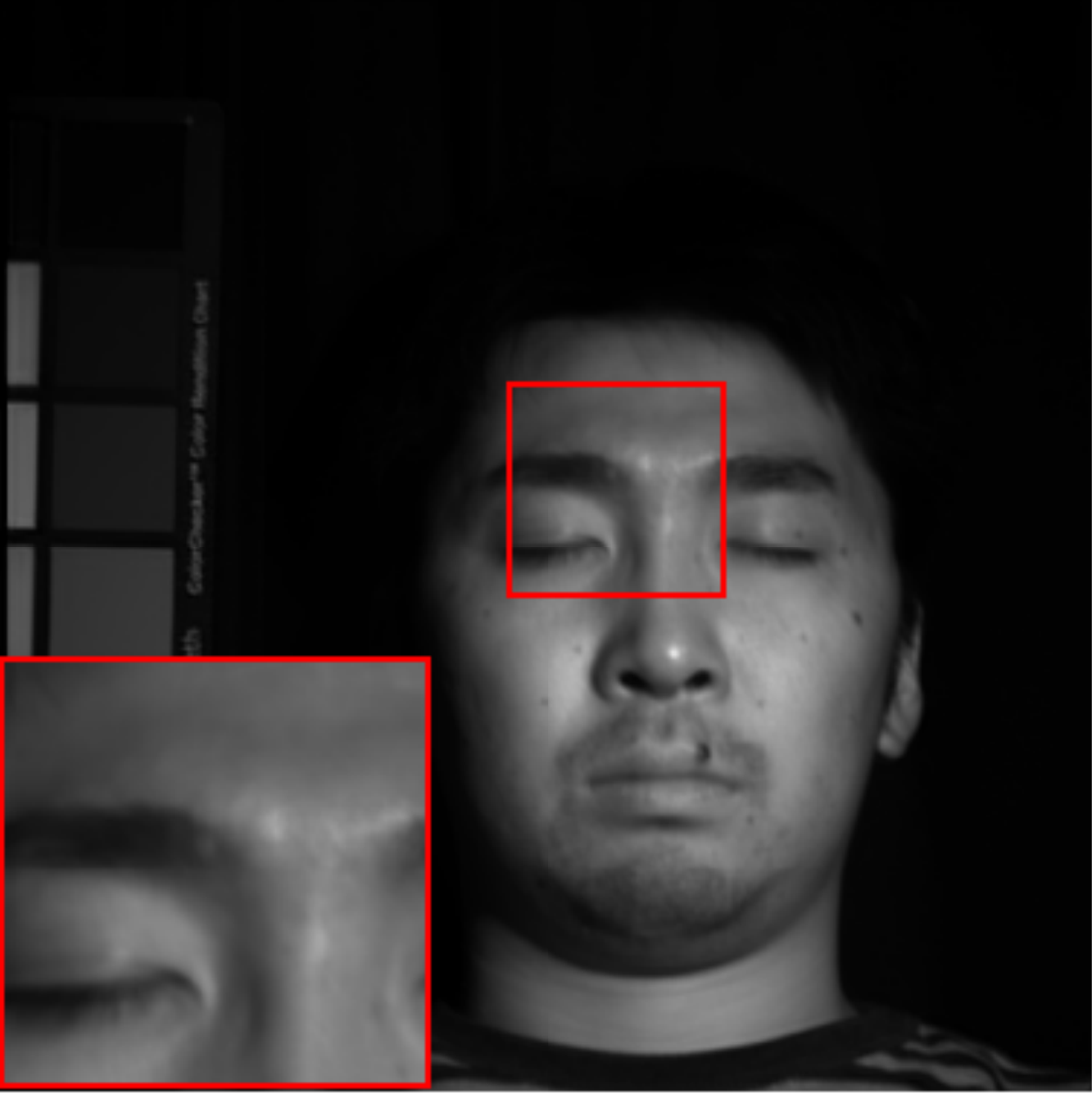}
\hspace{-0.15cm}
\includegraphics[height=1.0in, width=0.9in]{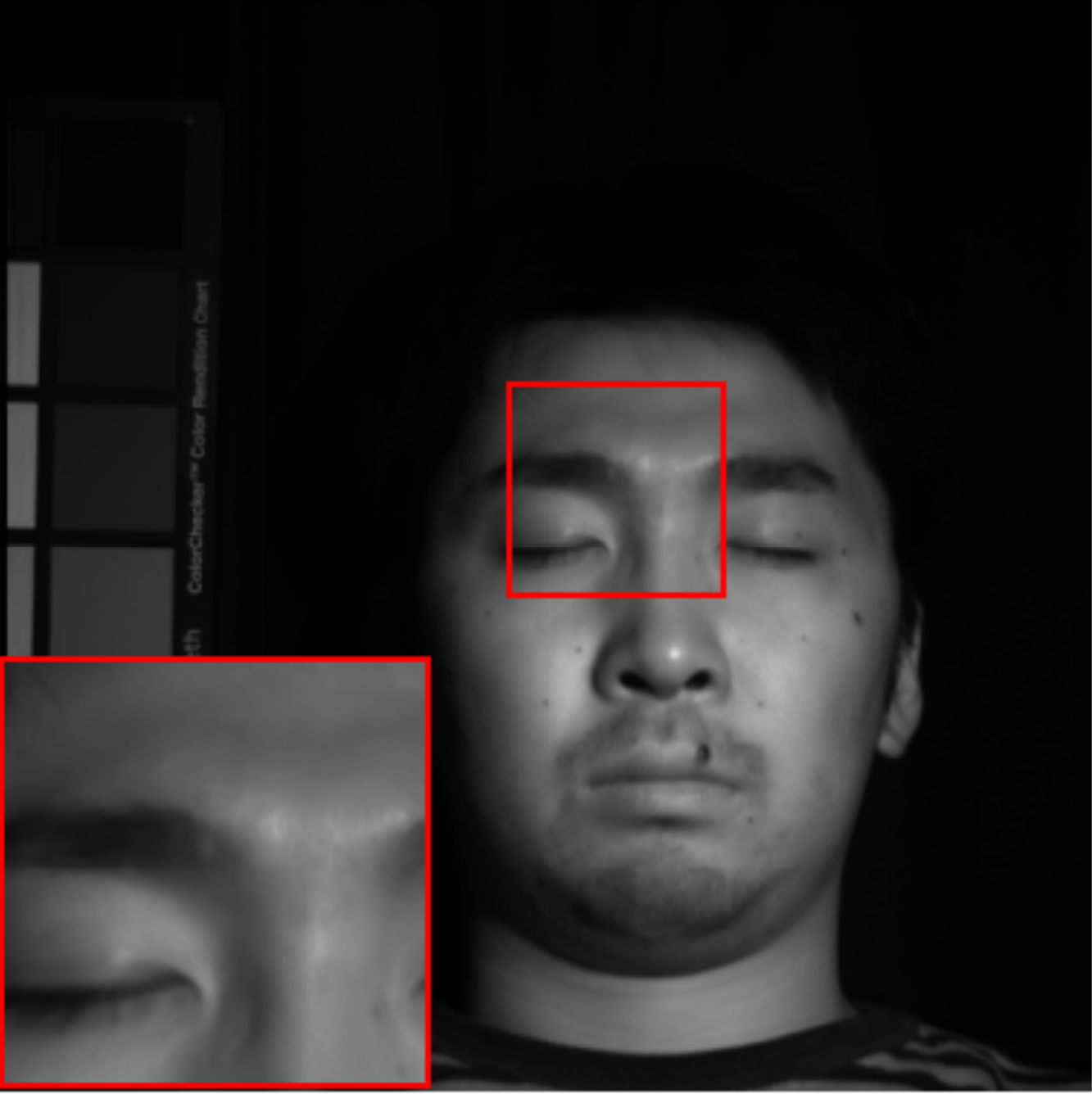}
\hspace{-0.15cm}
\includegraphics[height=1.0in, width=0.2in]{./error/blank.pdf}
\\
\vspace{-0.18cm}
\subfigure[BI~\cite{hou1978cubic}]{\includegraphics[height=1.0in, width=0.9in]{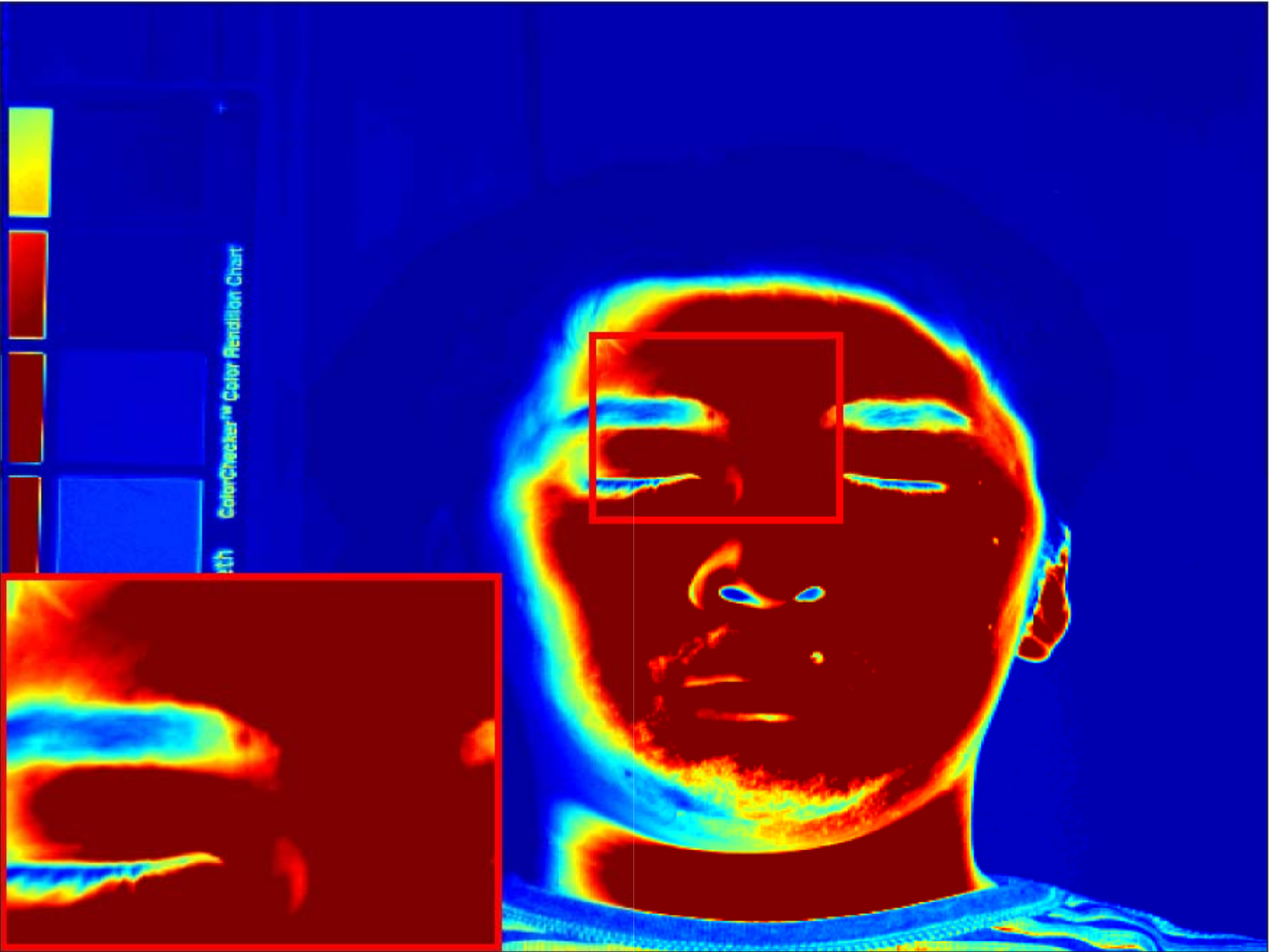}}
\hspace{-0.15cm}
\subfigure[Arad~\cite{arad2016sparse}]{\includegraphics[height=1.0in, width=0.9in]{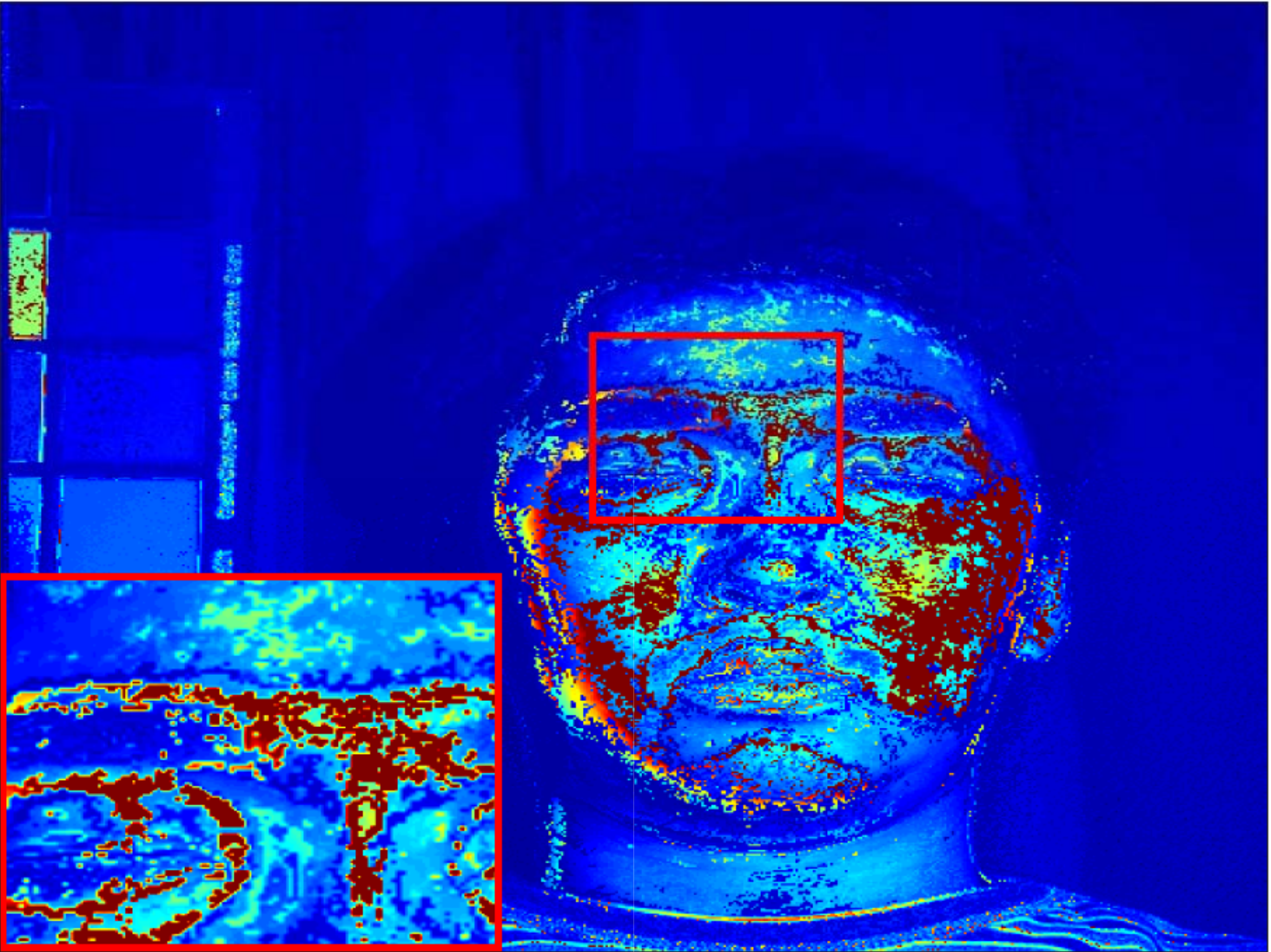}}
\hspace{-0.15cm}
\subfigure[Aitor~\cite{alvarez2017adversarial}]{\includegraphics[height=1.0in, width=0.9in]{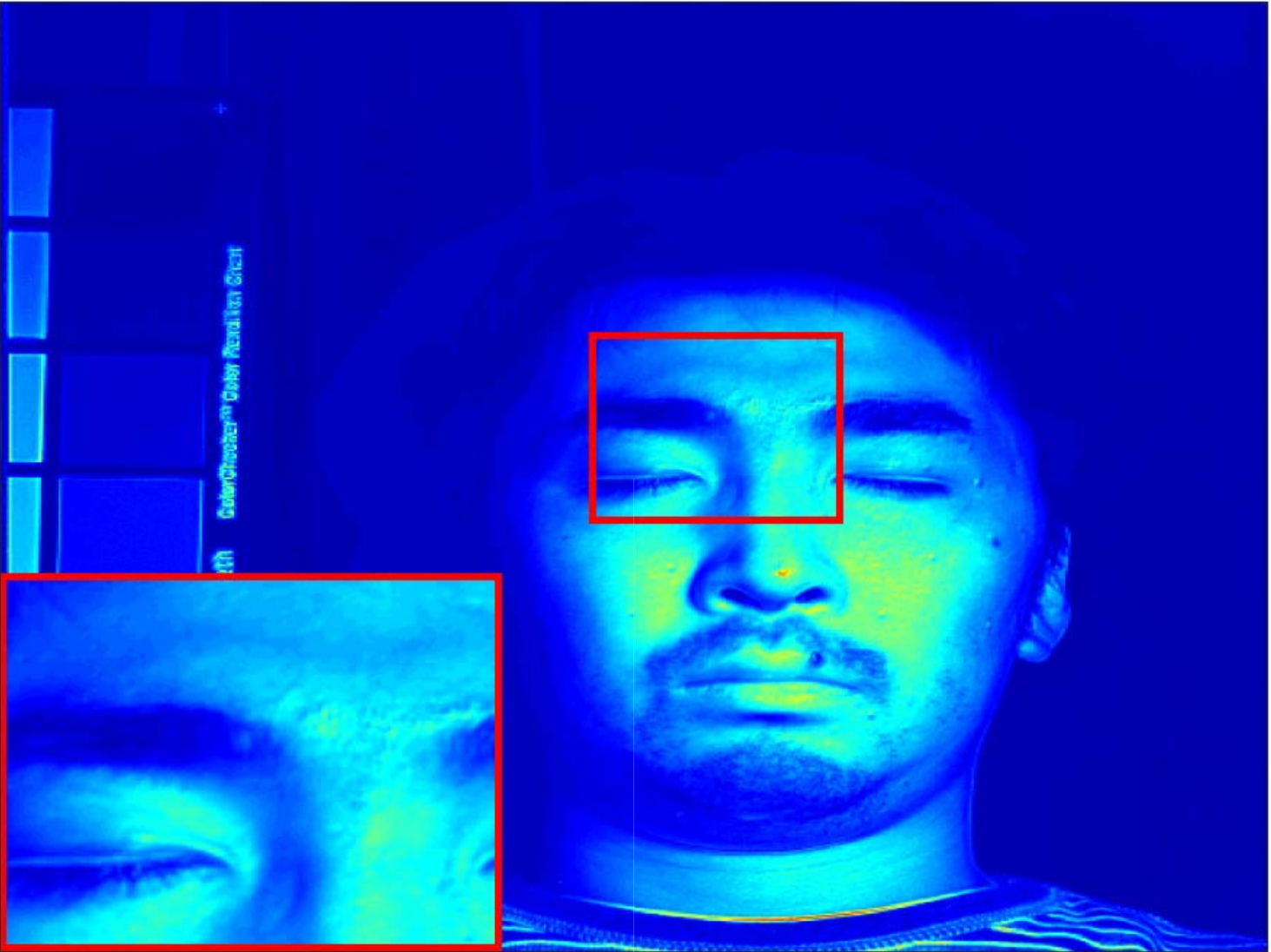}}
\hspace{-0.15cm}
\subfigure[HSCNN+~\cite{xiong2017hscnn}]{\includegraphics[height=1.0in, width=0.9in]{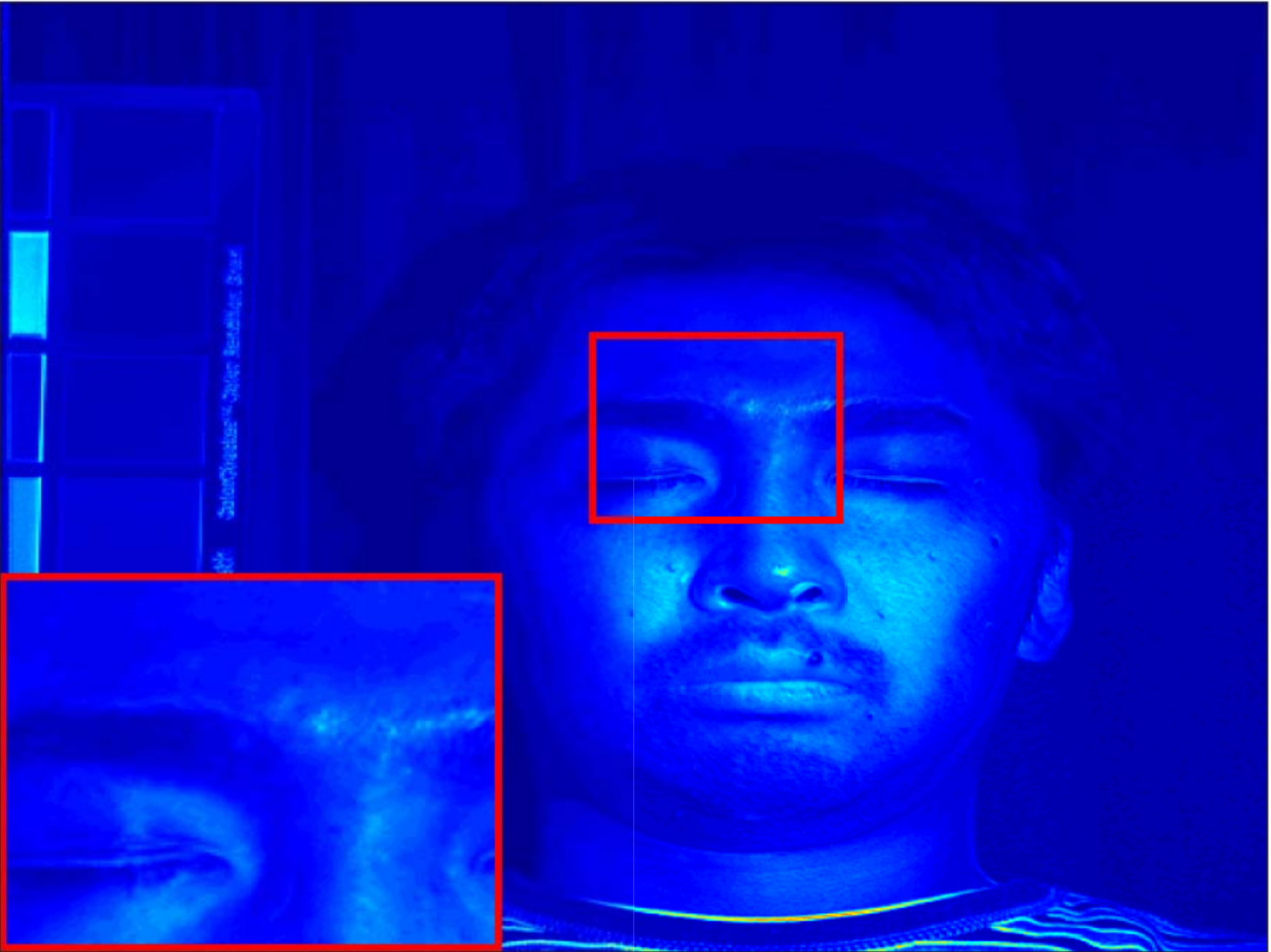}}
\hspace{-0.15cm}
\subfigure[DCNN]{\includegraphics[height=1.0in, width=0.9in]{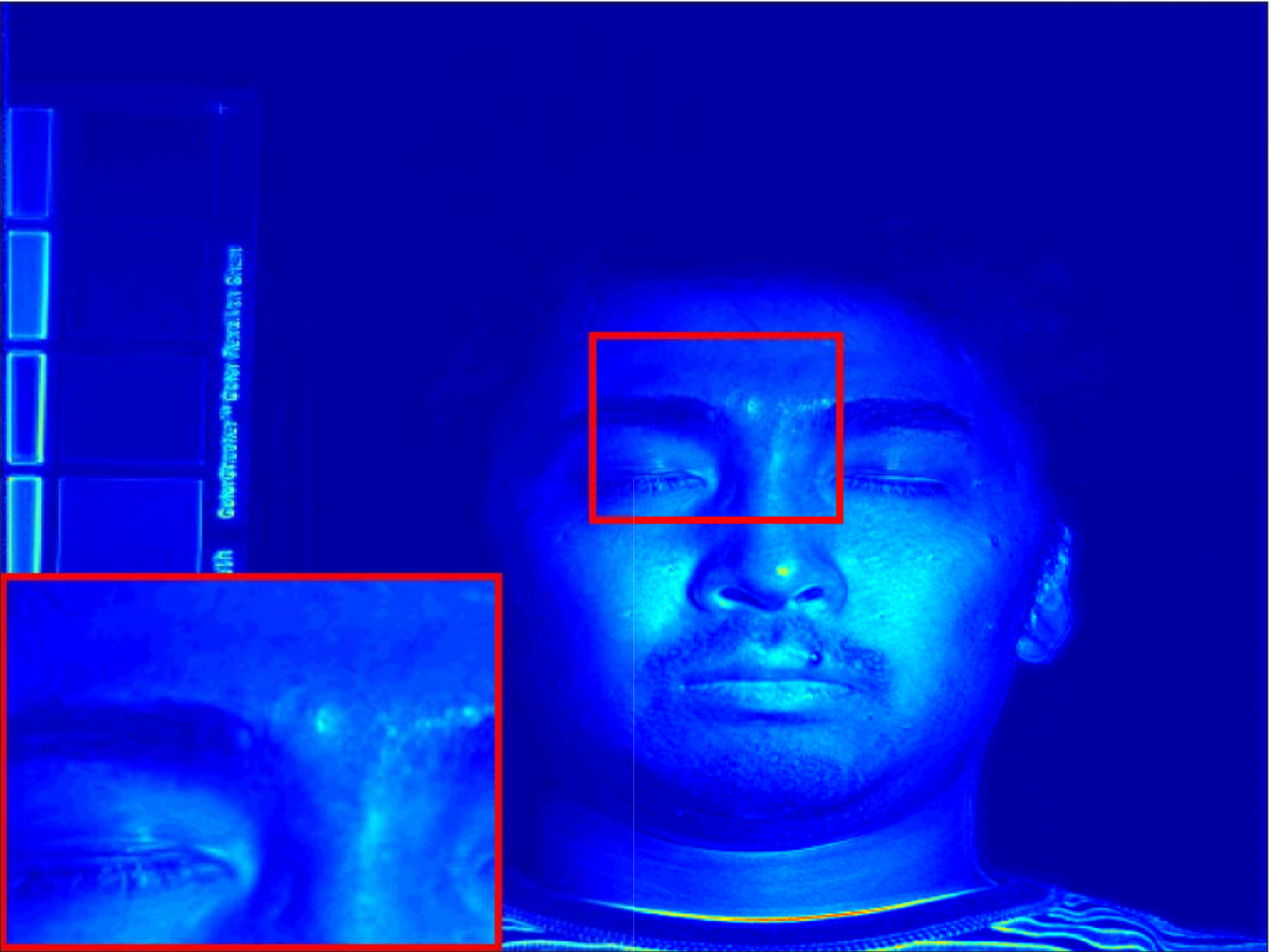}}
\hspace{-0.15cm}
\subfigure[MCNet]{\includegraphics[height=1.0in, width=0.9in]{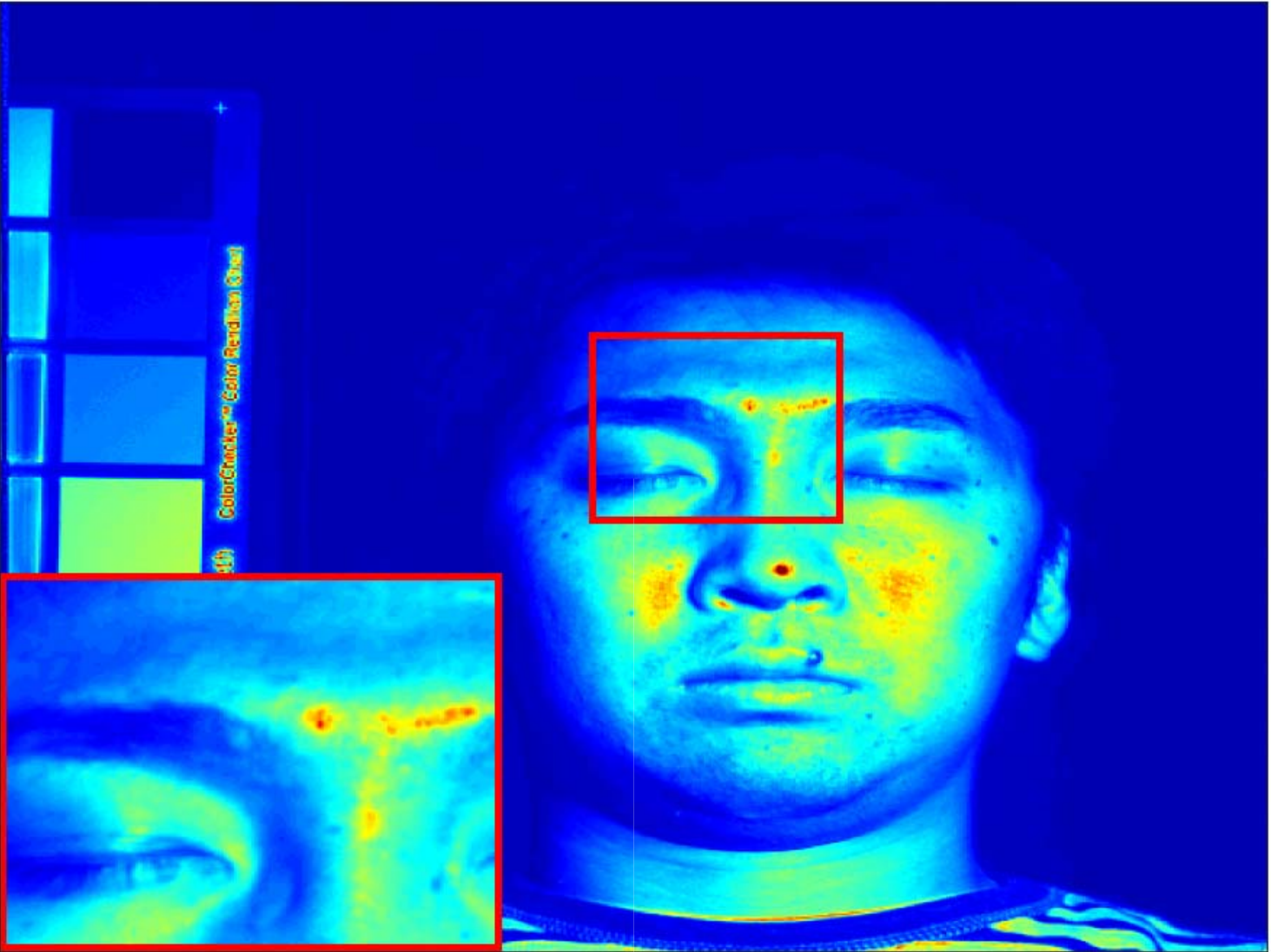}}
\hspace{-0.15cm}
\subfigure[Ours]{\includegraphics[height=1.0in, width=0.9in]{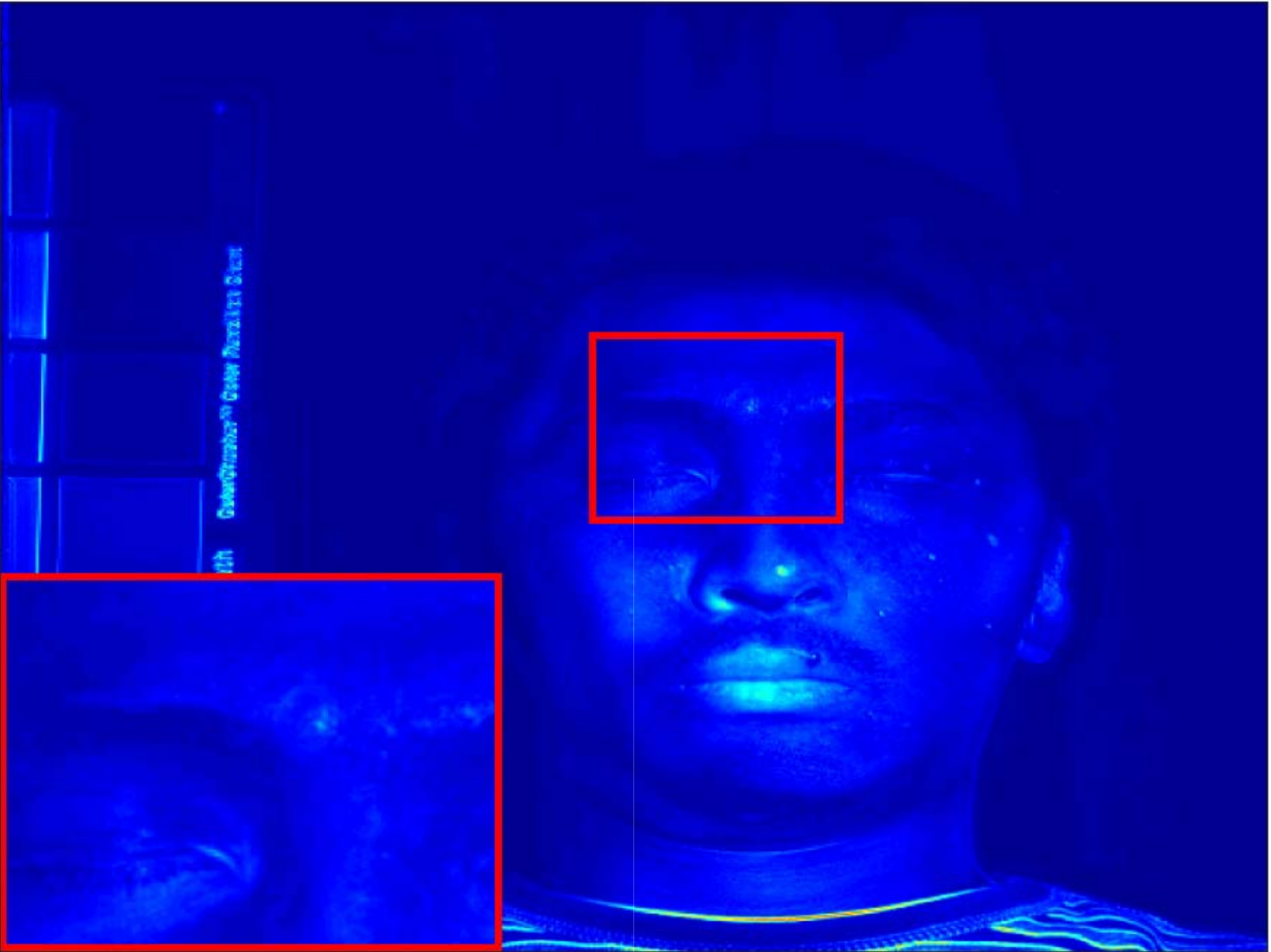}}
\hspace{-0.15cm}
\includegraphics[height=1.0in, width=0.2in]{./error/colorbar_30.pdf}
\caption{Visual super-resolution results of the 28-th band and the reconstruction error maps of an example image from the CAVE dataset for different methods. The reconstruction error is obtained by computing the mean-square error between two spectrum vectors from the super-resolution result and the ground truth at each pixel. Best view on the screen.}
\label{fig:visual-cave}
\vspace{-0.3cm}
\end{figure*}

\begin{figure*}[htbp]
\centering
\includegraphics[height=1.0in, width=0.9in]{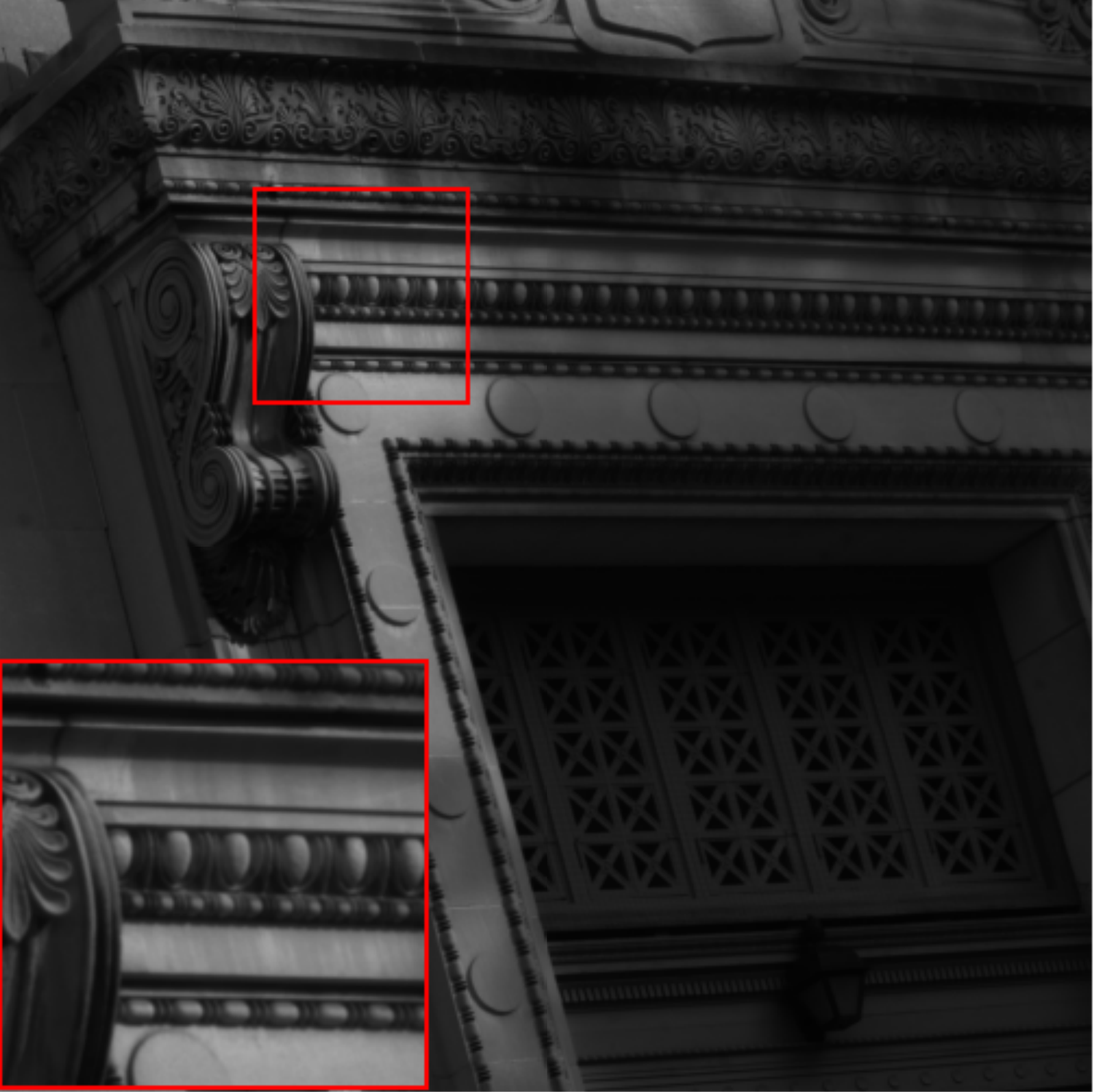}
\hspace{-0.15cm}
\includegraphics[height=1.0in, width=0.9in]{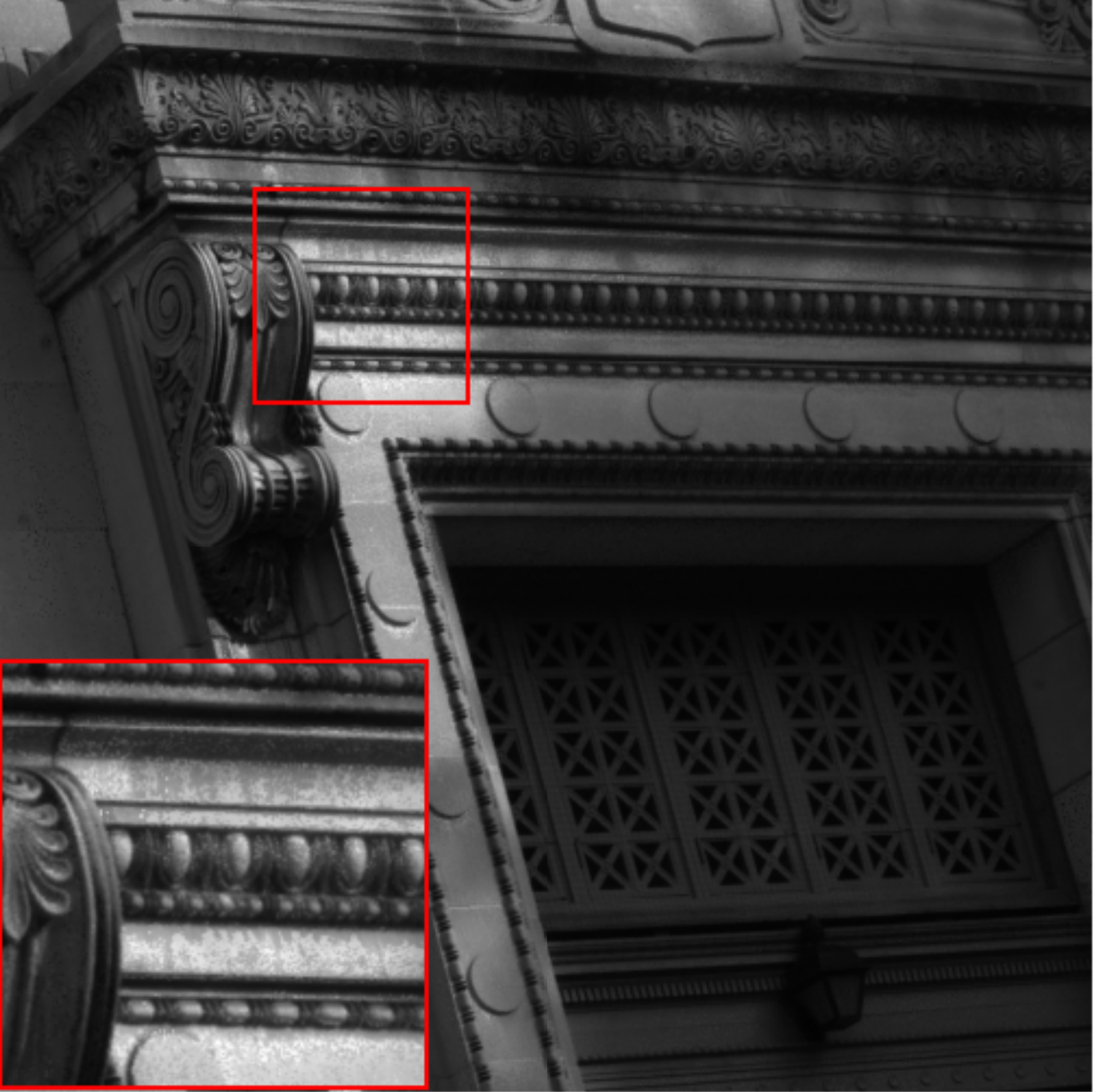}
\hspace{-0.15cm}
\includegraphics[height=1.0in, width=0.9in]{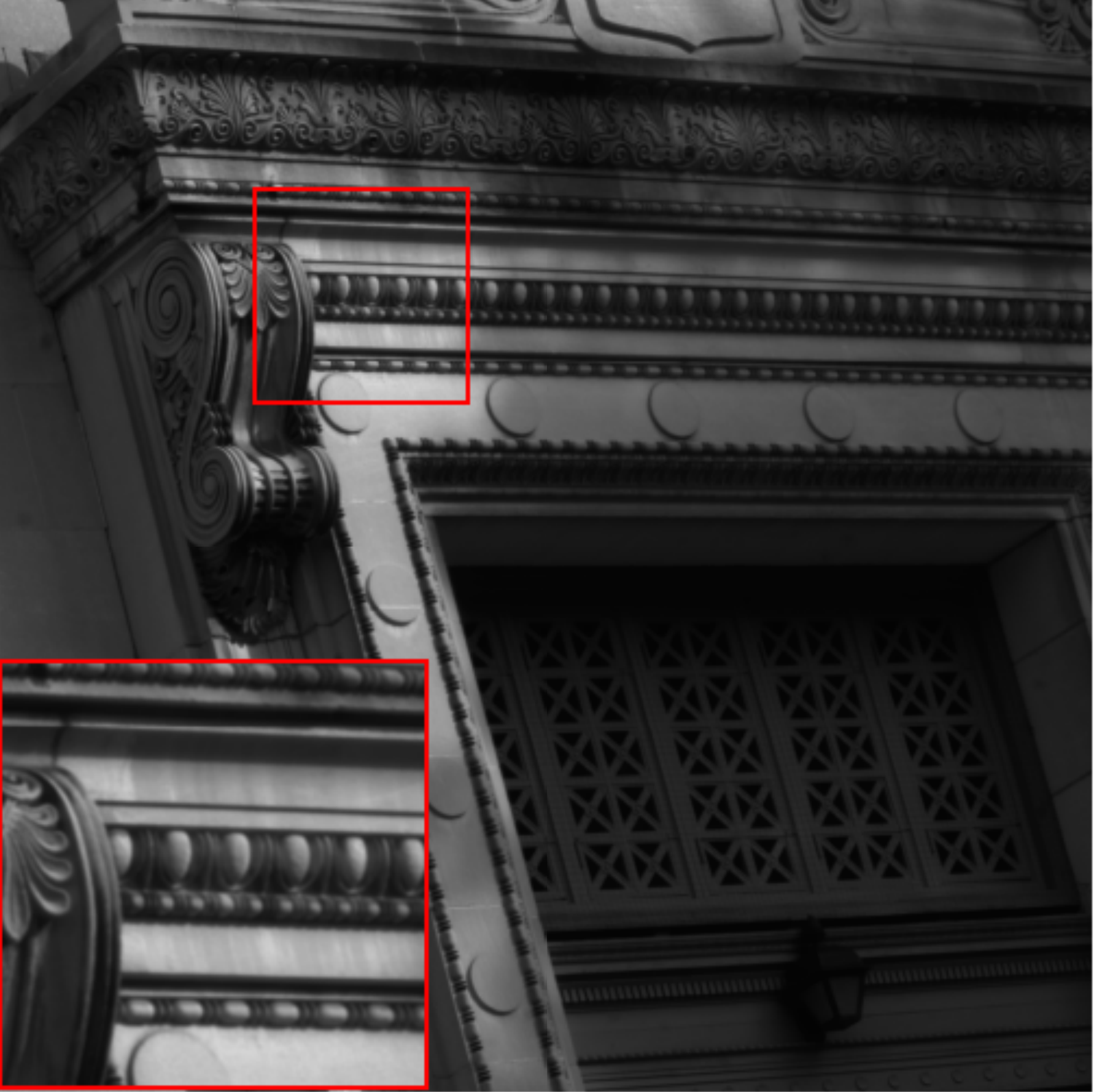}
\hspace{-0.15cm}
\includegraphics[height=1.0in, width=0.9in]{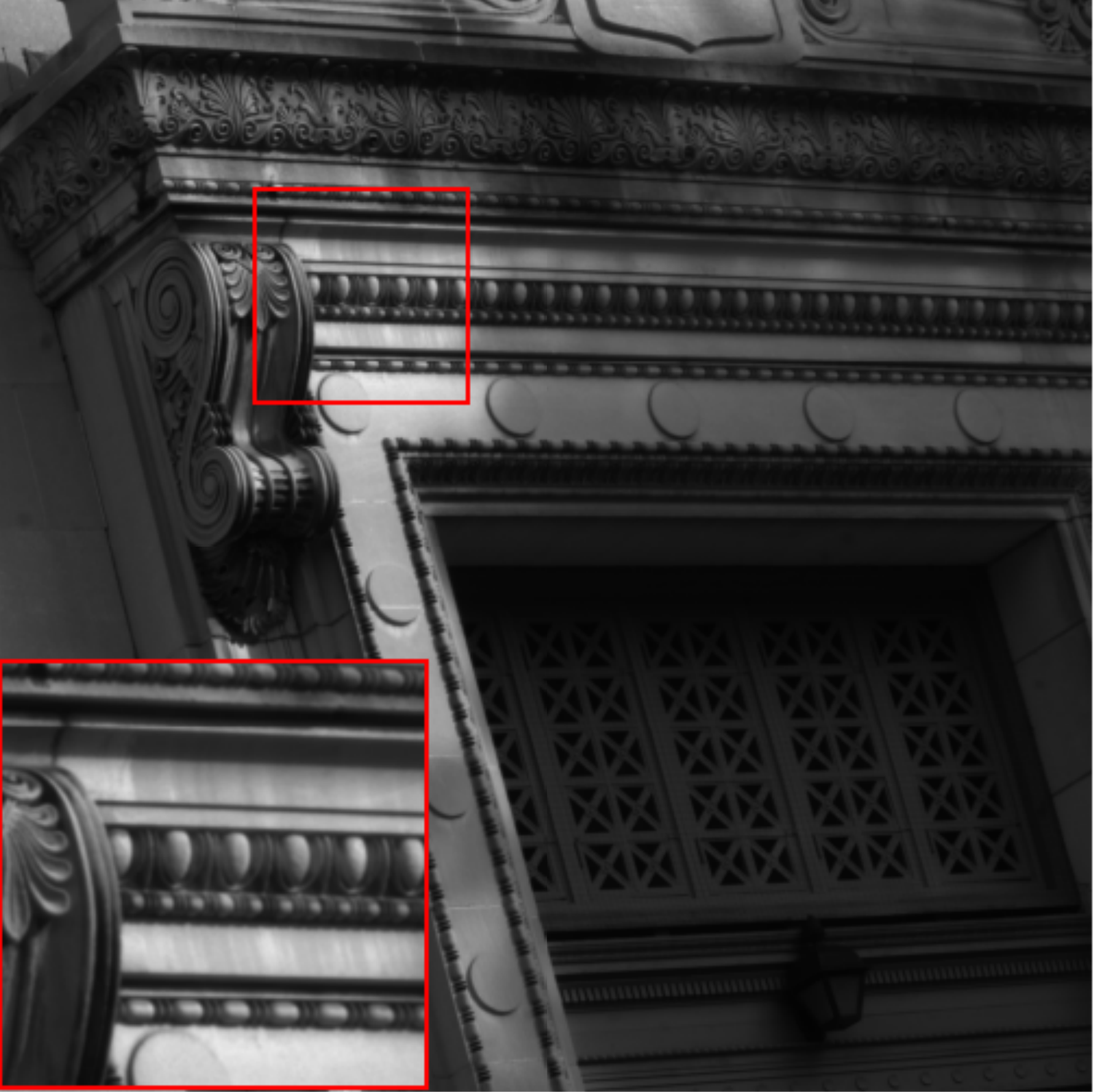}
\hspace{-0.15cm}
\includegraphics[height=1.0in, width=0.9in]{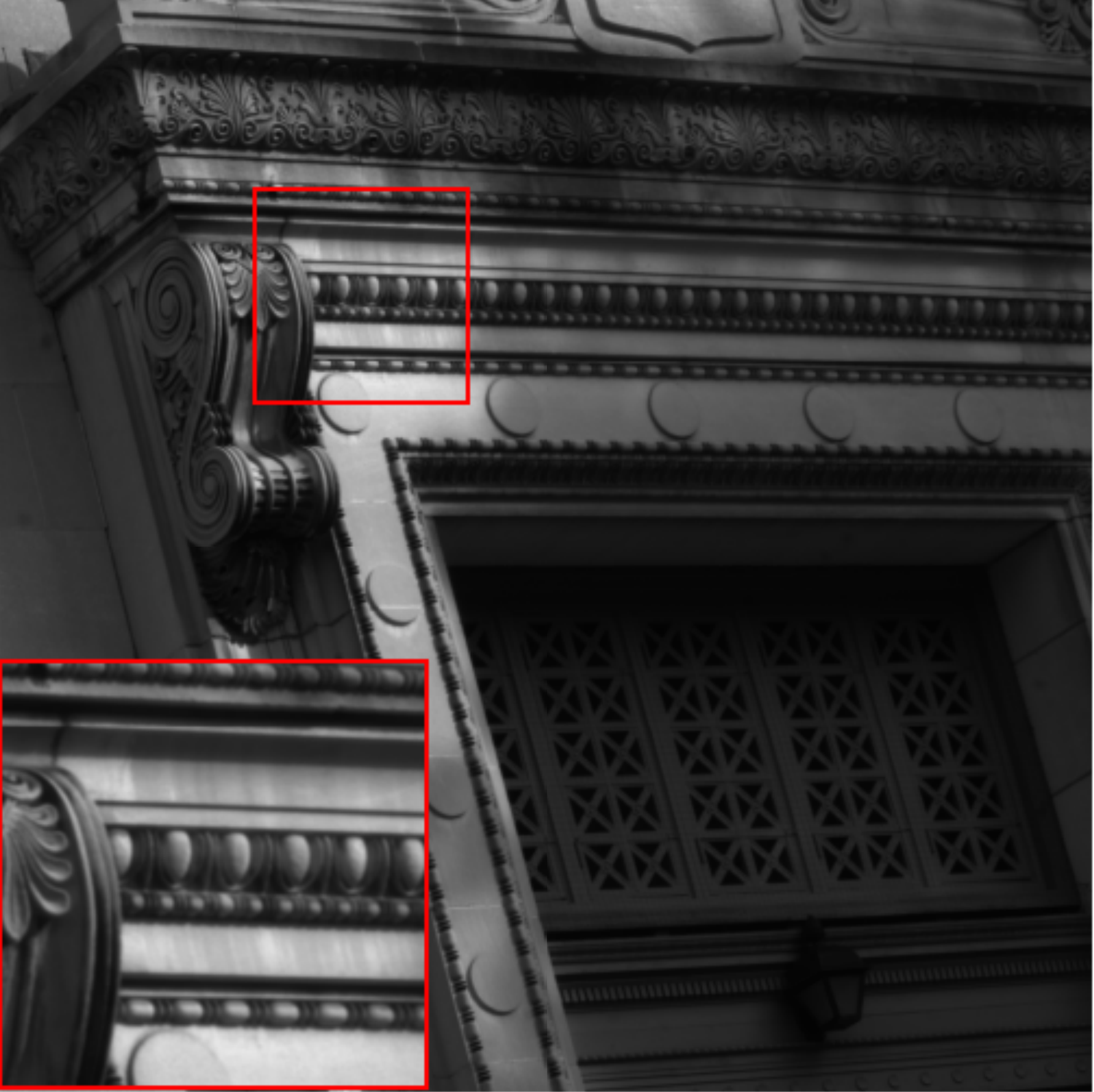}
\hspace{-0.15cm}
\includegraphics[height=1.0in, width=0.9in]{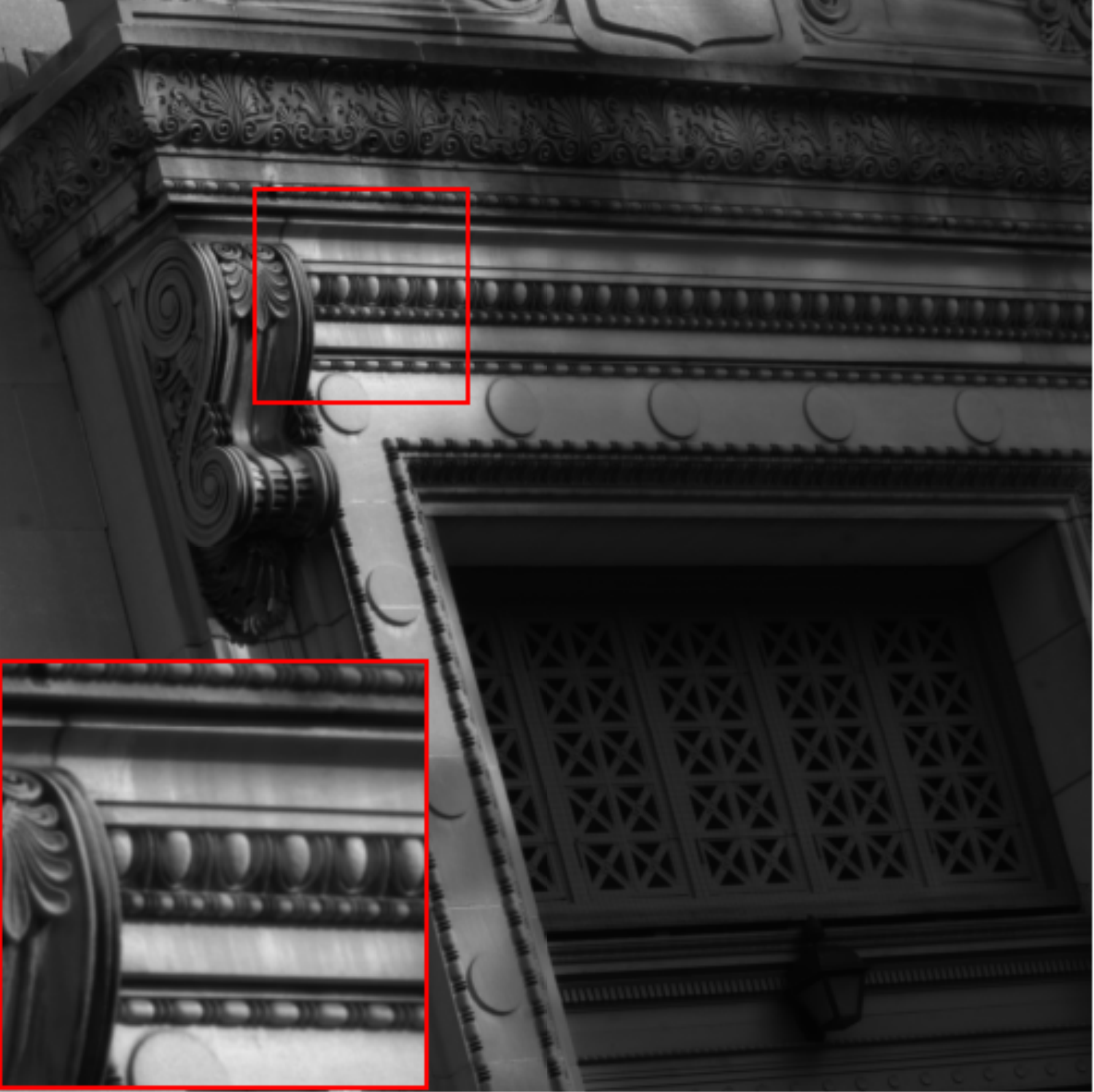}
\hspace{-0.15cm}
\includegraphics[height=1.0in, width=0.9in]{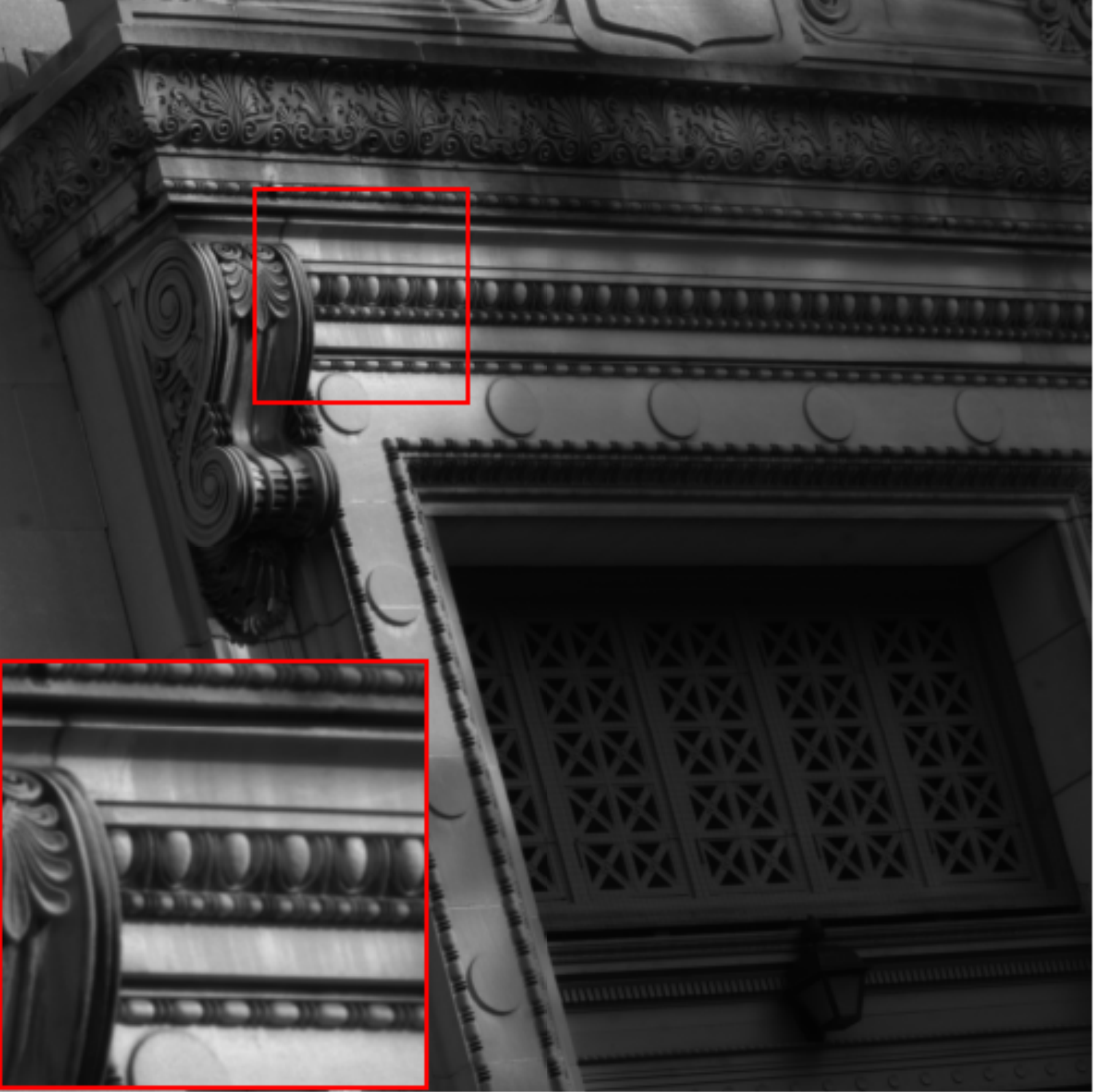}
\hspace{-0.15cm}
\includegraphics[height=1.0in, width=0.2in]{./error/blank.pdf}
\\
\vspace{-0.18cm}
\subfigure[BI~\cite{hou1978cubic}]{\includegraphics[height=1.0in, width=0.9in]{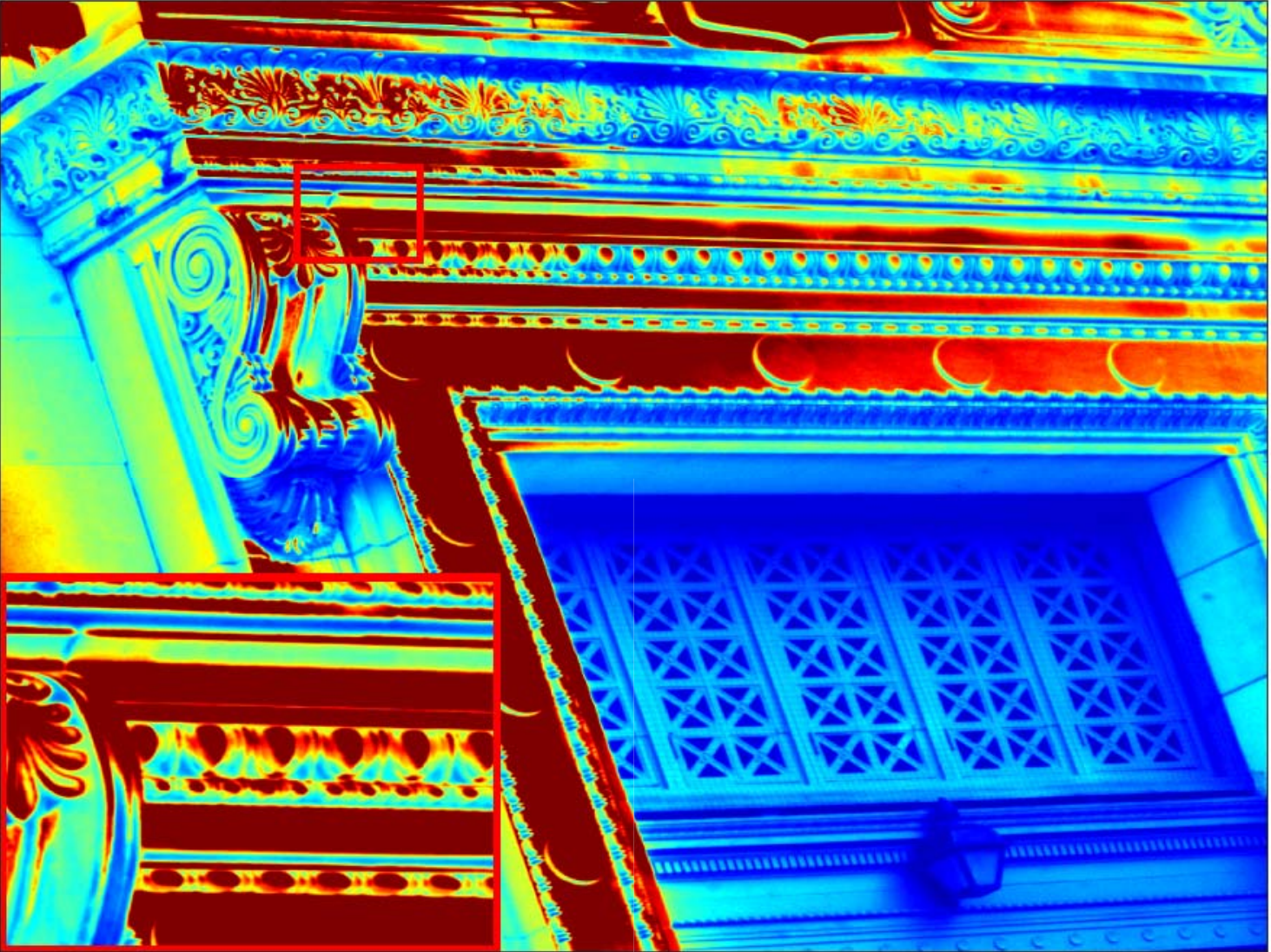}}
\hspace{-0.15cm}
\subfigure[Arad~\cite{arad2016sparse}]{\includegraphics[height=1.0in, width=0.9in]{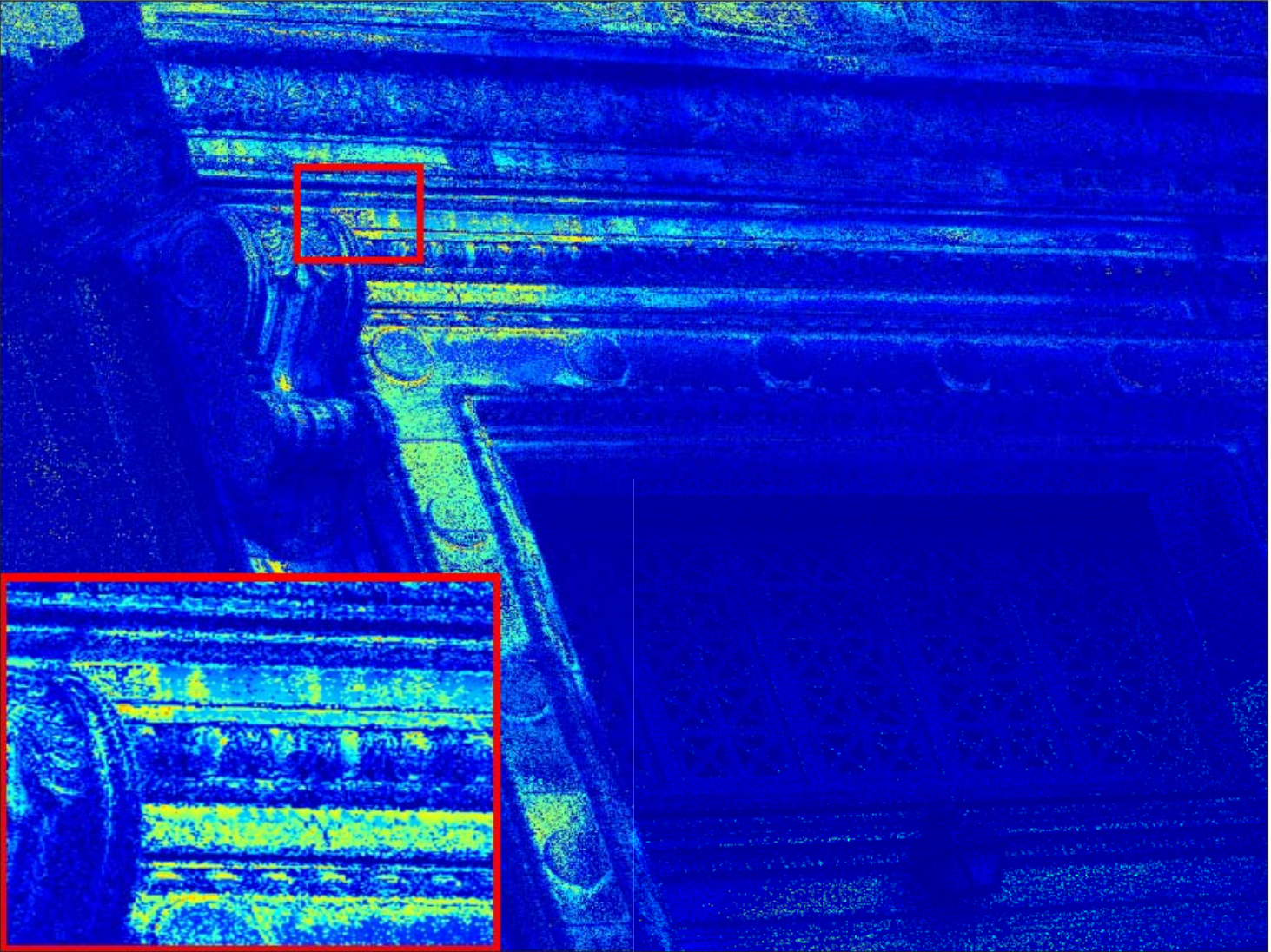}}
\hspace{-0.15cm}
\subfigure[Aitor~\cite{alvarez2017adversarial}]{\includegraphics[height=1.0in, width=0.9in]{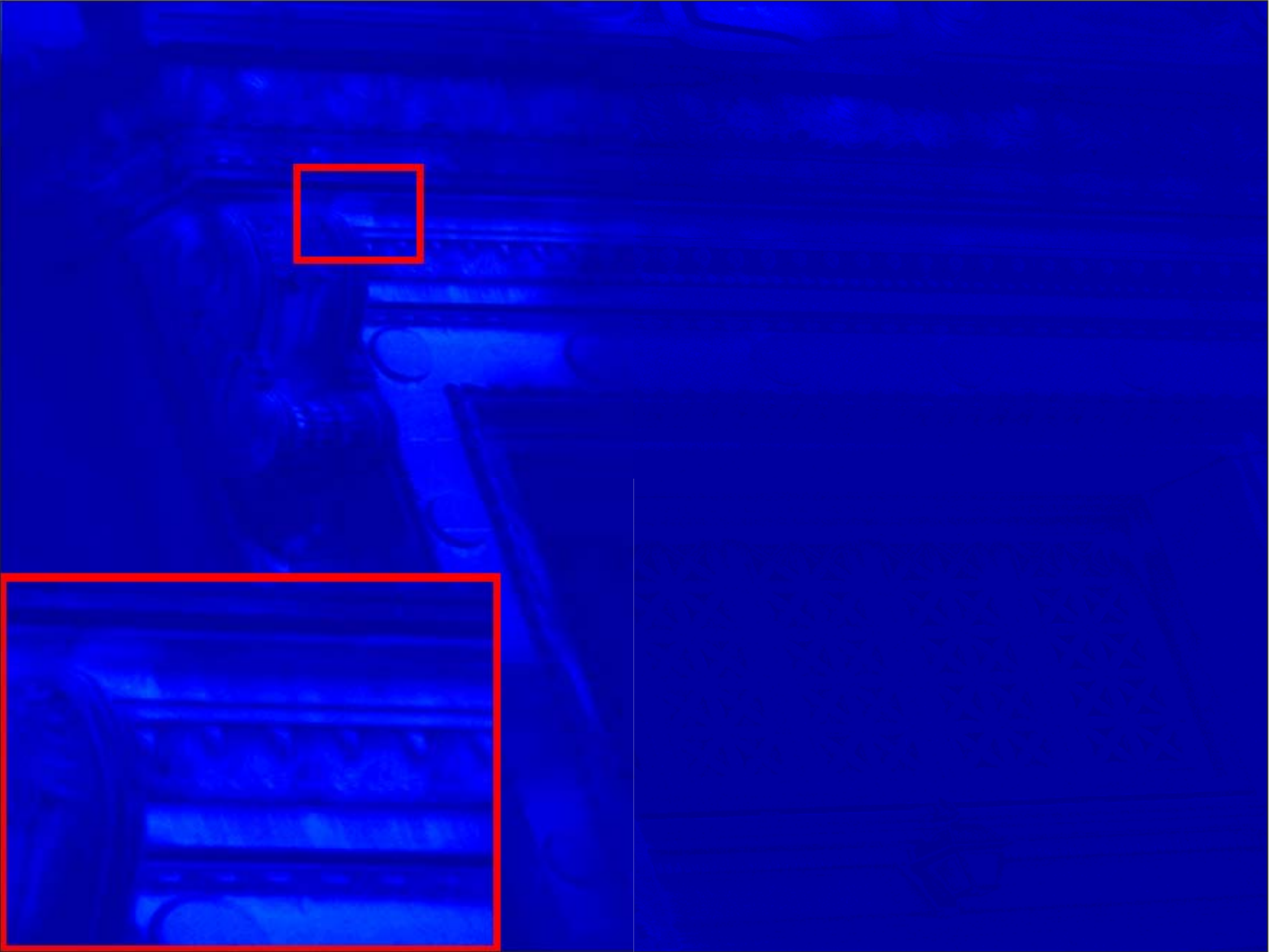}}
\hspace{-0.15cm}
\subfigure[HSCNN+~\cite{xiong2017hscnn}]{\includegraphics[height=1.0in, width=0.9in]{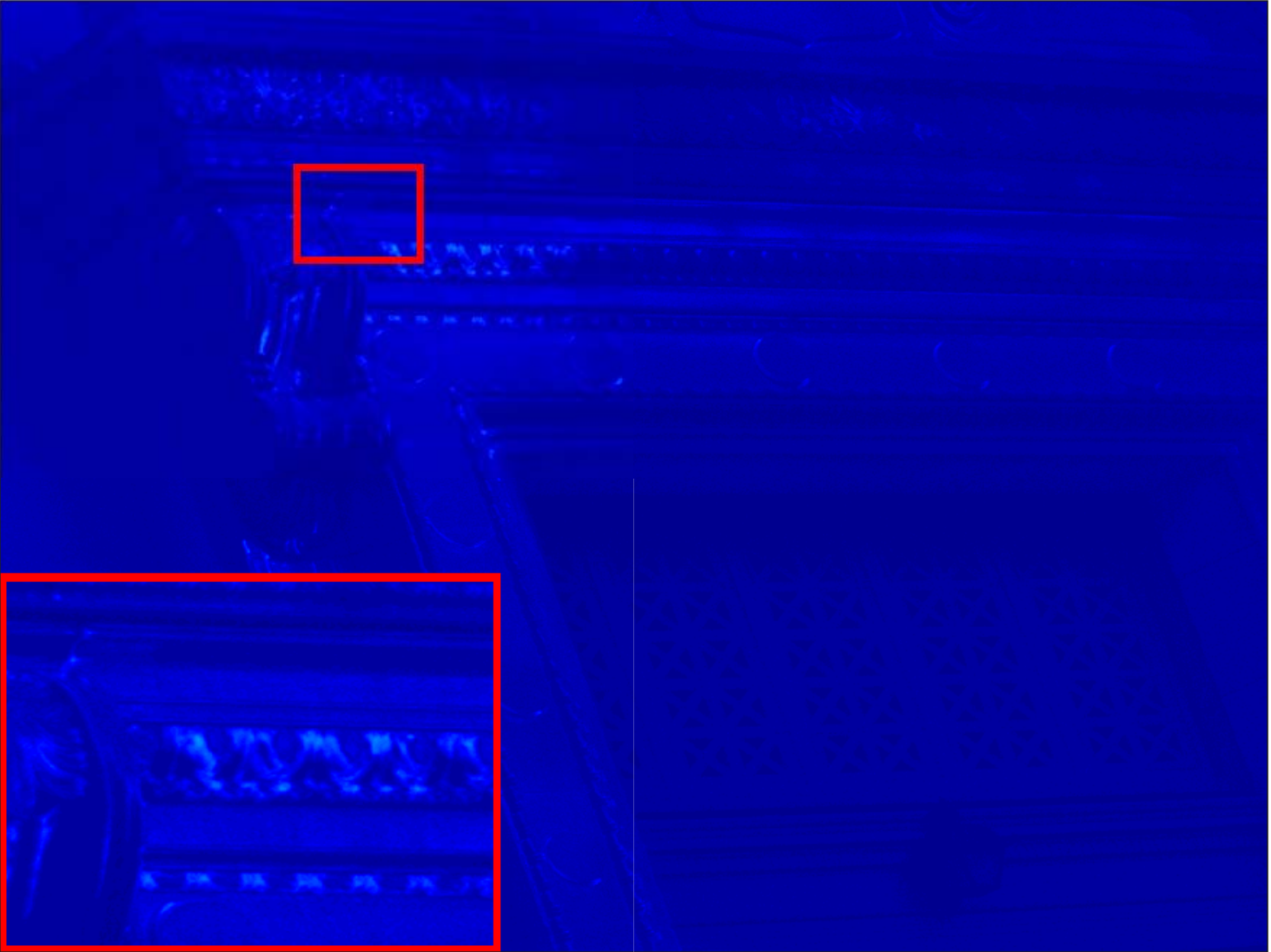}}
\hspace{-0.15cm}
\subfigure[DCNN]{\includegraphics[height=1.0in, width=0.9in]{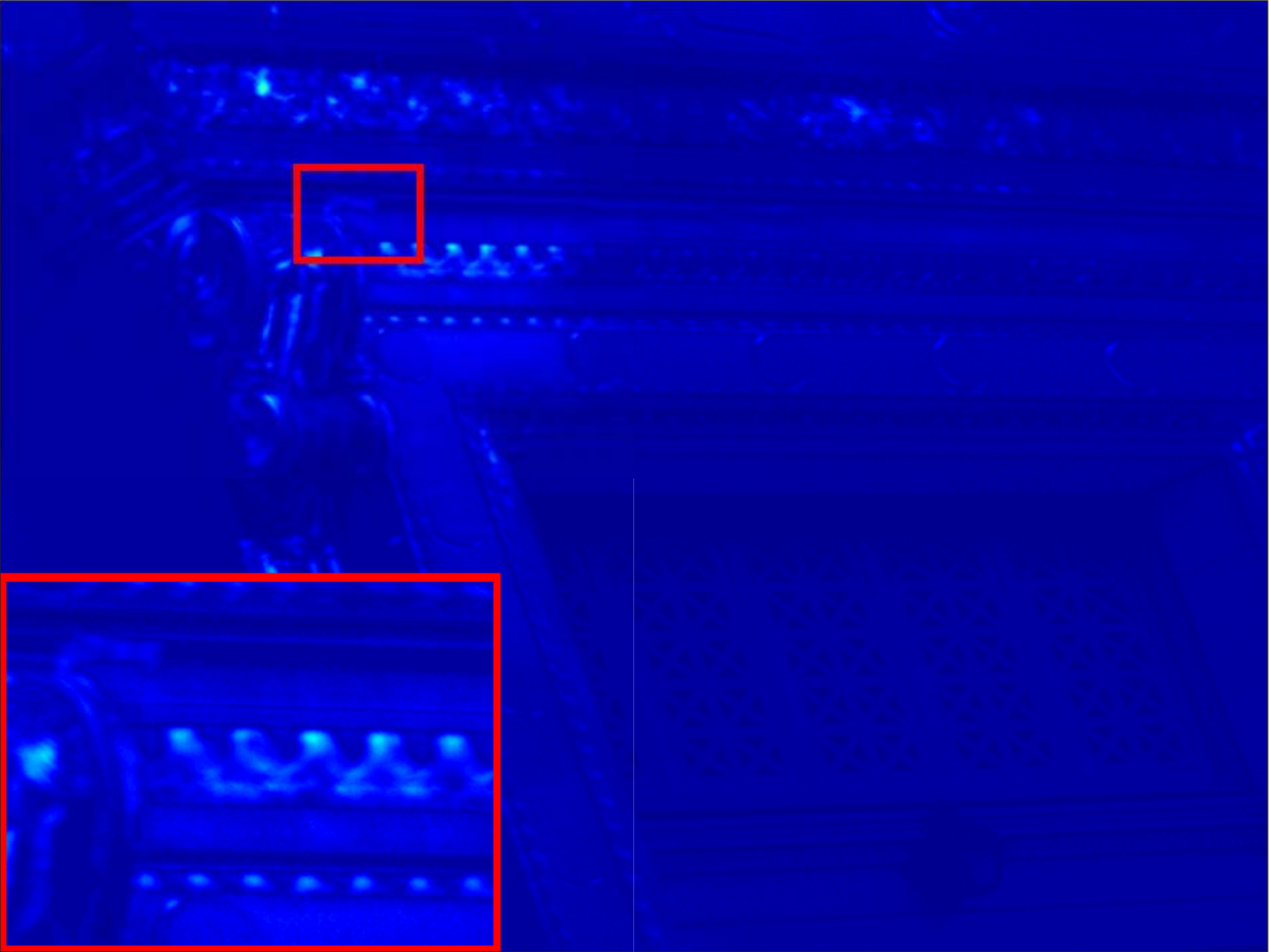}}
\hspace{-0.15cm}
\subfigure[MCNet]{\includegraphics[height=1.0in, width=0.9in]{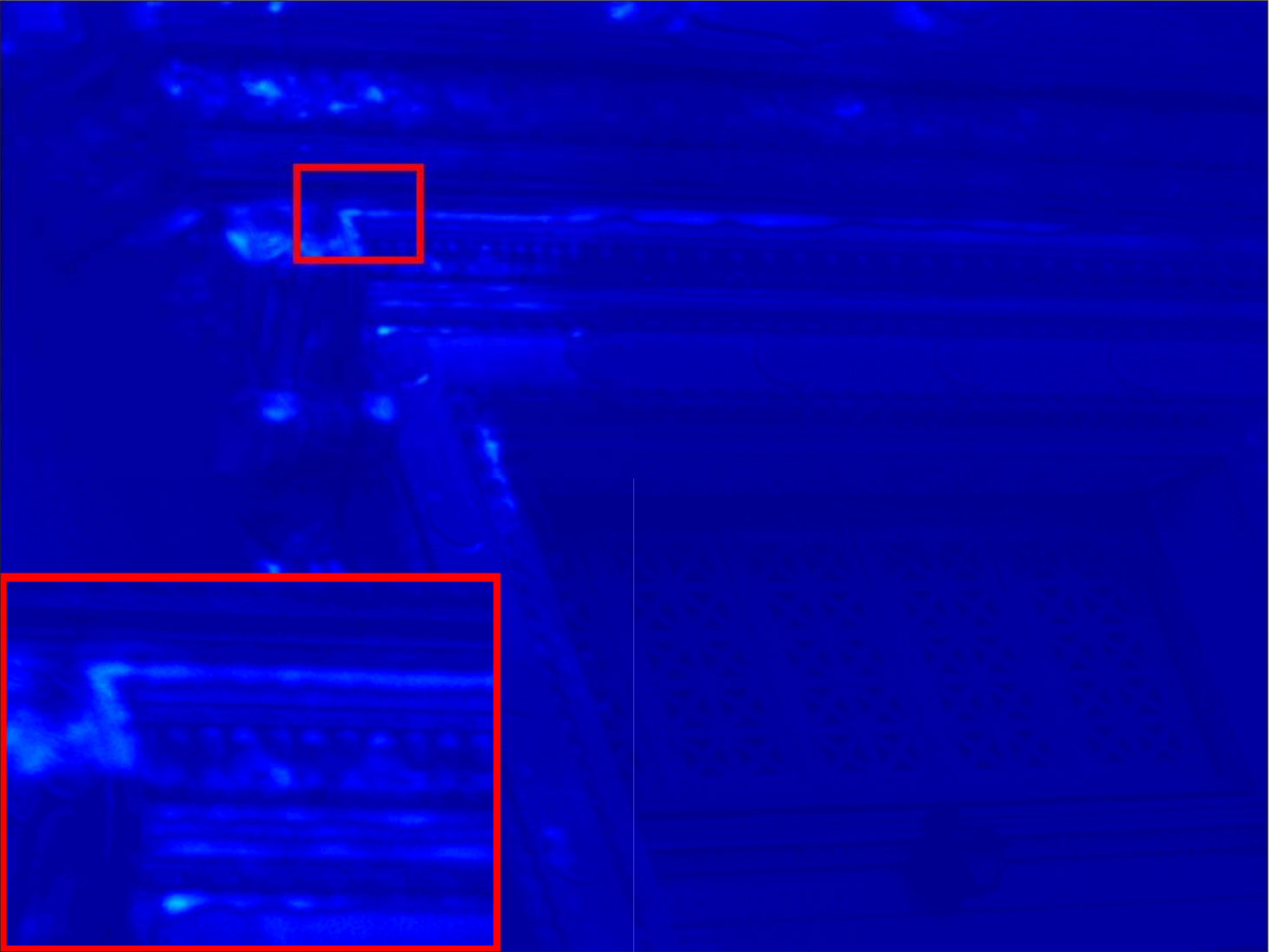}}
\hspace{-0.15cm}
\subfigure[Ours]{\includegraphics[height=1.0in, width=0.9in]{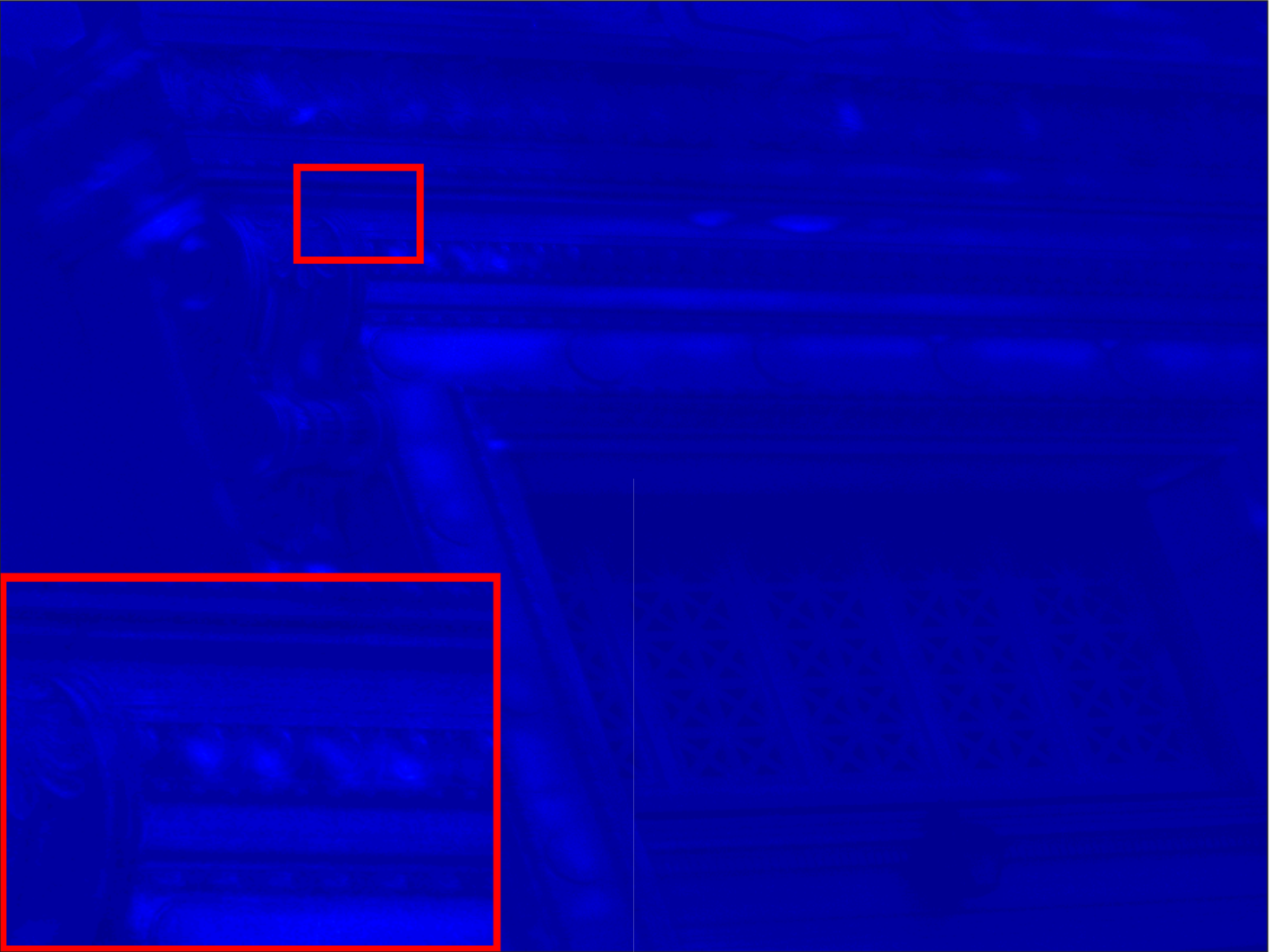}}
\hspace{-0.15cm}
\includegraphics[height=1.0in, width=0.2in]{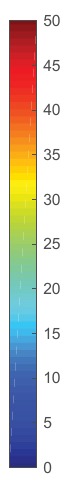}
\caption{Visual super-resolution results of the 18-th band and the reconstruction error maps of an example image from the Harvard dataset for different methods. The reconstruction error is obtained by computing the mean-square error between two spectrum vectors from the super-resolution result and the ground truth at each pixel. Best view on the screen.}
\label{fig:visual-harvard}
\vspace{-0.3cm}
\end{figure*}

\begin{figure*}[htbp]
\centering
\subfigure[NTIRE2018]{\includegraphics[height=1.0in, width=0.9in]{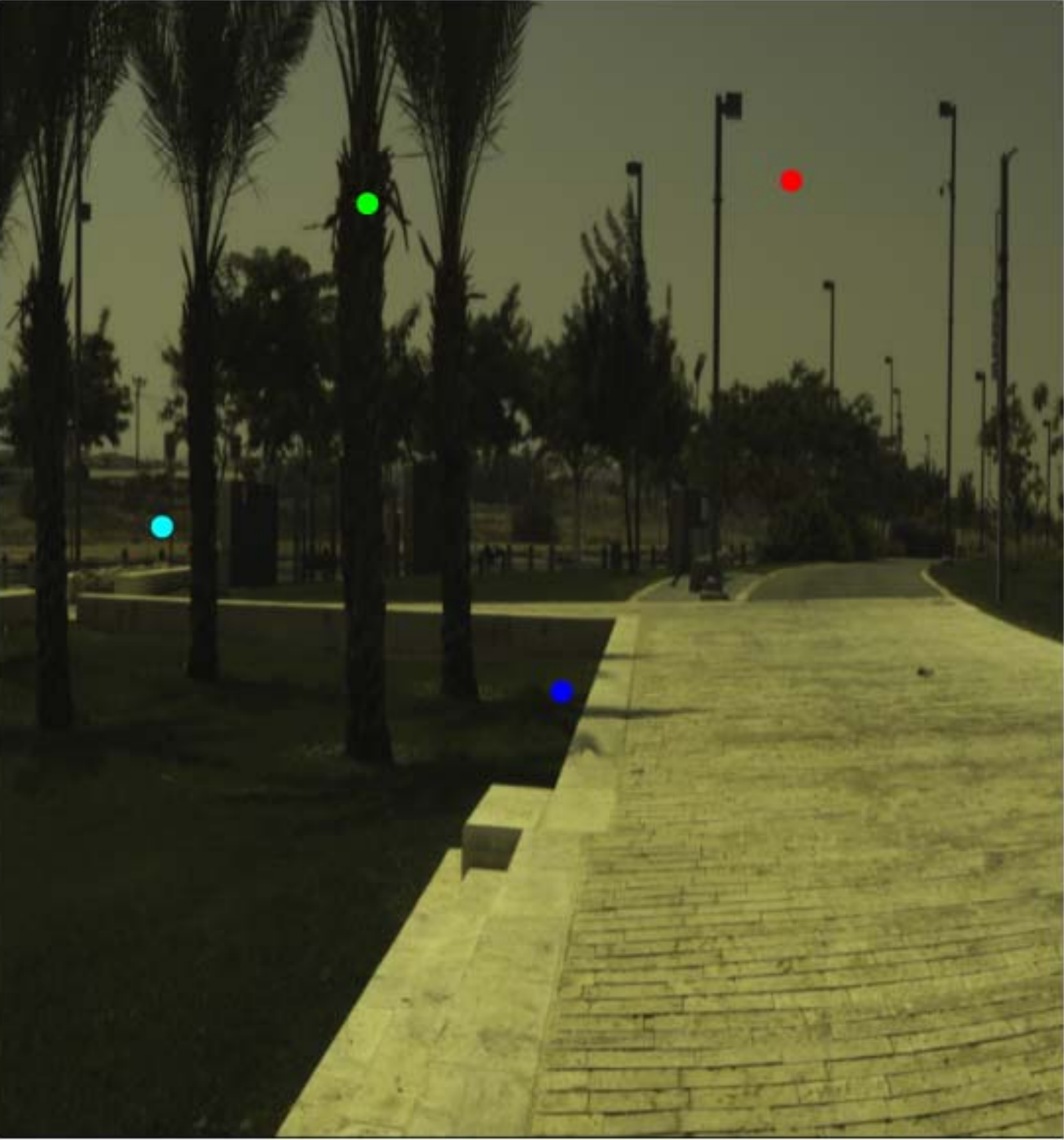}}
\hspace{-0.15cm}
\subfigure[Spectra]{\includegraphics[height=1.0in, width=1.3in]{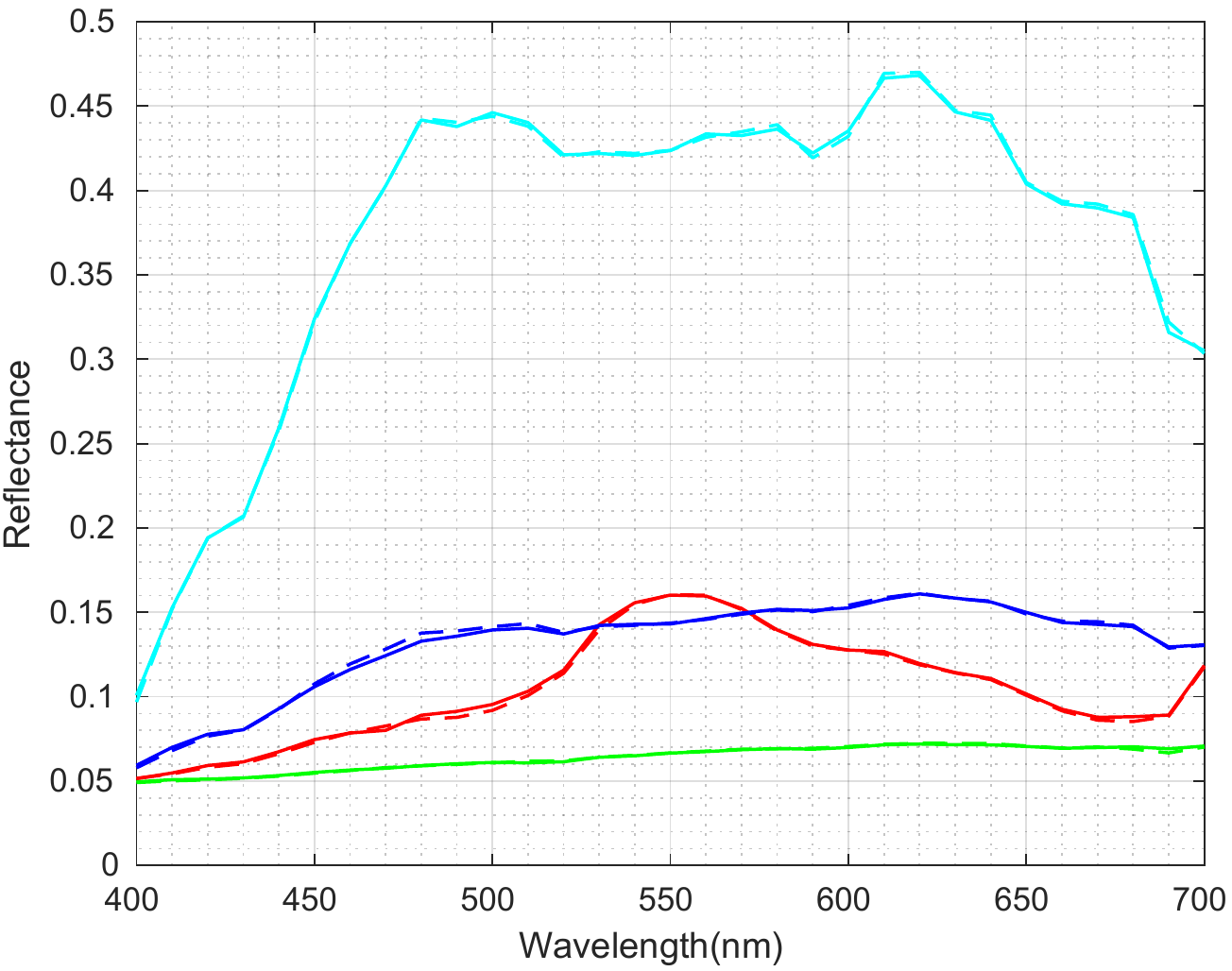}}
\hspace{-0.15cm}
\subfigure[CAVE]{\includegraphics[height=1.0in, width=0.9in]{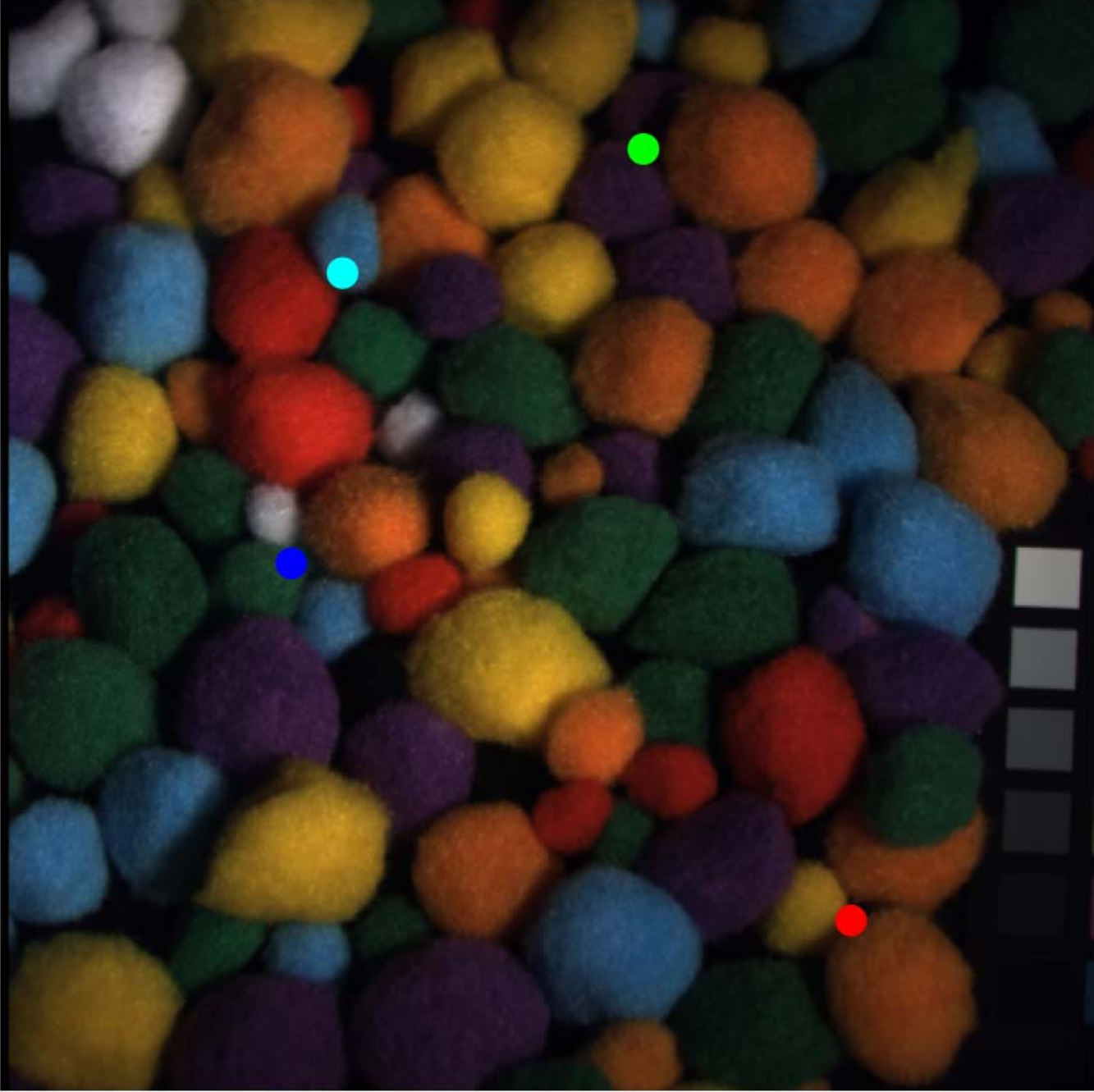}}
\hspace{-0.15cm}
\subfigure[Spectra]{\includegraphics[height=1.0in, width=1.3in]{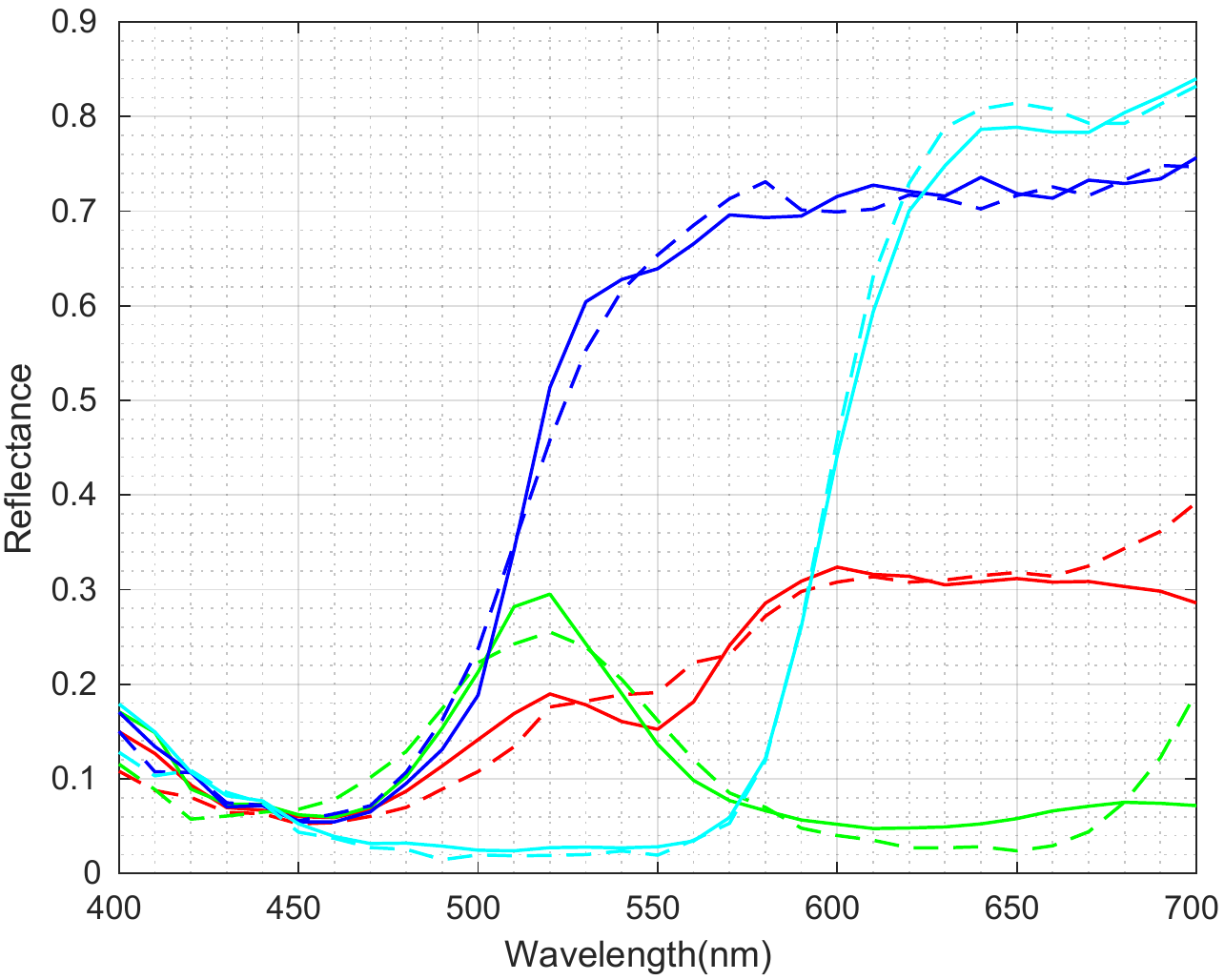}}
\hspace{-0.15cm}
\subfigure[Havard]{\includegraphics[height=1.0in, width=0.9in]{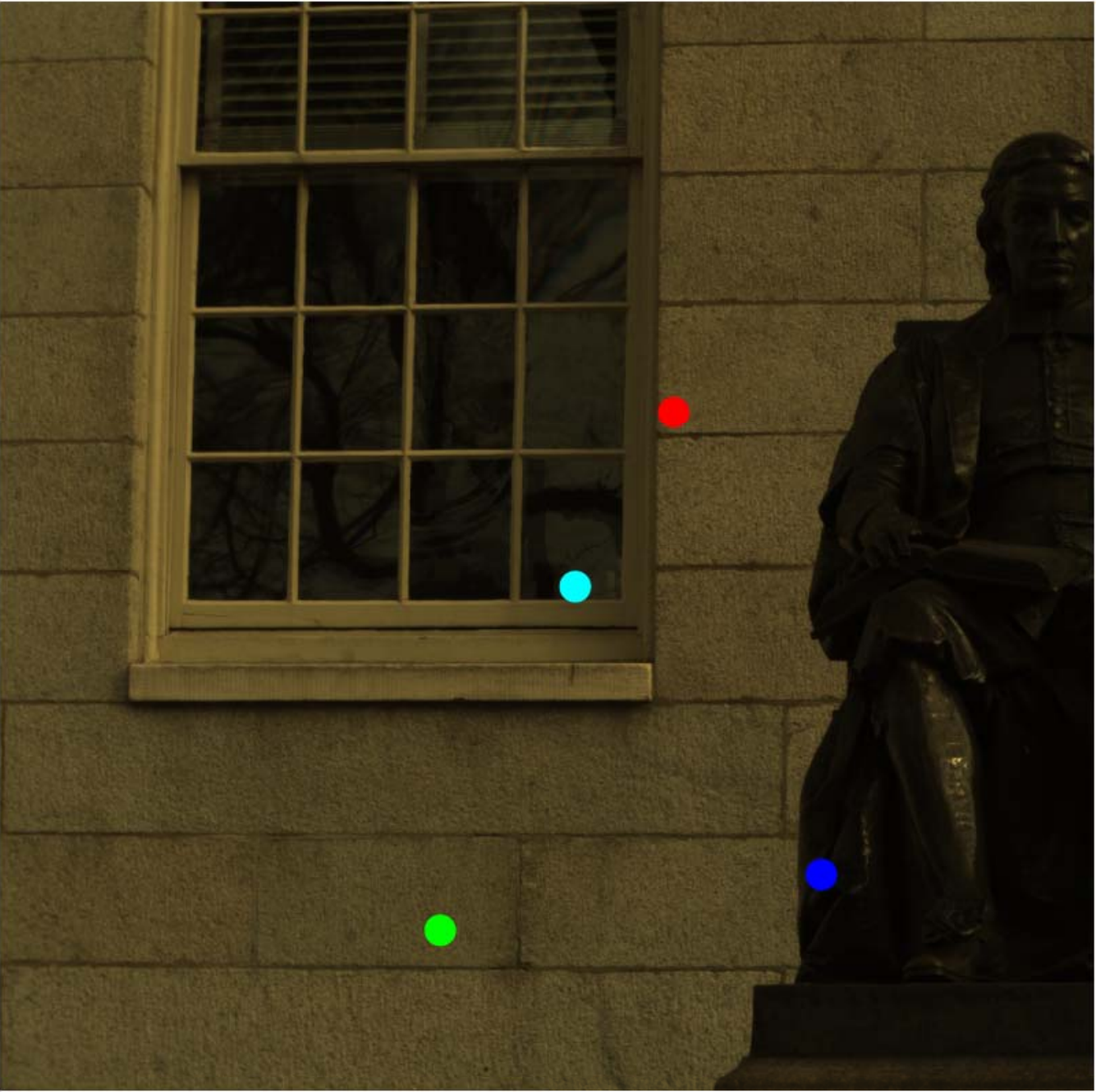}}
\hspace{-0.15cm}
\subfigure[Spectra]{\includegraphics[height=1.0in, width=1.3in]{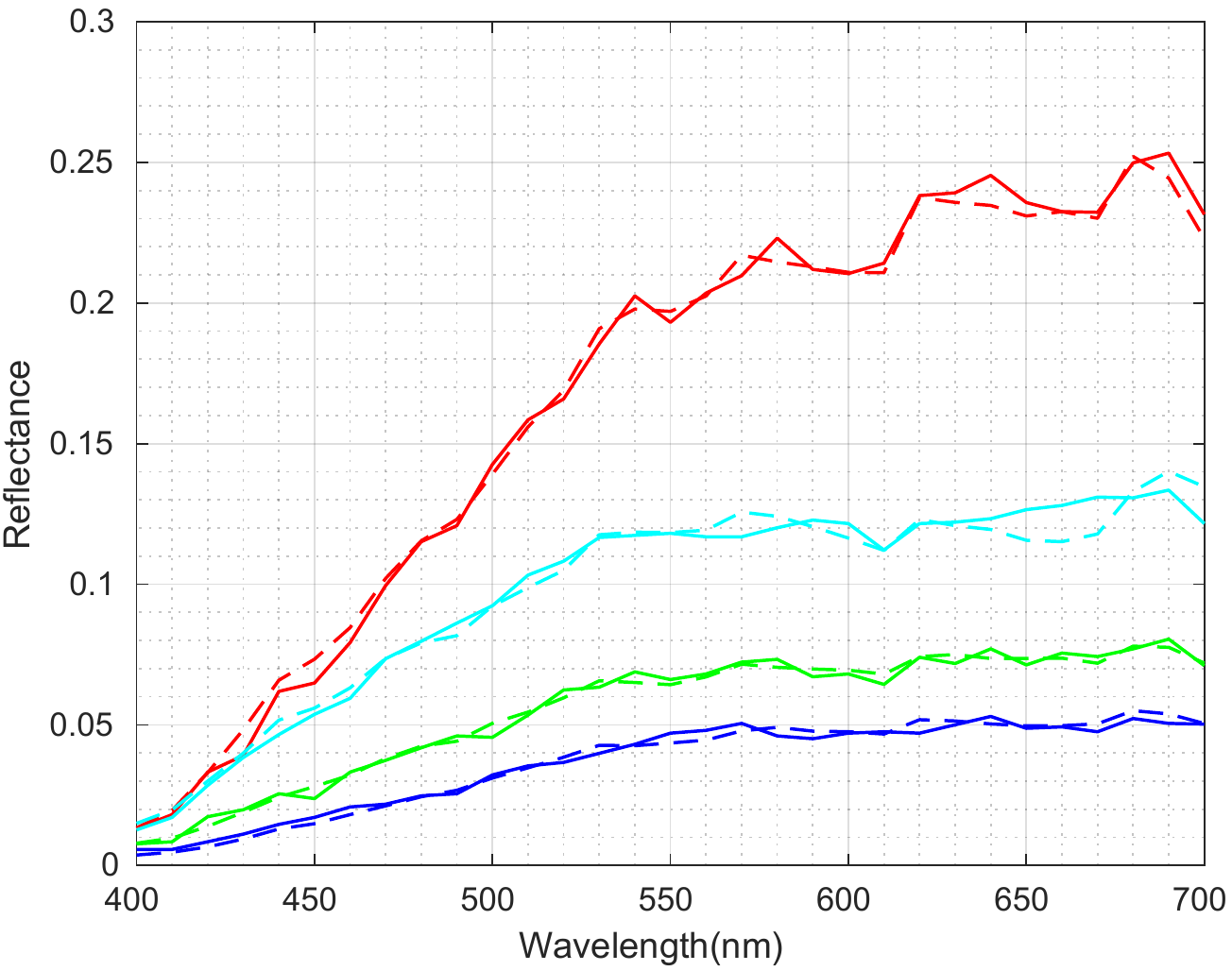}}
\caption{Recovered spectra form the super-resolution results of the proposed method on three example images chosen from three datasets. In each image, we select four different positions and plot the curves of the recovered spectra (\ie, denoted by dash lines) and the corresponding ground truth spectra (\ie, denoted by solid lines).}
\label{fig:spectrum}
\vspace{-0.3cm}
\end{figure*}

\begin{figure}[htbp]
\centering
\includegraphics[height=0.7in, width=0.75in]{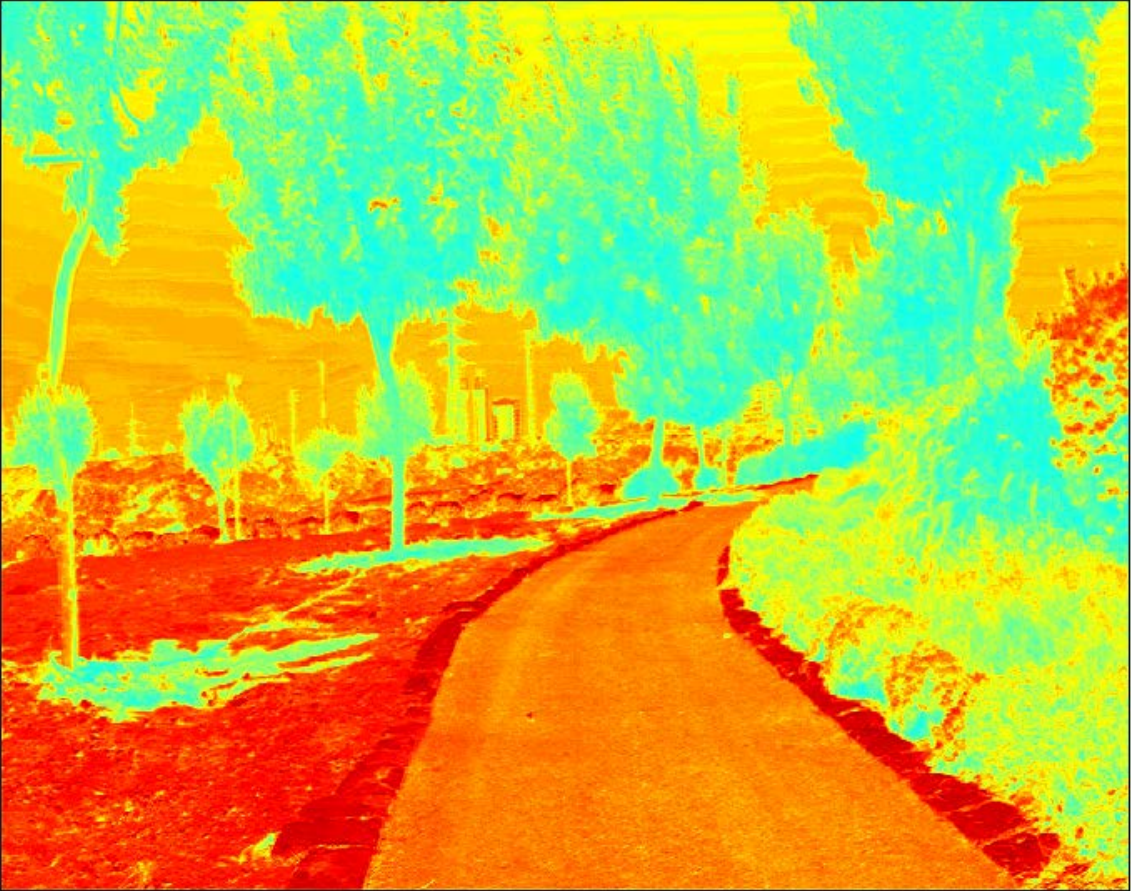}
\hspace{-0.15cm}
\includegraphics[height=0.7in, width=0.75in]{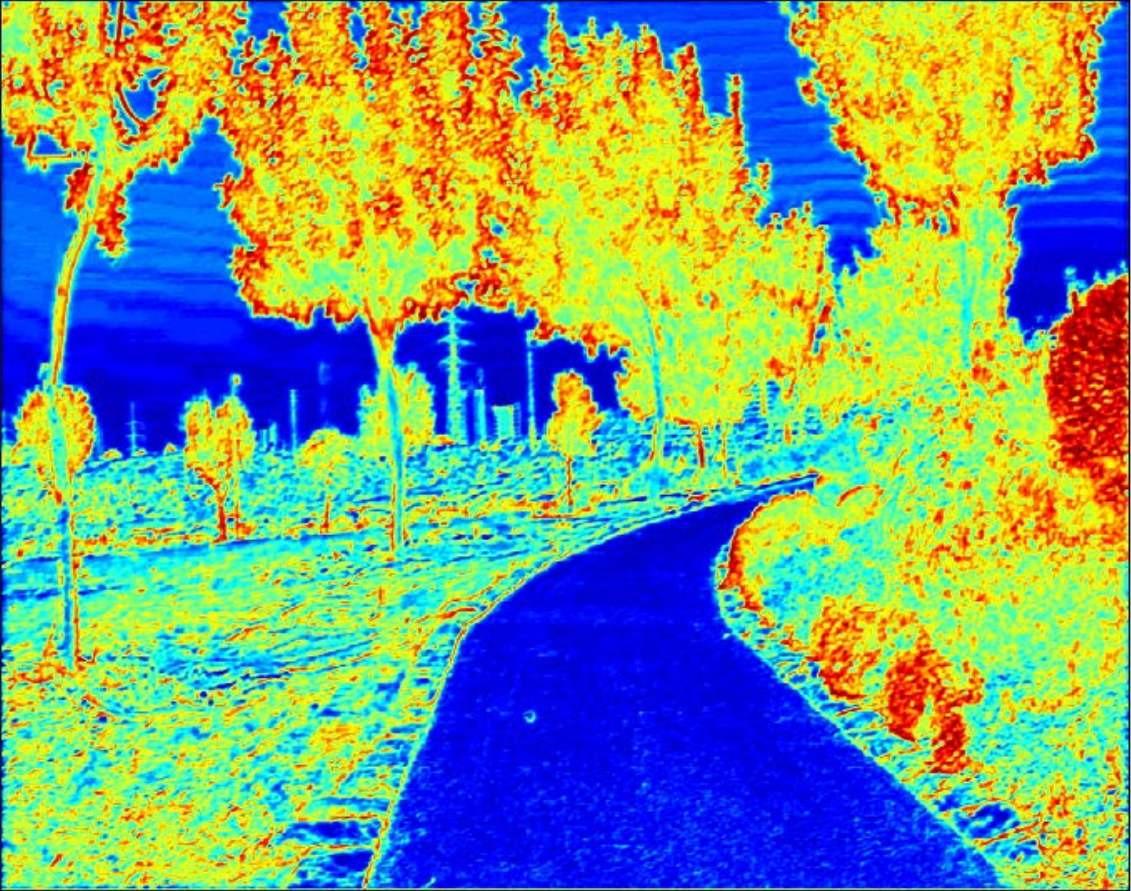}
\hspace{-0.15cm}
\includegraphics[height=0.7in, width=0.75in]{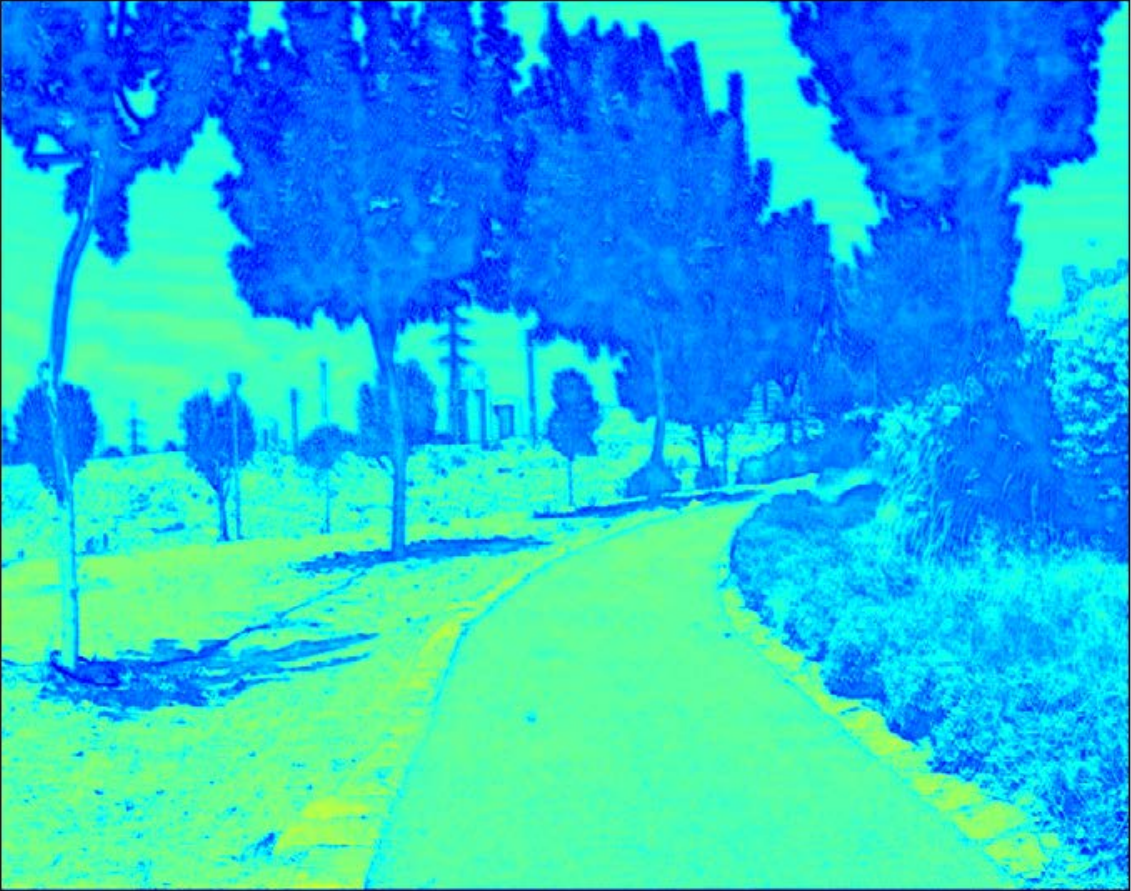}
\hspace{-0.15cm}
\includegraphics[height=0.7in, width=0.75in]{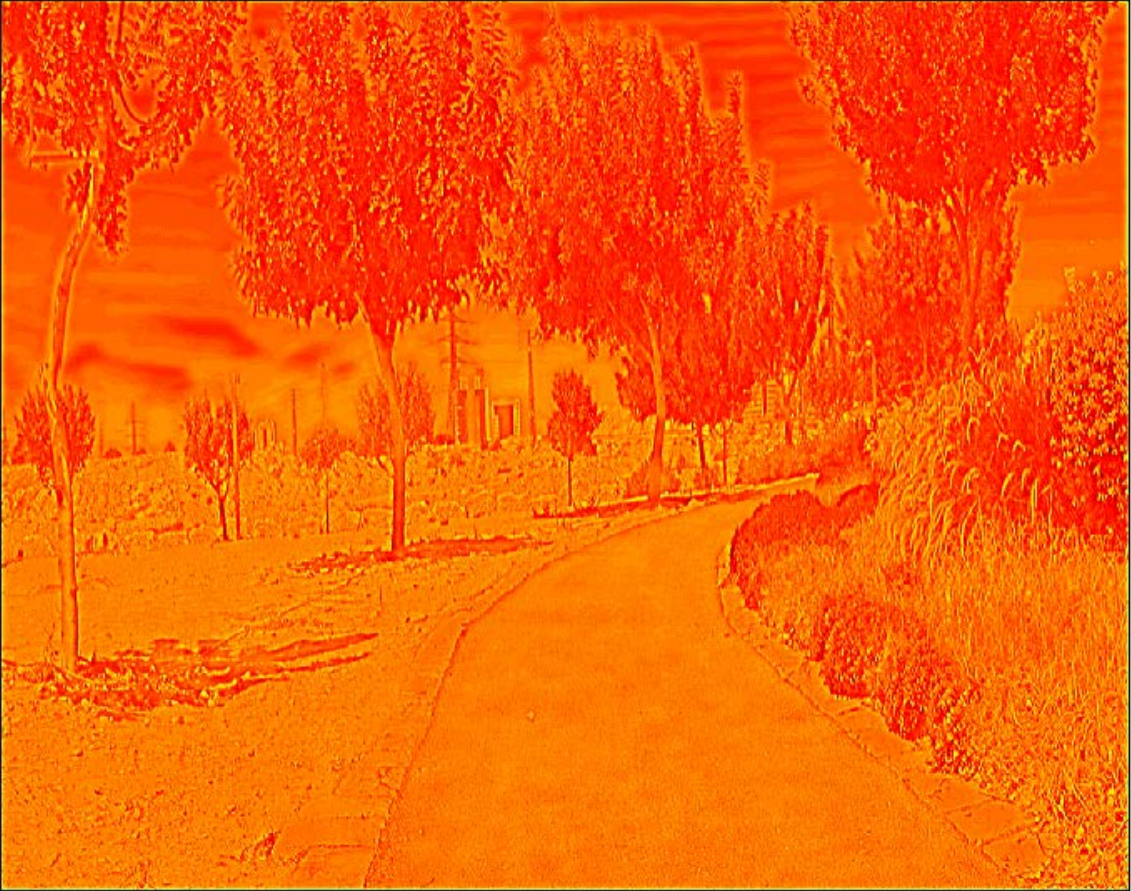}
\hspace{-0.15cm}
\includegraphics[height=0.7in, width=0.18in]{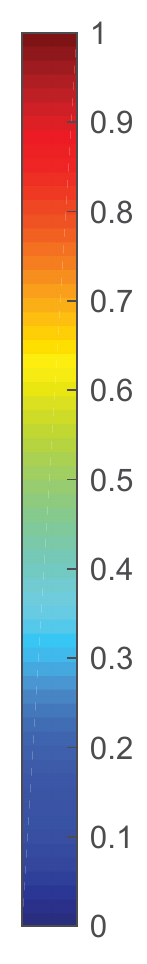}
\\
\includegraphics[height=0.7in, width=0.75in]{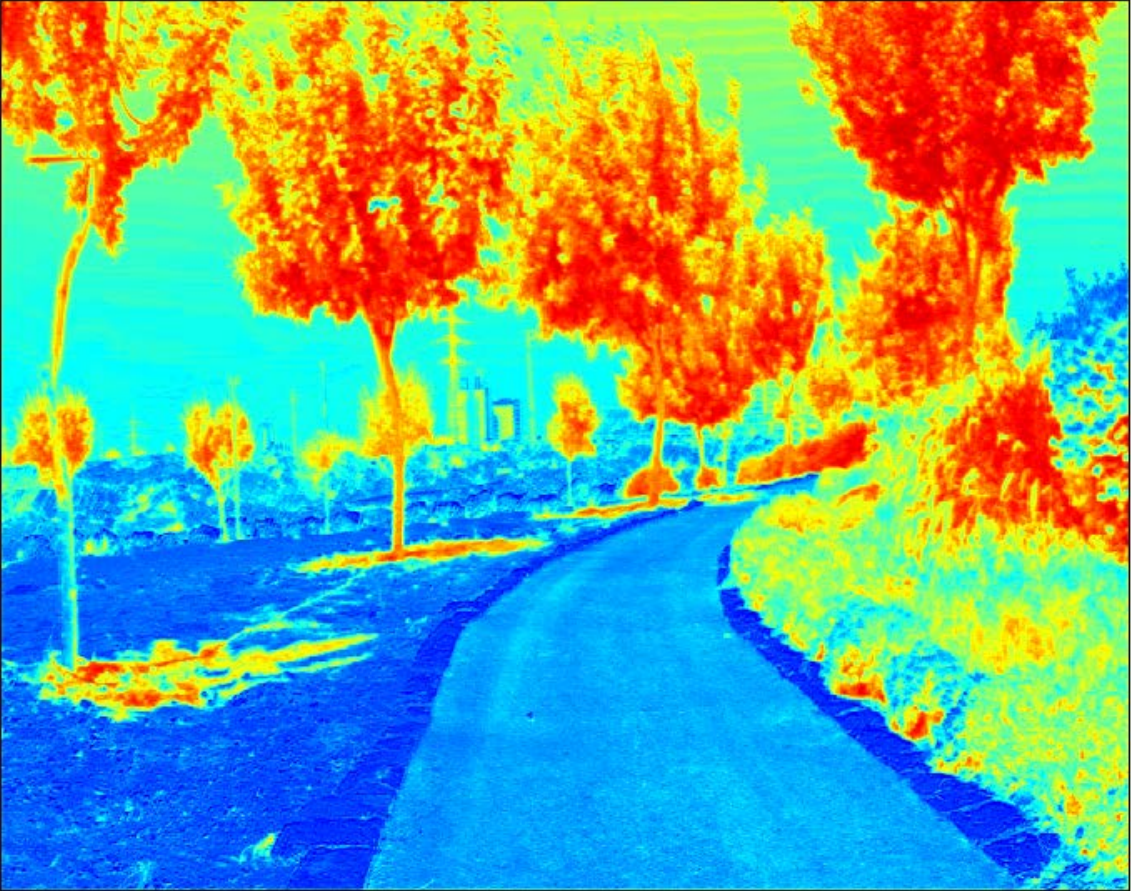}
\hspace{-0.15cm}
\includegraphics[height=0.7in, width=0.75in]{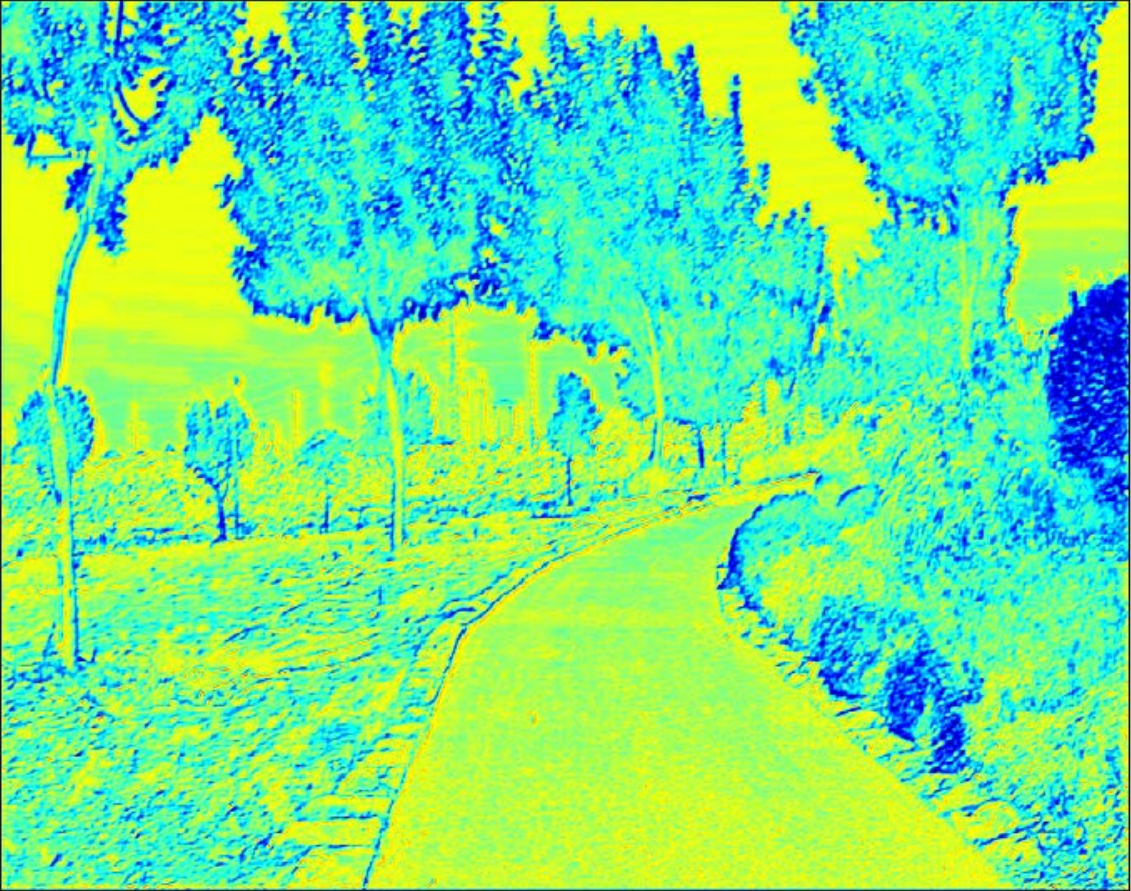}
\hspace{-0.15cm}
\includegraphics[height=0.7in, width=0.75in]{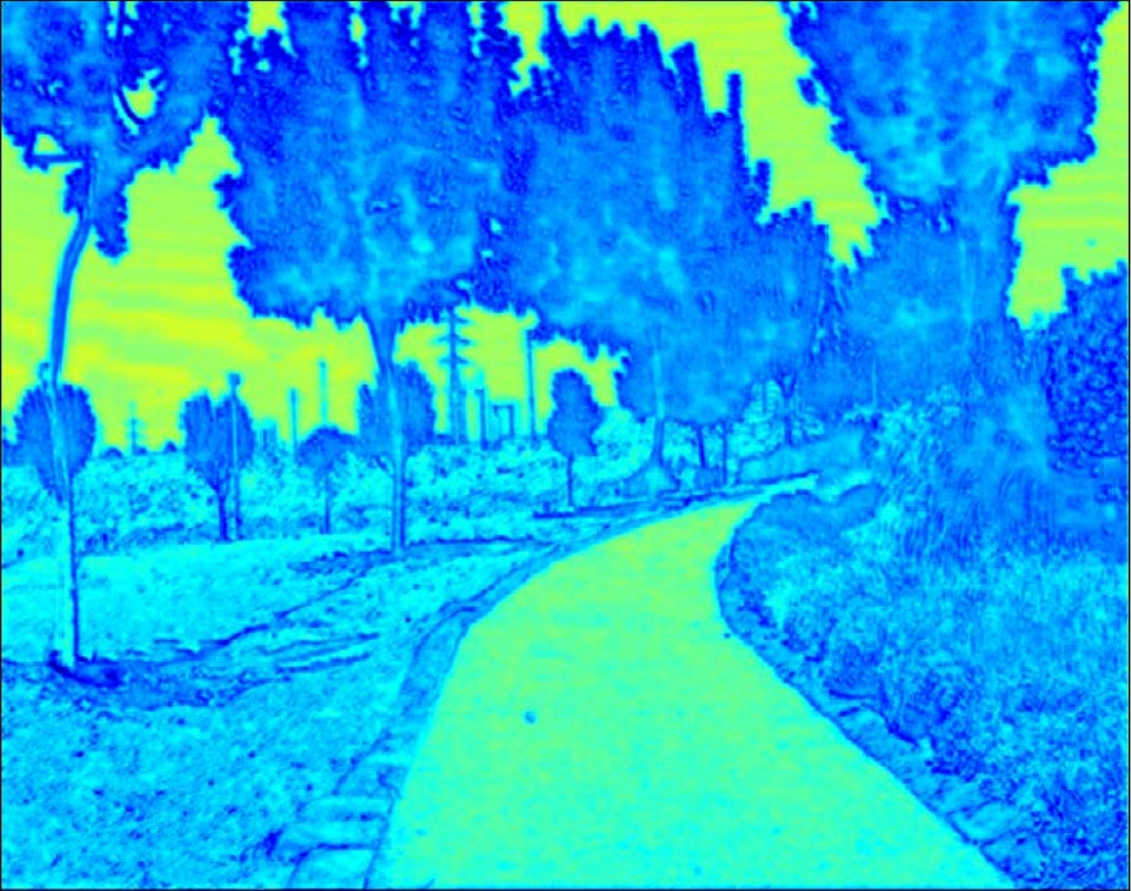}
\hspace{-0.15cm}
\includegraphics[height=0.7in, width=0.75in]{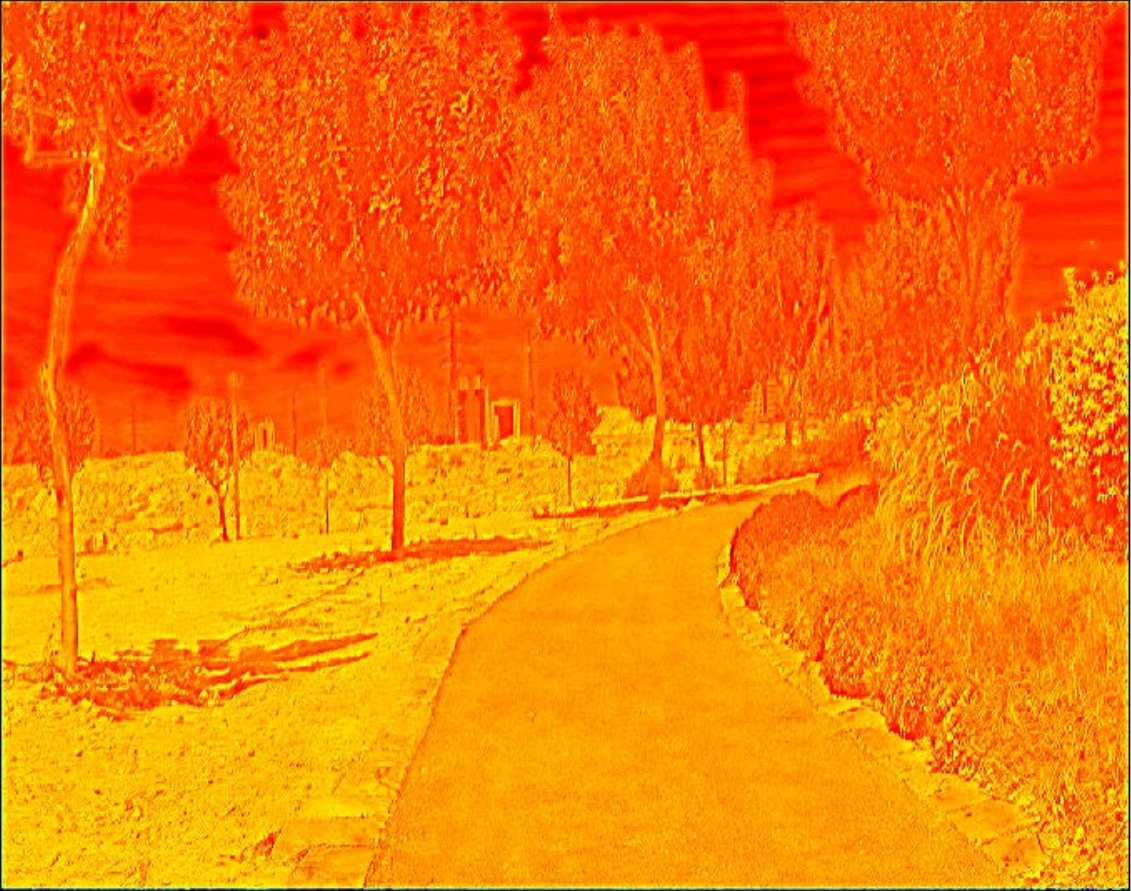}
\hspace{-0.15cm}
\includegraphics[height=0.7in, width=0.18in]{./mask/colorbar_1.pdf}
\\
\vspace{-0.18cm}
\subfigure[$\mathcal{F}^{1}$]{\includegraphics[height=0.7in, width=0.75in]{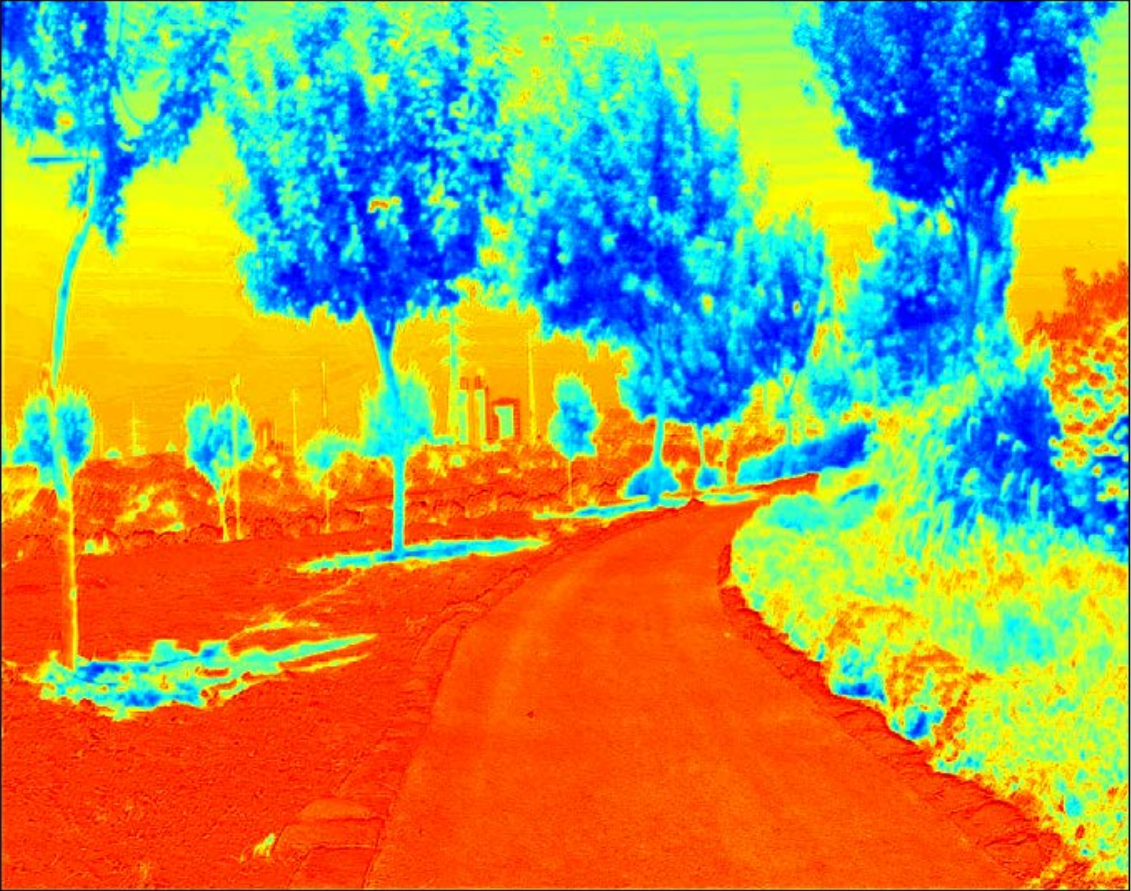}}
\hspace{-0.15cm}
\subfigure[$\mathcal{F}^{2}$]{\includegraphics[height=0.7in, width=0.75in]{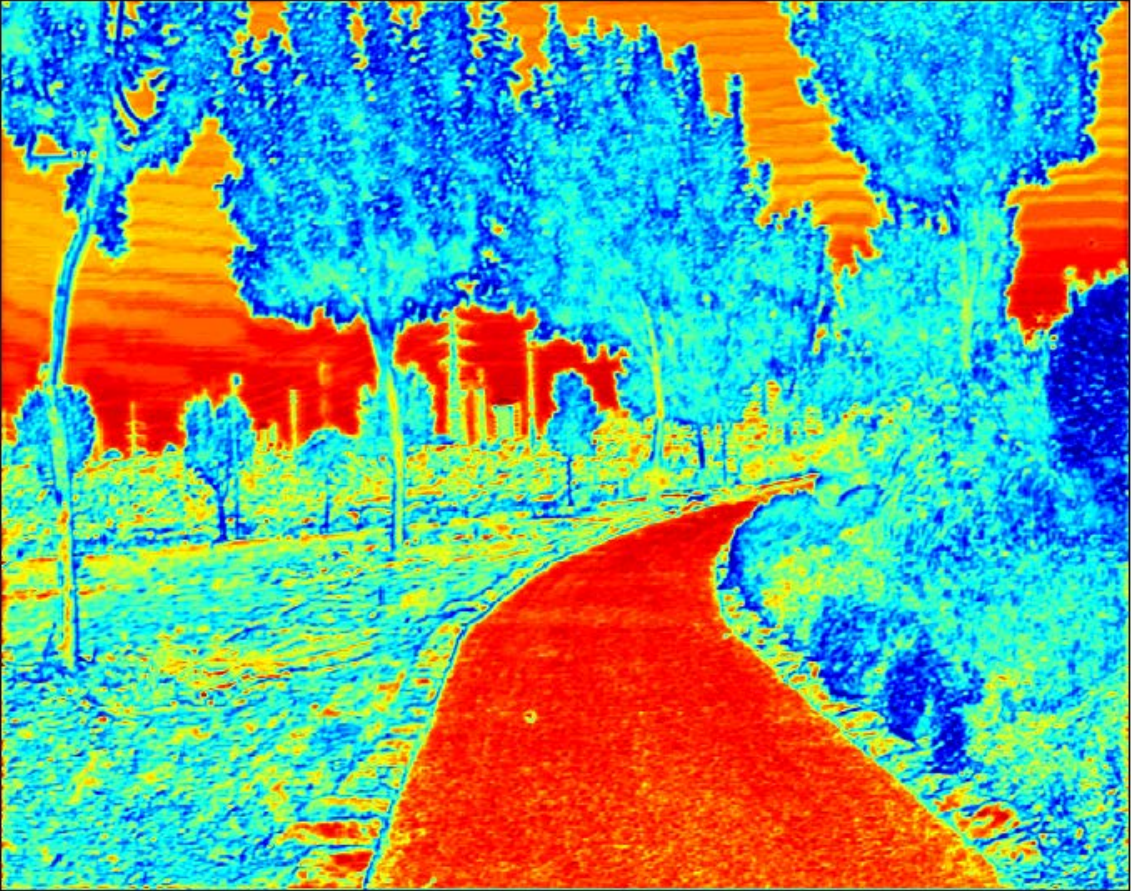}}
\hspace{-0.15cm}
\subfigure[$\mathcal{F}_{c}$]{\includegraphics[height=0.7in, width=0.75in]{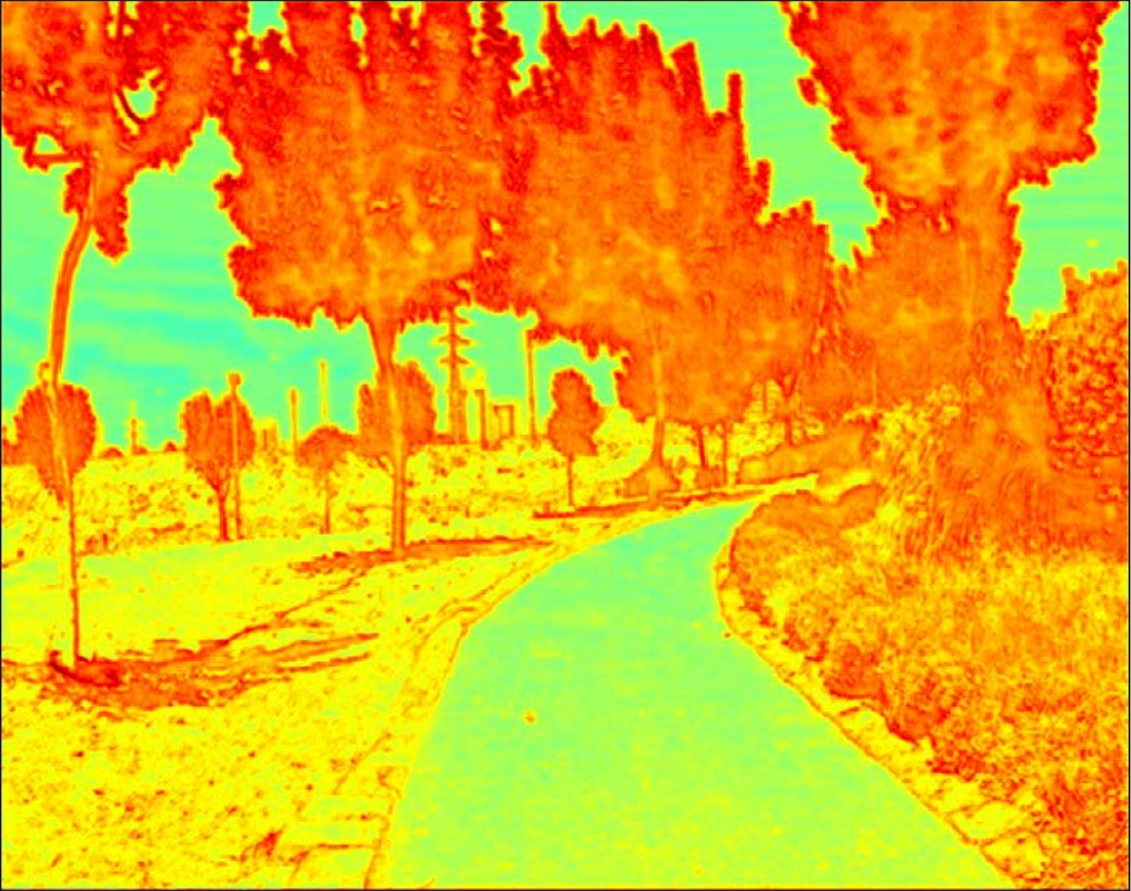}}
\hspace{-0.15cm}
\subfigure[$\mathcal{F}^{3}$]{\includegraphics[height=0.7in, width=0.75in]{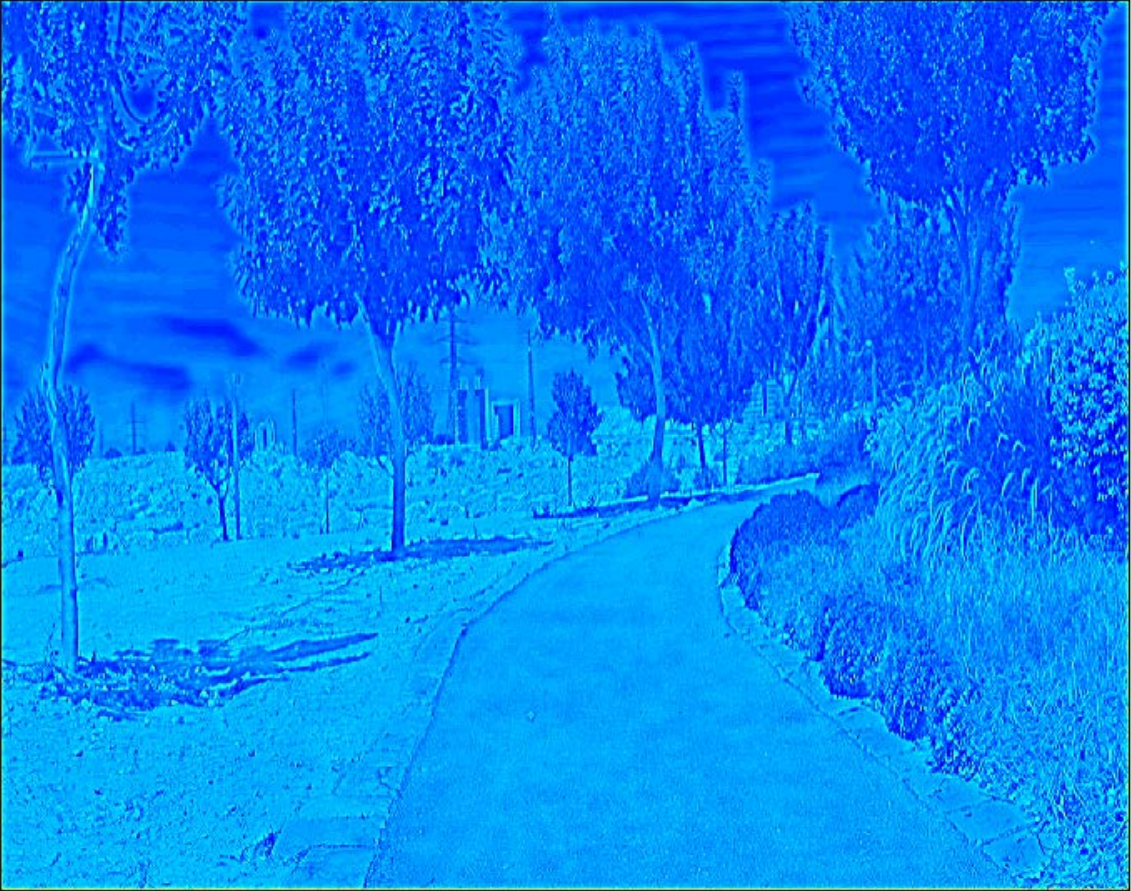}}
\hspace{-0.15cm}
\includegraphics[height=0.7in, width=0.18in]{./mask/colorbar_1.pdf}
\caption{Pixel-wise weights generated by the mixing function in different FM blocks of the proposed network. Figures in each column show the weight maps for three basis functions (from top to bottom: $f^u_1$, $f^u_2$ and $f^u_3$). For visualization convenience, we normalize each weight map into the range [0,1] using the inner maximum values and the minimum values.}
\label{fig:one-weight}
\vspace{-0.3cm}
\end{figure}


\begin{figure}[htbp]
\centering
\includegraphics[height=0.7in, width=0.75in]{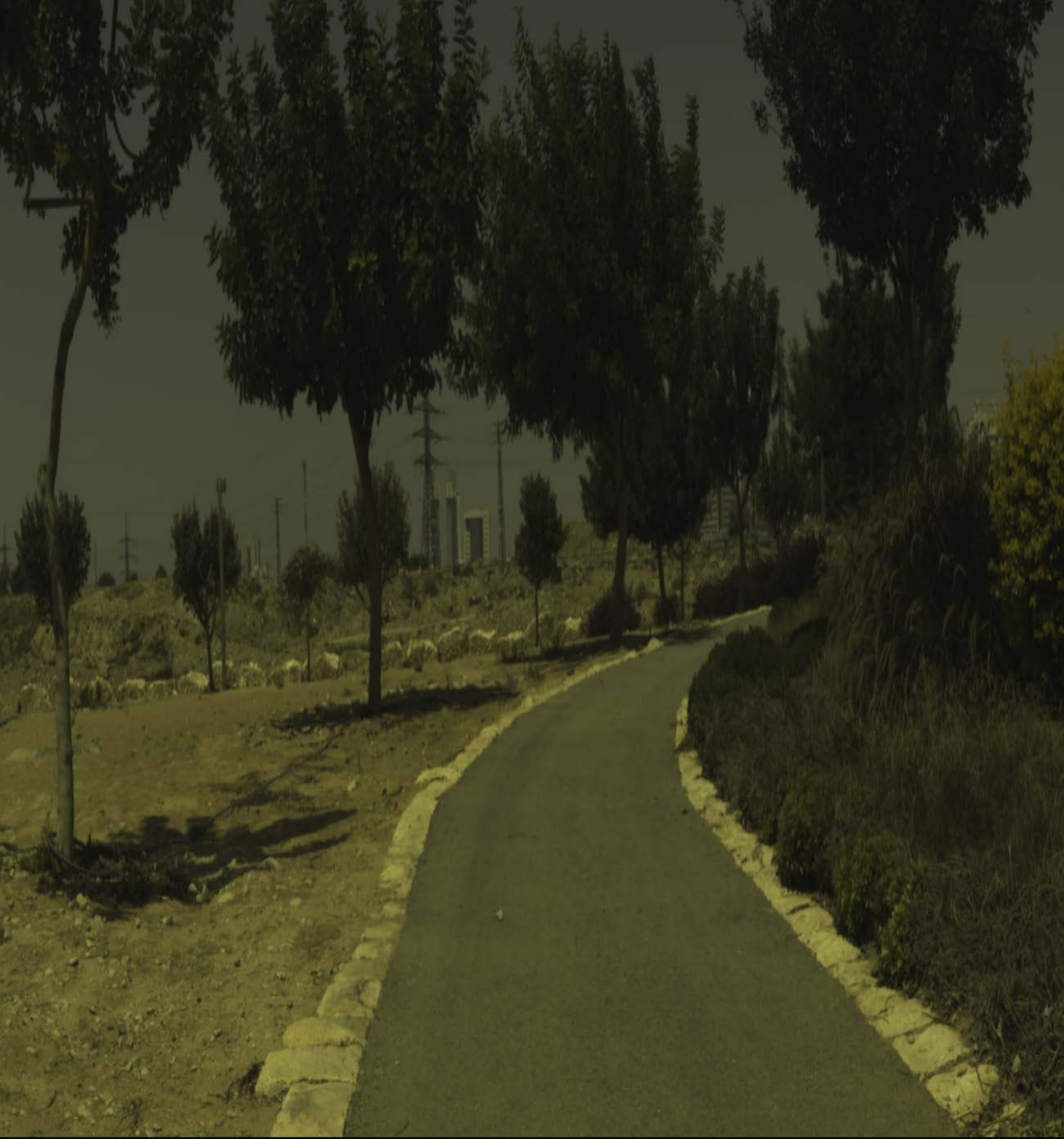}
\hspace{-0.15cm}
\includegraphics[height=0.7in, width=0.75in]{./mask/2_1.pdf}
\hspace{-0.15cm}
\includegraphics[height=0.7in, width=0.75in]{./mask/2_2.pdf}
\hspace{-0.15cm}
\includegraphics[height=0.7in, width=0.75in]{./mask/2_3.pdf}
\hspace{-0.15cm}
\includegraphics[height=0.7in, width=0.18in]{./mask/colorbar_1.pdf}
\\
\includegraphics[height=0.7in, width=0.75in]{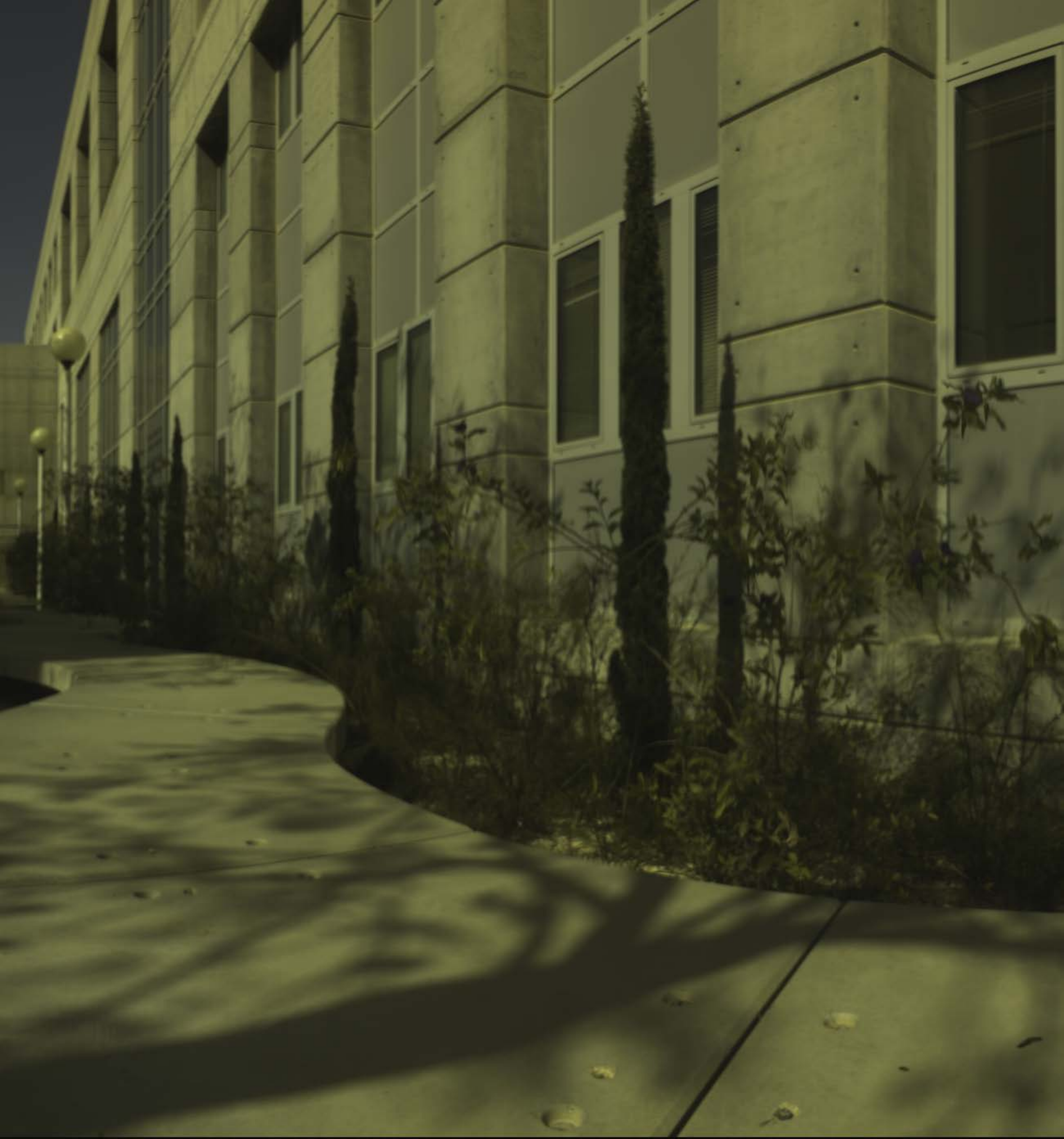}
\hspace{-0.15cm}
\includegraphics[height=0.7in, width=0.75in]{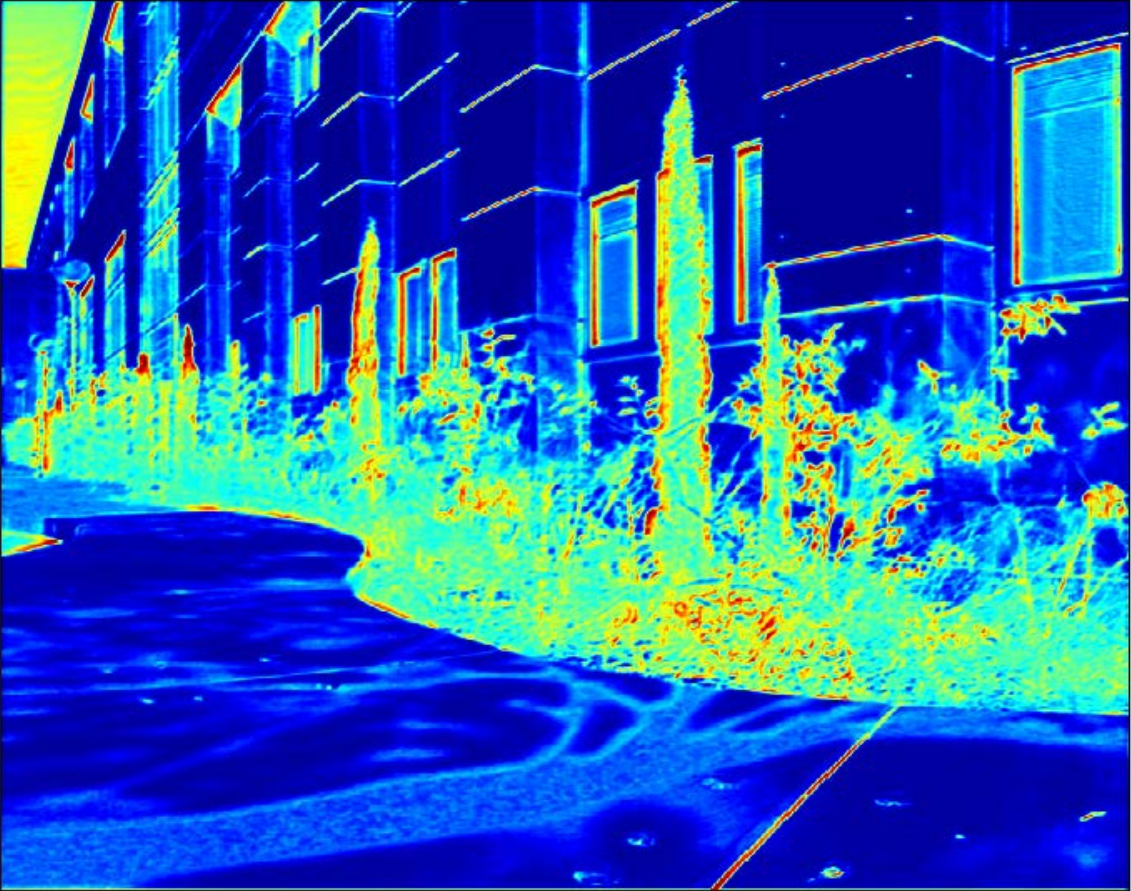}
\hspace{-0.15cm}
\includegraphics[height=0.7in, width=0.75in]{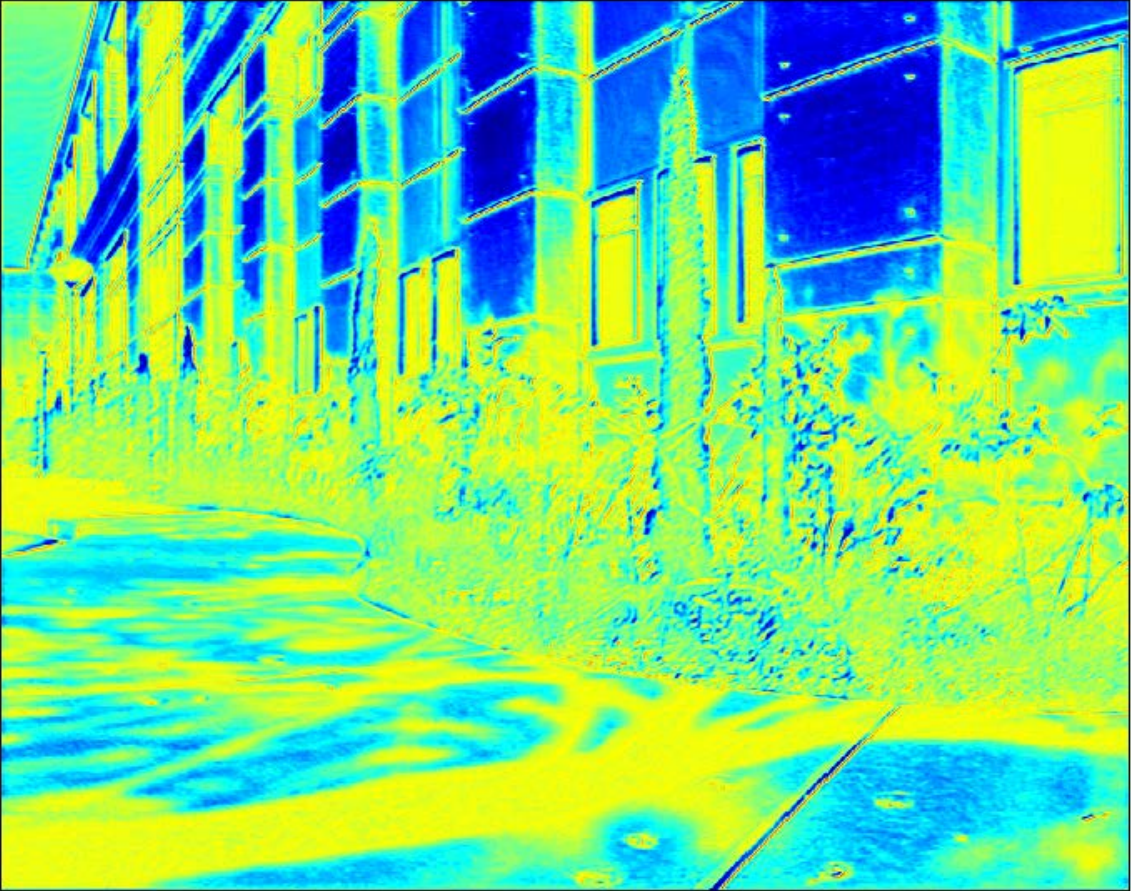}
\hspace{-0.15cm}
\includegraphics[height=0.7in, width=0.75in]{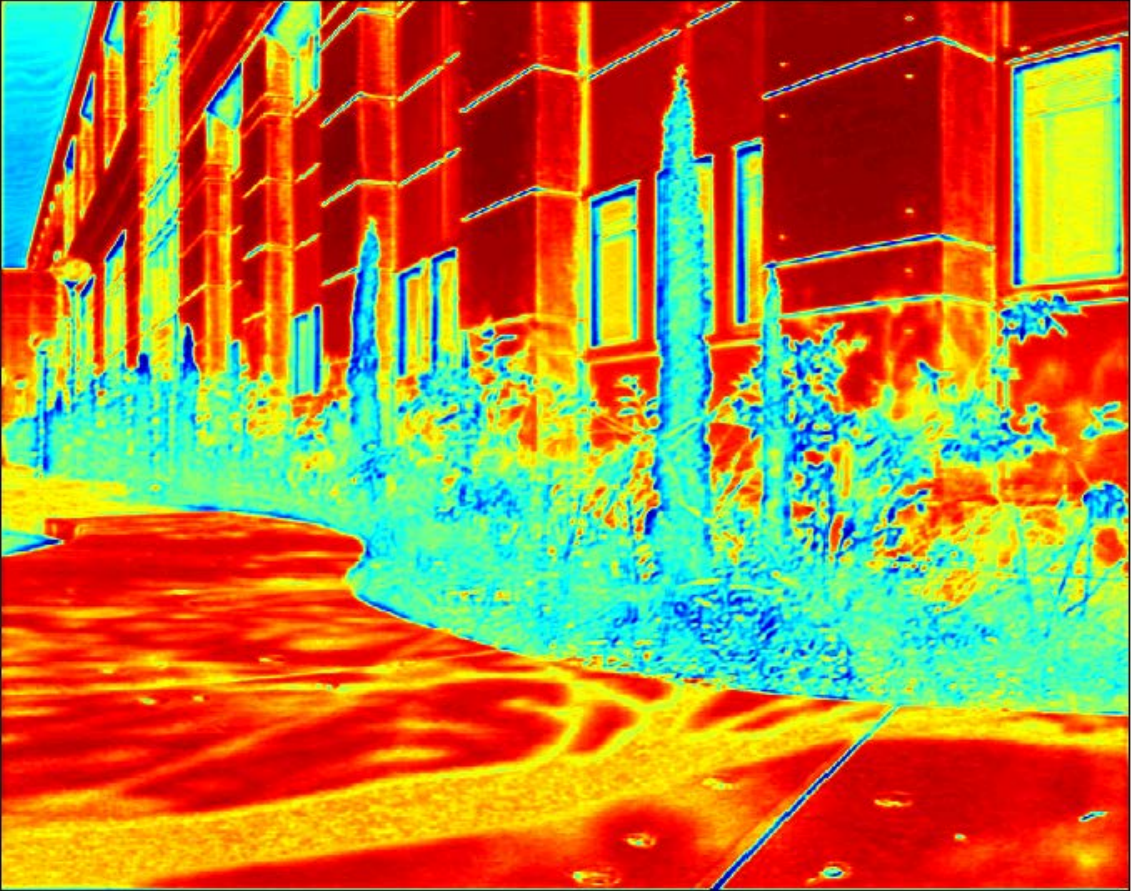}
\hspace{-0.15cm}
\includegraphics[height=0.7in, width=0.18in]{./mask/colorbar_1.pdf}
\\
\vspace{-0.18cm}
\subfigure[RGB image]{\includegraphics[height=0.7in, width=0.75in]{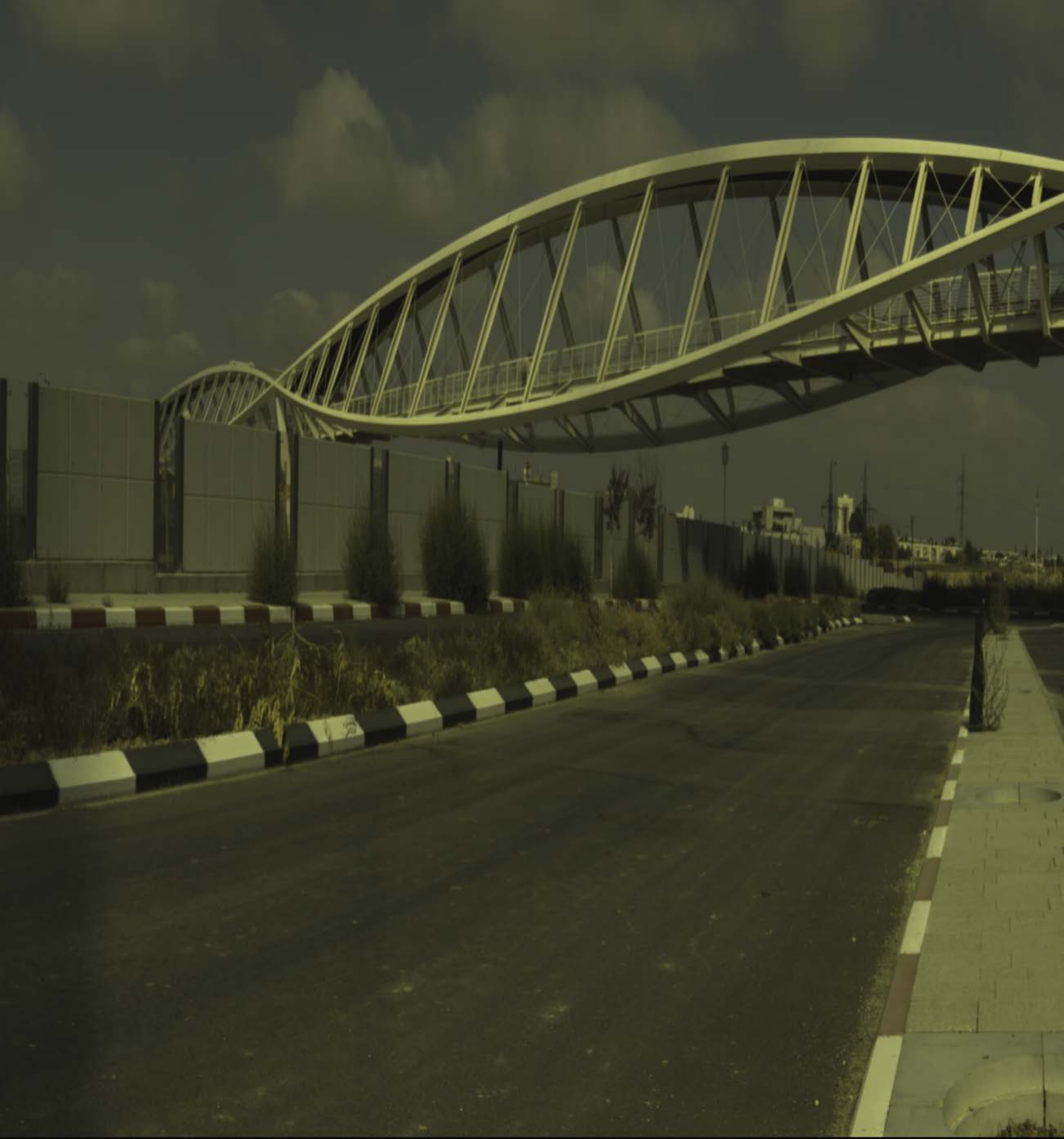}}
\hspace{-0.15cm}
\subfigure[$f^2_1$]{\includegraphics[height=0.7in, width=0.75in]{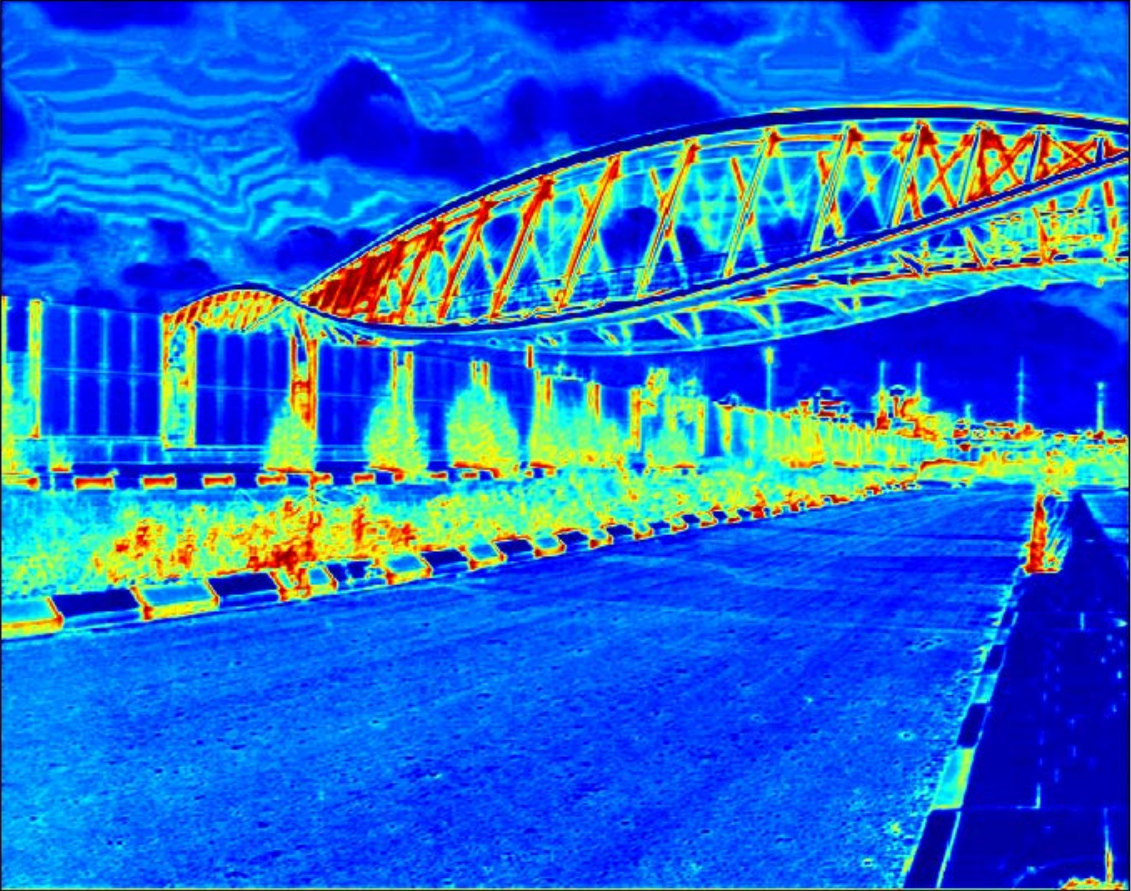}}
\hspace{-0.15cm}
\subfigure[$f^2_2$]{\includegraphics[height=0.7in, width=0.75in]{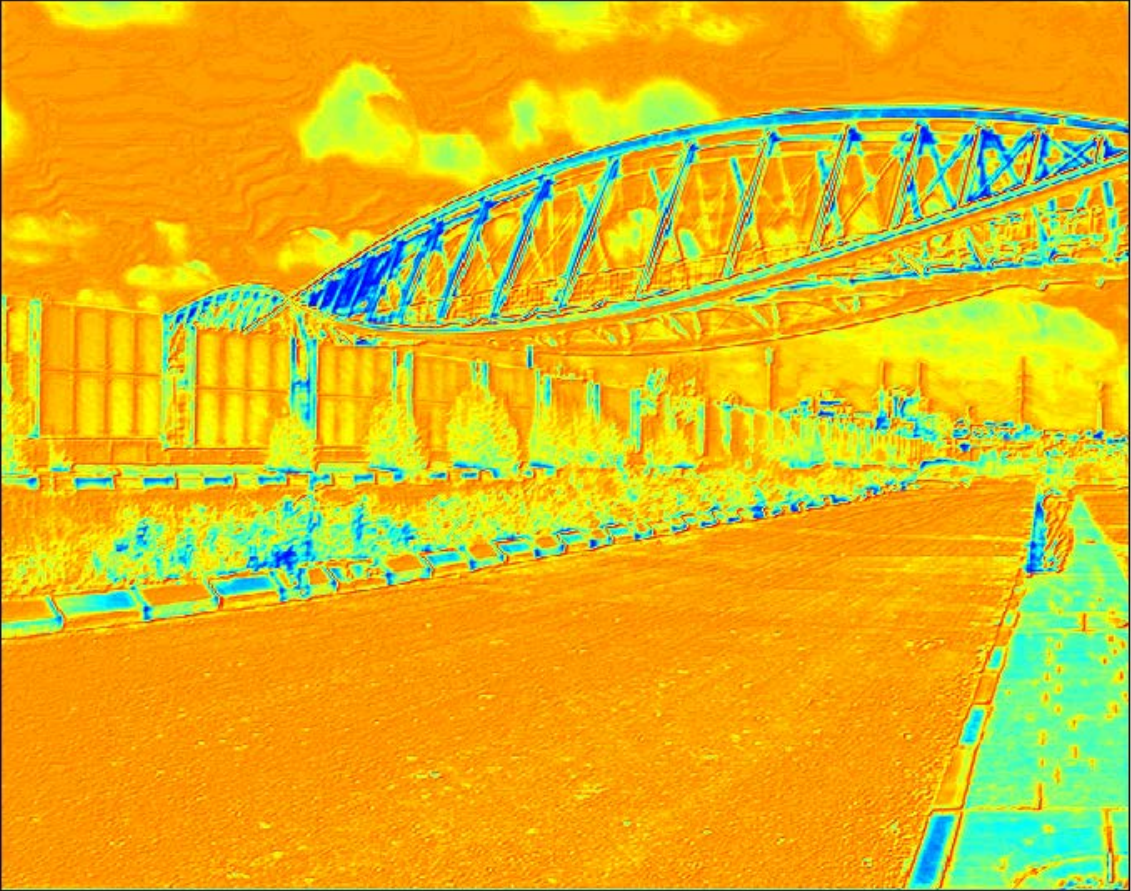}}
\hspace{-0.15cm}
\subfigure[$f^2_3$]{\includegraphics[height=0.7in, width=0.75in]{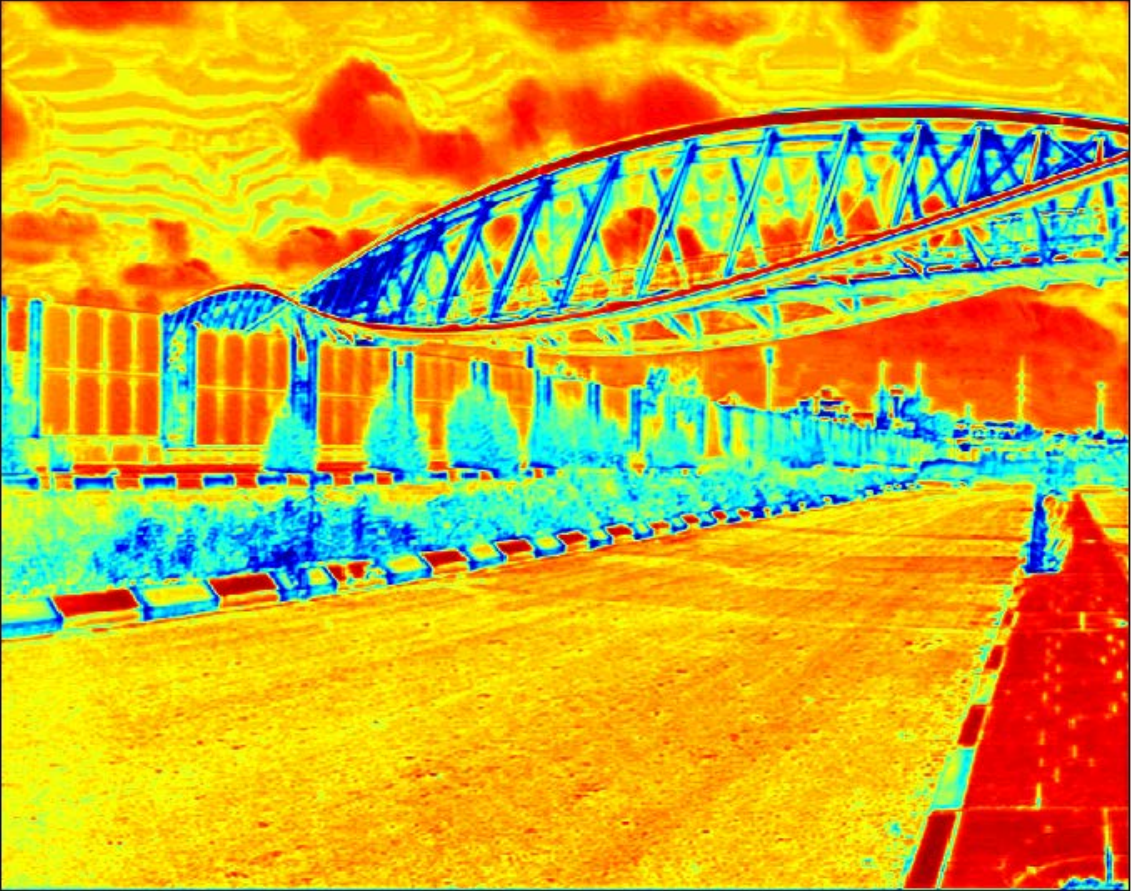}}
\hspace{-0.15cm}
\includegraphics[height=0.7in, width=0.18in]{./mask/colorbar_1.pdf}
\caption{Pixel-wise weights generated by the mixing function in the FM block $\mathcal{F}^{2}$ on different images. In each row, figures from left to right denote the input RGB image and three generated weight maps corresponding to the basis functions $f^2_1$, $f^2_2$ and $f^2_3$, respectively. For visualization convenience, we normalize each weight map into the range [0,1] using the inner maximum values and the minimum values.}
\label{fig:multiple-weight}
\vspace{-0.3cm}
\end{figure}


\subsection{Performance Evaluation}
\paragraph{Performance comparison} Under the same experimental settings, we evaluate all those methods on the testing set from each benchmark dataset. Their numerical results are reported in Table~\ref{table:numerical}. It can be seen that these DCNN based comparison methods often produce more accurate results than the interpolation or the sparsity induced SSR method. For example, on the NTIRE2018 dataset, the RMSE of the Aitor and HSCNN+ are less than 2.0 while that of the BI and Arad are higher than 4.0. Nevertheless, the proposed method obviously outperforms these DCNN based competitors. For example, compared with the state-of-the-art HSCNN+, the proposed method reduces the RMSE by 0.43 and improves the PSNR by 0.67db on the CAVE dataset. On the NTIRE2018 dataset, the decrease on RMSE is even up to 0.52 and the improvement on PSNR is up to 3.19db. This profits from the ability of the proposed method in adaptively determining the receptive field size and the mapping function for each pixel. With such an ability, the proposed method is able to handle each pixel more flexibly. Moreover, since various mapping functions can be approximated by the mixture of the learned basis functions, the proposed method can better generalize to the unknown pixels. 

In addition, as shown in Table~\ref{table:numerical}, the proposed method also performs better than two baselines, \ie, DCNN and MCNet. For example, on the NTIRE2018 dataset, the PSNR obtained by the proposed method is higher than that of DCNN by 1.89db and higher than that of MCNet by 0.86. Since the only difference between the proposed method and DCNN is the discrepancy between the convolutional block and the proposed FM block, the superiority of the proposed method demonstrates that the proposed FM block is much powerful than the convolutional block for SSR. Similarly, the advantage of the proposed method over MCNet clarifies that the proposed network architecture is more effective than the multi-column architecture in SSR.

To further clarify the above conclusions, we plot some visual super-resolution results of different methods on three datasets in Figure~\ref{fig:visual-ntr}, Figure~\ref{fig:visual-cave} and Figure~\ref{fig:visual-harvard}. As can be seen, the super-resolution results of the proposed method have more details and show less reconstruction error than other competitors. In addition, we also sketch the recovered spectrum curves of the proposed method in Figure~\ref{fig:spectrum}. It can be seen that the spectra produced by the proposed method are very close to the ground truth.
\vspace{-0.4cm}
\paragraph{Pixel-wise mixing weights} In this study, we mix the outputs of the basis functions with pixel-wise weights to adaptively learn the pixel-wise mapping. To validate that the proposed method can effectively produce the pixel-wise weights as expected, we choose an example image from the NTIRE2018 and visualize the produced pixel-wise weights in each FM block, as shown in Figure~\ref{fig:one-weight}. We can find that, i) pixels from different categories or spatial positions are often given different weights. For example, in the second weight map generated by $\mathcal{F}^1$, the weights for the pixels from 'road' are obviously smaller than that for the pixels from 'tree'. ii) Pixels from the same category are pone to be given similar weights. For example, pixels from 'road' are given similar weights in each weight map in Figure~\ref{fig:one-weight} (a)(b). To further clarify these two aspects of observations, we visualize the weight maps of some other images generated by the FM block $\mathcal{F}^2$ in Figure~\ref{fig:multiple-weight}, where similar phenomenon can be observed. iii) In the intermediate FM blocks (\ie, $\mathcal{F}^1$ and $\mathcal{F}^2$ in Figure~\ref{fig:one-weight}), the high level block (\eg, $\mathcal{F}^2$) can distinguish finer difference between pixels than the low level block (\eg, $\mathcal{F}^1$), viz., only highly similar pixels will be assigned to similar weights. iv) Due to being forced to match the output, in the weight maps generated by the ultimate output block $\mathcal{F}^3$, the weight difference between pixels from various categories is not as obvious as that in previous FM block (\eg, $\mathcal{F}^1$ and $\mathcal{F}^1$), as shown in Figure~\ref{fig:one-weight}(a)(b)(d).

According to the above observations, we can conclude that the proposed network can effectively generate the pixel-wise mixing weights and thus is able to pixel-wisely determine receptive field size and mapping function.

\subsection{Ablation study}
In this part, we carry out an ablation study on the NTIRE2018 dataset to demonstrate the effect of the different ingredients, the number of basis functions and the number of FM blocks on the proposed network.

\begin{figure}[htbp]
\centering
\includegraphics[height=1.4in, width=1.6in]{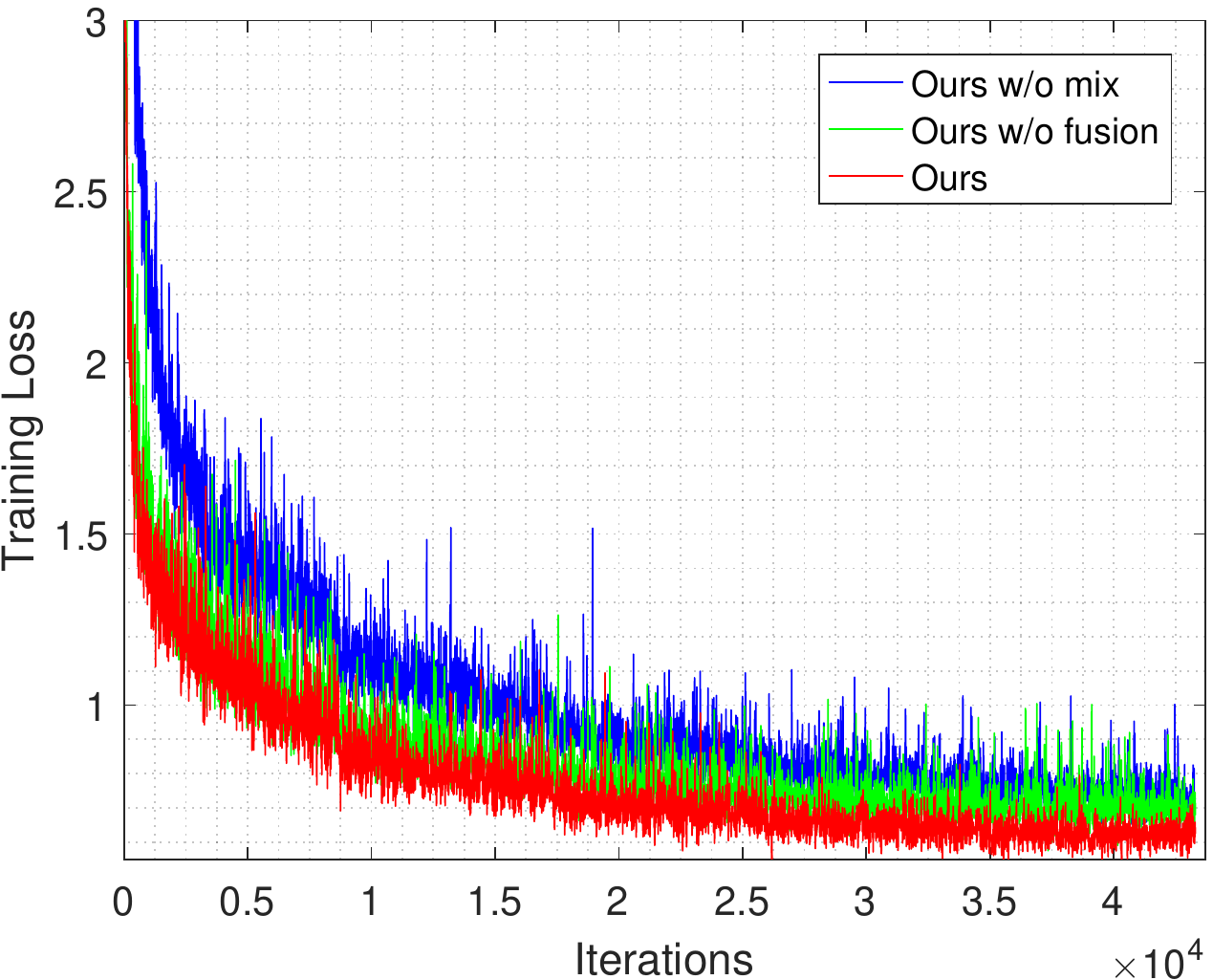}
\includegraphics[height=1.4in, width=1.6in]{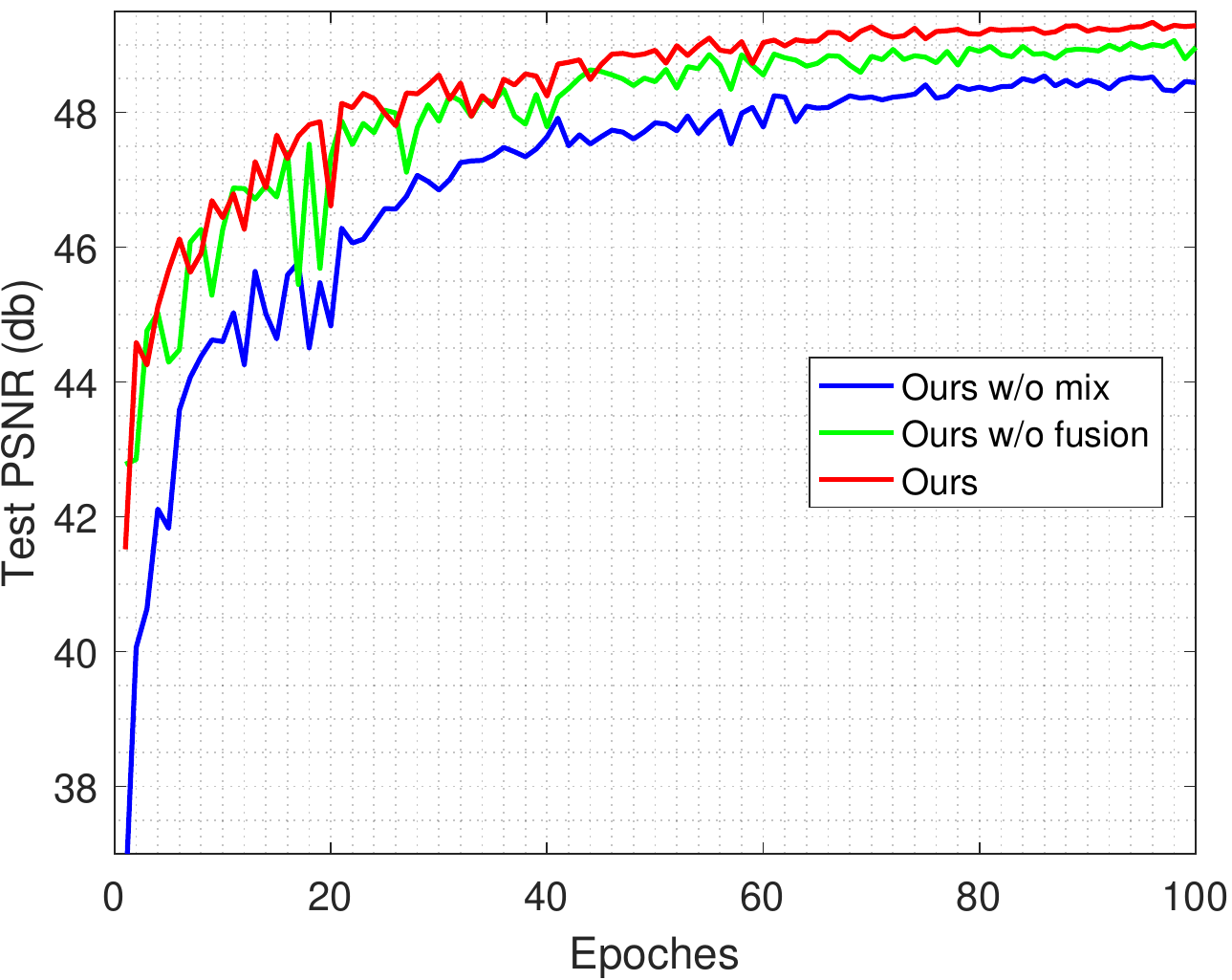}
\caption{Curves of training loss and test PSNR for the proposed method (\eg, 'Ours') and its two variants (\eg, 'Ours w/o mix', 'Our w/o fusion') during training on the NTIRE2018 dataset. (Ours w/o mix: without pixel-wise mixture; Ours w/o fusion: without intermediate feature fusion)}
\label{fig:curves}
\end{figure}

\begin{table}\small
\caption{Effect of the different ingredients (\ie, pixel-wise mixture \& intermediate feature fusion) in the proposed network.}
\label{table:ingradient}
\renewcommand{\arraystretch}{1.1}
\begin{center}
\begin{tabular}{l|c|c|c|c}
\hline
Methods & RMSE & PSNR & SAM & SSIM\\
\hline
Ours w/o mix & 1.10 & 48.44 & 1.16 & 0.9950\\
Ours w/o fusion & 1.05 & 48.97 & 1.09 & 0.9953\\
Ours & {\textbf{1.03}} & {\textbf{49.29}} & {\textbf{1.05}} & {\textbf{0.9955}}\\
\hline
\end{tabular}
\end{center}
\vspace{-0.3cm}
\end{table}

\begin{table}\small
\caption{Effect of the number $n$ of basis functions.}
\label{table:n_num}
\renewcommand{\arraystretch}{1.1}
\begin{center}
\begin{tabular}{l|c|c|c|c}
\hline
Methods & RMSE & PSNR & SAM & SSIM\\
\hline
Ours ($n=$1) & 1.47 & 45.82 & 1.57 & 0.9913\\
Ours ($n=$2) & 1.08 & 48.76 & 1.10 & 0.9952\\
Ours ($n=$3)& 1.03 & 49.29 & 1.05 & 0.9955\\
Ours ($n=$5) & {\textbf{0.98}} & {\textbf{49.87}} & {\textbf{1.00}} & {\textbf{0.9958}}\\
\hline
\end{tabular}
\end{center}
\vspace{-0.3cm}
\end{table}

\begin{table}\small
\caption{Effect of the number $p$ of FM blocks.}
\label{table:p_num}
\renewcommand{\arraystretch}{1.1}
\begin{center}
\begin{tabular}{l|c|c|c|c}
\hline
Methods & RMSE & PSNR & SAM & SSIM\\
\hline
Ours ($p=$2) & 1.05 & 48.95 & 1.09 & 0.9954\\
Ours ($p=$3) & 1.03 & 49.29 & 1.05 & 0.9955\\
Ours ($p=$4)& 1.05 & 49.42 & 1.05 & 0.9954\\
Ours ($p=$6) & {\textbf{1.00}} & {\textbf{49.59}} & {\textbf{1.02}} & {\textbf{0.9956}}\\
\hline
\end{tabular}
\end{center}
\vspace{-0.5cm}
\end{table} 

\vspace{-0.4cm}
\paragraph{Effect of Different Ingredients}
In the proposed FM network, there are two important ingredients, namely the pixel-wise mixture and the intermediate feature fusion. To demonstrate the effect of these two ingredients, we compare the proposed method with its two variants. One (\ie, 'Ours w/o mix') disables the pixel-wise mixture in the proposed network, which implies mixing the outputs of the basis functions with equal weights; while the other (\ie, 'Ours w/o fusion') disables the intermediate feature fusion, \ie, removing the skip connections as well as the FM block $\mathcal{F}_c$. We plot the training loss curves and the testing PSNR curves of these three methods in Figure~\ref{fig:curves}. As can be seen that the proposed method obtains the smallest training loss and the highest testing PSNR. More numerical results are reported in Table~\ref{table:ingradient}. It can be seen that the proposed method still obviously outperforms these two variants. This demonstrate that both the pixel-wise mixture and the intermediate feature fusion are crucial for the proposed network.
\vspace{-0.4cm}
\paragraph{Effect of the Number of Basis Functions}
In the above experiments, we fix the number of basis functions as $n=3$ in each FM block. Intuitively, increasing $n$ will enlarge the expressive capacity of the basis fictions and thus lead to better performance, vice versa. To validate this, we evaluate the proposed method on the NTIRE2018 dataset using different $n$, \ie, $n=$1, 2, 3 and 5. The obtained numerical results are provided in Table~\ref{table:n_num}. As can be seen, the reconstruction accuracy gradually increases as the number $n$ of basis functions increases. When $n=$1, the proposed method degenerates to the convolutional blocks based network, which shows the lowest reconstruction accuracy in Table~\ref{table:n_num}. When $n$ increases to $5$, the obtained RMSE is even lower than 1.0 and the PSNR is close to 50db. However, there is also no free lunch in our case and a larger $n$ often results in higher computational complexity. Therefore, we make a balance between the accuracy and efficiency by tuning $n$. This makes it possible to customize the proposed network for a specific device.
\vspace{-0.4cm}
\paragraph{Effect of the Number of FM Blocks}
In addition to the number of basis functions, the model complexity of the proposed method also depends on the number $p$ of the FM blocks. To demonstrate the effect of $p$ on the proposed method, we evaluate the proposed method on the NTIRE2018 dataset using different number of FM blocks, \ie, $p$=2,3,4 and 6. The obtained numerical results are reported in Table~\ref{table:p_num}. Similar as the case of $n$, the performance of the proposed method can be gradually improved as the number $p$ of FM blocks increases. We also find an interesting thing, increasing $n$ may be more effective than increasing $p$ in terms of boosting the performance of the proposed method.

\section{Conclusion}
In this study, to flexibly handle the pixels from different categories or spatial positions in HSIs and consequently improve the performance, we present a pixel-aware deep function-mixture network for SSR, which is composed of multiple FM blocks. Each FM block consists of one mixing function and some basis functions, which are implemented as parallel DCNN based subnets. Thereinto, the basis functions take different sized receptive fields and learn distinct mapping schemes; while the mixing function generates the pixel-wise weights to linearly mix the outputs of all these basis functions. This enables to pixel-wisely determine the receptive field size and mapping function. Moreover, we stack several such FM block in the network to further increase its flexibility in learning the pixel-wise mapping. To boost the SSR performance, we also fuse the intermediate features generated by the FM blocks for feature reuse. With extensive experiments on three benchmark SSR datasets, the proposed method shows superior performance over several existing state-of-the-art competitors.

It is worth noting that this study employs the linear mixture to approximate the pixel-wise mapping function. In the future, it is interesting to exploit the non-linear mixture. In addition, it is promising to generalize the idea in this study to other tasks requiring pixel-wise modelling, \eg, semantic segmentation, colorization \etc 

{\small
\bibliographystyle{ieee}
\bibliography{egbib}
}

\end{document}